\pdfoutput=1

\documentclass[11pt]{article}

\usepackage[preprint]{acl}
\usepackage{times}
\usepackage{latexsym}
\usepackage{makecell}
\usepackage{tcolorbox}
\usepackage{paralist}
\usepackage{fontawesome5}

\newtcolorbox{Summary}{
    colback=gray!10,
    colframe=black,
    boxrule=0.5pt,
    arc=4pt,
    auto outer arc,
    left=4pt,
    right=4pt,
    top=2pt,              
    bottom=2pt,           
    boxsep=2pt,           
    before skip=10pt,      
    after skip=0pt,       
    fonttitle=\bfseries
}

\usepackage[T1]{fontenc}

\usepackage[utf8]{inputenc}

\usepackage{microtype}

\usepackage{inconsolata}

\usepackage{graphicx}

\usepackage{booktabs}
\usepackage{amsmath}
\usepackage{amssymb}
\usepackage{bbm}
\usepackage{paralist} 
\usepackage{multicol}
\usepackage{multirow}
\usepackage{subcaption}
\usepackage{xurl}
\usepackage{hyperref}
\usepackage{footmisc}

\newcommand{\inlinecomment}[1]{}

\usepackage[textsize=tiny]{todonotes}
\newcommand{\note}[4][]{\todo[author=#2,color=#3,size=\scriptsize,fancyline,caption={},#1]{#4}} 

\newcommand{\yongxin}[2][]{\note[#1]{\textbf{Yongxin}}{orange!30}{#2}}

\title{Observable Patterns Are Not Explanations:\\A Causal-Geometric Analysis of Latent Reasoning Models
}

\author{
  \textbf{Darpan Aswal$^{1}$} \quad 
  \textbf{Thomas Palmeira Ferraz$^{1,2}$} \quad 
  \textbf{Yongxin Zhou$^1$} \quad 
  \textbf{Maxime Peyrard$^1$} \\
  $^1$Universit\'e Grenoble Alpes, CNRS, Grenoble INP, LIG  \\ $^2$NAVER LABS Europe \\
  \texttt{darpan.aswal@univ-grenoble-alpes.fr}\\
  \makebox[\linewidth][c]{%
    \small
    \begin{tabular}{@{}c@{\hspace{3em}}c@{}}
      Code: \faGithub ~ \href{https://github.com/darpanaswal/InterpretingLatents}{darpanaswal/InterpretingLatents} &
      Models: {%
        \raisebox{-0.3em}{\includegraphics[height=1.3em]{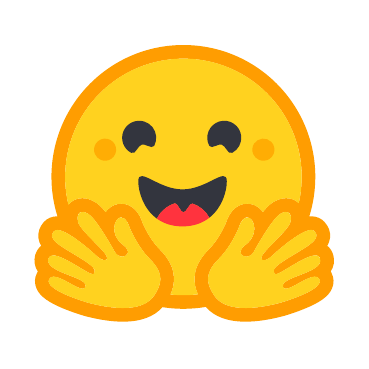}} \href{https://hf.co/collections/darpanaswal/interpretinglatents}{darpanaswal/InterpretingLatents}}
    \end{tabular}%
  }
}

\begin{document}
\maketitle

\begin{abstract}
Latent reasoning models (LRMs) replace explicit chain-of-thought with continuous thoughts.
Recent work treats observable latent-state patterns, such as BFS-like frontiers and decodable arithmetic computation, as evidence for internal reasoning mechanisms.
Evaluating two LRMs (Coconut and CODI) against controls lacking the proposed recurrence or curriculum, 
we find these patterns also appear in the controls and do not consistently affect behavior. 
Causal interventions reveal that latent-thought utilization is not binary but graded, scaling with a thought's causal effect on model behavior. Geometric analyses reveal this effect concentrates in low-dimensional subspaces whose step-wise geometry grows more structured as their behavioral influence increases.
Latent thoughts should therefore be treated as hidden computation, not hidden explanation: decodability, attention, or static structure alone cannot establish LRM reasoning mechanisms; that requires matched controls and causal tests.
\end{abstract}

\section{Introduction}
\label{sec:introduction}

Chain-of-thought (CoT) prompting has become a standard approach for eliciting reasoning through verbalized intermediate steps in natural language~\citep{wei2022chain,li2026thinking,lyu2023faithful}.
However, CoT is computationally costly and fundamentally constrained by the discreteness of natural-language tokens~\citep{zhang2026soft}.
To address these limitations, latent reasoning models (LRMs) perform intermediate computation in continuous hidden states~\citep{coconut,shen2025codi}.
LRMs promise improved efficiency, but they also remove the primary artifact available for monitoring and oversight: visible CoT traces.
Despite well-known faithfulness concerns~\citep{chen2025reasoning,turpin2023language}, CoT traces still provide human-interpretable signals that can be monitored for harmful or misaligned behavior~\citep{bereska2024mechanistic,korbak2025cotmonitorability}.
In contrast, LRMs reason through continuous states, raising new safety and interpretability concerns as they scale and are deployed in agentic systems~\citep{chan2023harms}.
This calls for a better understanding of their internal dynamics.

Early interpretability efforts have searched for latent-state analogues of CoT traces, aiming to decode interpretable patterns from latent thoughts.
For example, \citet{coconut} and \citet{reasoningbysuperposition25} argue that continuous latent thoughts demonstrate breadth-first-search-like reasoning over multiple superposed candidate reasoning paths, while \citet{shen2025codi} and \citet{wei2025sim} suggest that implicit supervision can yield interpretable intermediate reasoning steps.
Later work has also used local interventions to investigate these findings~\citep{CODImechinterp25,scratchpadthinking}.
However, several methodological concerns remain, leaving important questions unaddressed.
First, \textit{emergence}: whether the observed pattern emerges specifically from latent-reasoning mechanisms, or also appears in matched non-LRM controls.
Second, \textit{generalizability}: whether the pattern is characteristic of latent reasoning more broadly, rather than specific to one task, model, curriculum, or training recipe.
Third, \textit{causality}: whether the pattern actually contributes to the model's behavior, rather than merely correlating with successful outcomes. Without addressing these questions, observable patterns may lead to superficial evidence for reasoning.

In this work, instead of treating latent thoughts as hidden explanations, we argue in favor of treating them as extended hidden computation states whose influence strength must be measured by their causal effect on model behavior. Our approach contrasts with prior work also taking a causal perspective by asking binarily whether a latent thought is ``used'' or ``not used''~\citep{zhang2025latenttokensthinkcausal,cui2026how}, shifting the question to where causal influence is concentrated within the latent representation. These causally effective regions then become the object of analysis: once isolated, their geometry and dynamics can be studied.
This framing leads to the following research questions:

\noindent\textbf{RQ1.} \textit{Do observable patterns in LRM latent space uniquely emerge from, and explain the performance of, latent-reasoning mechanisms?}
In \S~\ref{sec:epiphenomena}, we compare LRMs against matched controls and find that reasoning patterns previously attributed to latent recurrence can also emerge in curriculum-matched non-recurrent models, and even in untrained models with forced thought positions. This shows that observable patterns alone do not sufficiently establish mechanism.

\noindent\textbf{RQ2.} \textit{When do latent thoughts influence model predictions, and where in the latent state does this influence come from?}
In \S~\ref{sec:thought_use}, we combine latent thought ablations, causal tracing, and gradient-subspace interventions to show that latent thought influence lies on a continuum.
In particular, the effect of latent thoughts is often concentrated in low-dimensional, loss-sensitive directions (a causal gradient subspace) rather than distributed across the full latent representation.

\noindent\textbf{RQ3.} \textit{Do behavior-influential latent thoughts differ from weakly influential ones in how their representations evolve across steps?}
In \S~\ref{sec:geometry}, we study the geometry and dynamics of both full latent thought trajectories and causally influential subspaces.
We find that weakly influential latent thoughts are often nearly static across steps, while behavior-influential thoughts exhibit more structured evolution.

Overall, our work shifts the target of interpretability from observational proxies to the actual causal trajectories within latent computation. Finally, we discuss practical implications in more detail in \S~\ref{sec:discussions}.

\section{Background and Related Work}


\subsection{Latent Reasoning Models}
\label{subsec:lat_reasoning_definition}

Unlike standard Chain-of-Thought (CoT) models that generate textual rationales, latent reasoning models such as implicit CoT~\citep{deng2023implicitchainthoughtreasoning} and recurrent/looped transformers~\citep{pmlr-v202-giannou23a,dehghani2018universal} replace intermediate steps with hidden state computations to bypass vocabulary projection. We focus on \emph{autoregressive continuous-thought} models which insert $K$ latent positions between the prompt $x$ and the answer $y$ by recursively feeding projected soft-tokens $e^{\mathrm{lat}}_{t+1}$ from hidden-states:
\[
    h_t = F_{\theta}\!\left(E(x), e^{\mathrm{lat}}_{1:t}\right)_{\mathrm{last}},
    \qquad
    e^{\mathrm{lat}}_{t+1}=g(h_t),
\]%
\vspace{-2pt}%
\noindent%
with $F_\theta$ a causal transformer, $E$ the token embedding map, and $g$ identity or learned\inlinecomment{transforming $h_t$ into soft-tokens $e^{\mathrm{lat}}_{t+1}$}. We call $h_t \in \mathbb{R}^d$ the \emph{latent thought}. After $K$ latent steps, autoregressive decoding resumes\inlinecomment{with each latent thought soft-tokens occupying an autoregressive position as causal context for subsequent thoughts and answer tokens}:
\vspace{-2pt}
\[
    p_\theta(y \mid x, e^{\mathrm{lat}}_{1:K})
    =
    \prod_m p_\theta(y_m \mid x, e^{\mathrm{lat}}_{1:K}, y_{<m}).
\]

\vspace{-2pt}
We study two instances: \textbf{COCONUT} gradually replaces CoT segments with latent states via a staged curriculum~\citep{coconut}, while \textbf{CODI} compresses textual rationales into latents via self-distillation~\citep{shen2025codi}. 

\subsection{Observational vs. Causal Interpretability}
Mechanistic interpretability seeks to reverse engineer neural network computations into human-understandable algorithms~\citep{bereska2024mechanistic,geiger2025causal}. A key distinction separates information that is \emph{observable} from activations from information that is \emph{causally used}: probes and visualization methods can reveal correlated structure without establishing mechanism~\citep{belinkov2022probing,jain2019attention,elazar-etal-2021-amnesic,lasri-etal-2022-probing,exp_unders}. This issue is central for LRMs, where claims about latent reasoning often rely on observational readouts, without showing that these structures drive performance. This risks conflating \textit{decodability} with \textit{mechanism}, an analogue of the \emph{dead salmon} effect~\citep{meloux2026the}, symptoms of a general tendency of interpretability research to produce false positive findings~\citep{hewitt-liang-2019-designing, ravichander-etal-2021-probing, kantamneni2025are, meloux2026the}. We rely on intervention-based methods (e.g., causal tracing, activation patching)~\citep{meng2022locating,wang2023interpretability,chan2022causal,heimersheim2024activation,monea-etal-2024-glitch} to test whether latent thoughts affect model behavior~\footnote{while recognizing that causal interventions can still admit multiple compatible explanations~\citep{meloux2025everything}.}.

\subsection{Mechanistic Analyses of Latent Reasoning}
\label{sec:mechinterp_latent_reasoning}

\paragraph{Observable Latent Patterns.}
Prior work finds structured intermediates in latent states. In CODI, logit-lens, attention-mass, and activation-patching analyses are used to argue that models use latent thoughts as a \textit{computational scratchpad} for reasoning: at each step, these states store operands and intermediate values for subsequent steps~\citep{shen2025codi, CODImechinterp25,scratchpadthinking}; however, the evidence remains largely single-task and observational.~\citet{coconut} interpret Coconut's latent thoughts as superposed reasoning paths with BFS-like dynamics, representing multiple candidate nodes at increasing graph depths, progressively concentrating mass on target-reaching nodes. Later work attributes similar patterns to shortcut behavior and task-specific heuristics~\citep{cui2026how, rizvi-martel2026the}. Our work investigates whether such observable patterns reflect genuine reasoning mechanisms.

\paragraph{Shortcut Behavior.}
Continuous thoughts may be decodable without driving reasoning: models can solve tasks via prompt or KV-cache shortcuts, or commit to answers before latent computation. Prior work frames thought use as binary, showing performance persisting under perturbations and ablations of latent states~\citep{zhang2025latenttokensthinkcausal, cui2026how},
\inlinecomment{and traces collapsing to partial intermediates or direct answer mappings instead of faithful step-by-step reasoning~\citep{liang2026latentcotmodelsthinkstepbystep,LatentReasoningInterpretable2026}.}
suggesting stronger supervision as the cure. However, such findings cannot establish whether models are simply incapable of using their thoughts or if they just find other circuits in their absence. We therefore treat latent-thought influence as a graded and localizable causal quantity, motivating the question of whether this influence is concentrated in specific directions or distributed~\citep{gur2018gradient,bernas2026revisiting}.

\paragraph{Geometry, Stability, and Latent Dynamics.}
A complementary perspective on LLM interpretability studies reasoning as movement through the representation space~\cite{elhage2022toy,park2024the,gurnee2024language,geva2023dissecting,bhatia-etal-2025-date,bhatia2026reallycontrolstemporalreasoning,bernas2026revisiting,zhou2026the,kumar2026geometriclatentreasoninginduces}. In particular, \citet{MarkovCoTHelp2026} model explicit CoT as a Markov chain, with CoT being useful only when transitions are stable and aligned. However, such analyses remain limited in LRM interpretability. \citet{reasoningbysuperposition25} compare the geometry of latent thoughts with optimal node embeddings, suggesting a superpositional graph-search representation. \citet{wei2025sim} report that scaled implicit reasoning can become unstable, with representations homogenizing and drifting from the vocabulary manifold. These studies investigate latent state evolution, not its causal effect on model behavior, motivating pairing geometry with intervention.

\paragraph{Concurrent Causal Analyses.}
 Recent work has also examined the causal role of latent thoughts. \citet{li2026dynamics} study how causal influence is distributed across latent-thought steps, finding uneven and non-local effects and a gap between early output bias and later representational commitment. \citet{unlockingblackboxlatent2026} combine semantic, geometric, and causal analyses of individual latent thoughts, identifying early thoughts as structured and steerable causal hubs. Complementarily, \citet{seddik2026formalizinglatentthoughtsaxioms} propose causality, minimality, separability, and stability as desiderata for thought representations, finding that downstream accuracy can mask representational failures. Our work complements these analyses with within-thought causal localization, isolating low-dimensional, loss-sensitive directions that influence behavior.

\section{Experimental Setup}
\label{sec:experimental_setup}
\paragraph{Datasets.}
We evaluate graph-hopping on ProsQA~\cite{coconut} and arithmetic-reasoning on GSM8k~\cite{cobbe2021training}.
%
%
\paragraph{Models and Controls.}
All models are trained separately for each task from pretrained GPT-2 small~\cite{radford2019language} (Llama-3.2-1B-Instruct~\cite{meta_llama32} extension in Appendix~\ref{app:llama_results}). Our design compares target LRM models with \underline{different curricula} but the \underline{same recurrence mechanism} (\textbf{Coconut (C)} and \textbf{CODI}), and the following controls:%
\begin{list}{$\bullet$}{
    \setlength{\leftmargin}{0.5em}
    \setlength{\labelwidth}{0.6em}
    \setlength{\labelsep}{0.2em}
    \setlength{\itemsep}{0.05em}
    \setlength{\topsep}{0.05em}
    \setlength{\parsep}{0pt}
    \setlength{\partopsep}{0pt}
} 
    \item \textbf{Recurrence control:} \textbf{Pause-as-thought (PaT)} follows Coconut's format and curriculum but replaces recurrence with learned thought tokens.
    \item \textbf{Curriculum control:} \textbf{Coconut$_u$ (C$_u$)}, a curriculum-perturbed Coconut variant that samples other stages with probability $u=0.3$.
    \item \textbf{Observational controls:} \textbf{Base GPT-2 (B)} and \textbf{Explicit-CoT GPT-2 (CoT)} are included in \S~\ref{sec:epiphenomena} to test whether similar patterns can be recovered without latent-reasoning training, either from the probe itself or from ordinary task-solving. They are excluded from causal and geometric analyses, which require dedicated latent thought positions.
\end{list}%
Details on models and training are in Appendix~\ref{app:models_and_hyperparameters}.
%
\paragraph{Statistical Evaluation.}
We report 95\% bootstrapped confidence intervals (1{,}000 resamples) and use McNemar $p$-values for paired comparisons. Details and significance results are in Appendix~\ref{app:statistical_tests}.
%
%
%
%
%


\begin{figure*}[!h]
    \centering
    \begin{subfigure}[t]{\linewidth}
        \centering
        \includegraphics[width=\linewidth]{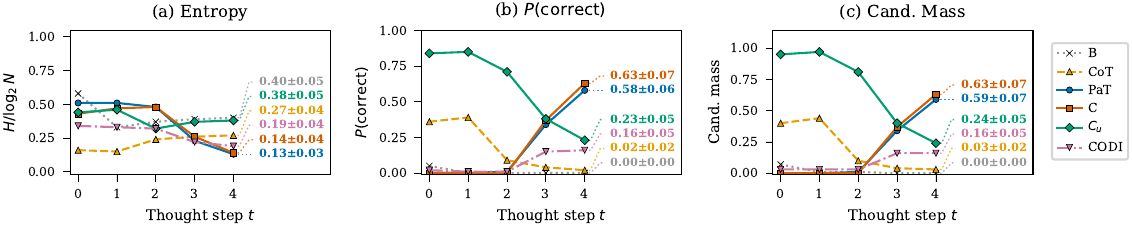}
        \caption{Case Study 1: Superposition and BFS-like search on Graph-Hopping}
        \label{fig:epiphenomena_superposition}
    \end{subfigure}
    
    \vspace{0.5em}
    
    \begin{subfigure}[t]{\linewidth}
        \centering
        \includegraphics[width=\linewidth]{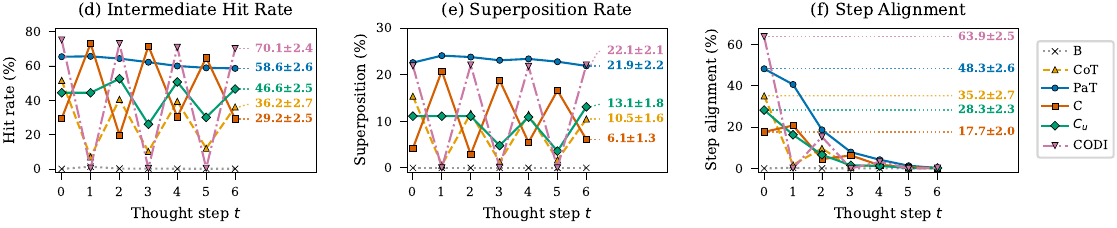}
        \caption{Case Study 2: Scratchpad Thinking on Arithmetic-Reasoning via Logit-Lens}
        \label{fig:epiphenomena_scratchpad}
    \end{subfigure}
    
    \caption{Probing for breadth-first-search (BFS) patterns on graph-hopping (top) and scratchpad-thinking on arithmetic-reasoning (bottom). \textbf{\textit{Takeaway:}} Matched controls reproduce or invert observable patterns, showing they are not specific to the proposed recurrence or curriculum mechanisms and, thus, are insufficient alone for mechanistic attribution.}
    \label{fig:epiphenomena}
\end{figure*}%
\section{Observable Structures Are Insufficient for Mechanistic Attribution}
\label{sec:epiphenomena}

We revisit two LRM interpretability patterns from prior work, discussed in \S~\ref{sec:mechinterp_latent_reasoning}: Coconut's apparent simultaneous encoding of multiple candidate reasoning paths (\textit{superposition}) and progressive refinement toward the correct path (\textit{breadth-first-search exploration}) on graph-tasks, and CODI's claimed reasoning through decodable intermediate computations (\textit{scratchpad}) on arithmetics. To address RQ1, we compare models to non-LRM model and curriculum controls introduced in \S~\ref{sec:experimental_setup}.


\subsection{Case Study 1: Superposition and BFS-like search on Graph-Hopping}
Following~\citet{coconut}, we probe the \emph{depth frontier} after $k$ latent thoughts by measuring the unnormalized joint probability $p(\text{concept}) = \prod_i p(\text{tok}_i \mid \text{ctx},\ \text{``Every''},\ \text{tok}_{<i})$ over candidate nodes at depth $\max(k,1)$ from the root: children for $k{=}1$; grandchildren for $k{=}2$.~\citet{coconut} read this as two patterns. \textit{Superposition}: early latent thoughts hold multiple candidates simultaneously (high entropy; candidate mass spread across nodes). \textit{BFS}: increasing mass on target-reaching nodes with recurrence (rising $P(correct)$; falling entropy). If recurrent feedback drives this BFS-like exploration, C should show this pattern more strongly than PaT. Results in Figure~\ref{fig:epiphenomena_superposition}. 


B and CoT, lacking latent training, never form the frontier (B near-zero mass; CoT concentrates mass early but degrades with depth). While C reproduces the BFS signature (rising $P(correct)$, falling entropy), PaT matches it with no recurrence (but same curriculum). Moreover, C$_u$ which perturbs C's curriculum inverts this pattern: high mass at shallow depth that \textbf{degrades} with depth. CODI, which lacks the curriculum, weakly mirrors C and PaT. Candidate mass tracks $P(correct)$ closely, indicating that frontier mass concentrates on target-reaching nodes. However, superposition requires competing incorrect candidates held simultaneously. Thus, BFS signature is not recurrence-specific: PaT matches it without recurrence, while C$_u$ changes it with recurrence preserved.%
%
%
\subsection{Case Study 2: Scratchpad Thinking on Arithmetic-Reasoning via Logit-Lens}
Following~\citet{shen2025codi}, we project final-layer hidden states at thought positions through the LM head and compare decoded tokens against ground-truth intermediate CoT annotations.~\citet{shen2025codi} read two patterns as evidence of latent scratchpad. \textit{Hit Rate}: decoded tokens match a ground-truth CoT intermediate. \textit{Step Alignment} (ordered-scratchpad): the $t$-th thought matches the $t$-th CoT step. We additionally test Superposition Rate (fraction of steps decoding $\geq$ 2 distinct intermediates) to test generalizability of~\citet{coconut} claims~\footnote{We also track top-1 decoded token changes across steps to assess trajectory structure on both tasks (Appendix~\ref{app:extended_analyses}).}. Results in Figure~\ref{fig:epiphenomena_scratchpad}.

PaT lacks recurrence but exceeds CODI on hit-rate overall. C, CoT (no latent training) and C$_u$ also reproduce CODI-like step-wise alternation, although with different magnitudes and phase structure. This suggests the pattern is not specific to CODI's latent-distillation objective: it could reflect CoT fine-tuning, probe-setup, or another source, but its appearance in controls already contradicts that attribution. Moreover, step-alignment decays sharply with depth for every model, contradicting an ordered scratchpad. B is near zero throughout. Superposition stays low across all models, failing to reproduce COCONUT's frontier in arithmetic. Overall, the logit-lens readouts recover scratchpad-like signatures without uniquely tracking performance or requiring CODI's distillation mechanism.


\begin{Summary}
    \paragraph{Summary.}
    Both case studies reproduce the original reported findings but challenge the conclusions. 
    Similar patterns in non-LRM controls indicate that the observed patterns are not specific to LRMs. Going beyond observable patterns requires a causal perspective.
    
\end{Summary}



\section{When and How do LRMs use Latent Thoughts? }
\label{sec:thought_use}


Having shown that observational readouts are not specific enough for mechanistic attribution, we now investigate how latent thoughts causally influence model behavior. 

\begin{figure*}[!t]
  \centering
  \includegraphics[width=\textwidth]{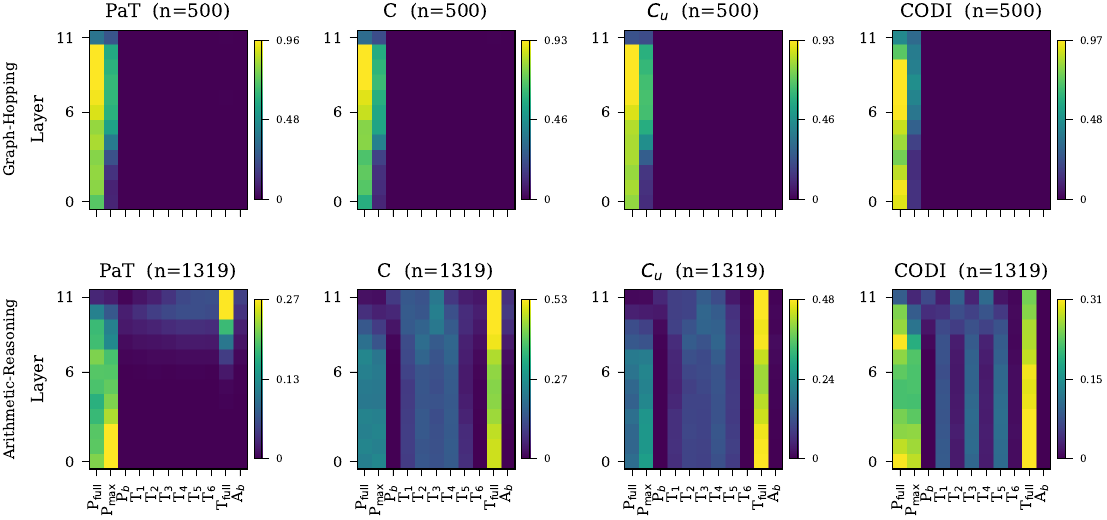}
  \caption{Per-layer $\mathrm{IE}_{\mathrm{KL}}$ on the residual stream under partner-prompt corruption across buckets $\{P_{\text{full}}, P_{\max}, P_b, T_1,\dots,T_K, T_{\text{full}}, A_b\}$. Prompt positions recover best across models and tasks. Thought positions $T_t$ contribute near zero on graph-hopping; on arithmetic-reasoning, $T_{\text{full}}$ yields the strongest recovery for C, $C_u$, and CODI. \textbf{\textit{Takeaway:}} Latent-thought influence is graded---when large, it can override corrupted prompt positions and steer models toward correct paths.}

  \label{fig:causal_trace}
\end{figure*}

\begin{table}[!t]
\centering
\small
\setlength{\tabcolsep}{3pt}
\begin{tabular}{lcc|cc}
\toprule
Model & \makecell{Graph- \\ Hopping} & \makecell{Thoughts \\ Removed} & \makecell{Arithmetic- \\ Reasoning} & \makecell{Thoughts \\ Removed} \\
\midrule
B     & $2.4 {\scriptstyle \pm 1.3}$  & -- & $1.4 {\scriptstyle \pm 0.6}$  & -- \\
CoT   & $79.0 {\scriptstyle \pm 3.6}$ & -- & $41.9 {\scriptstyle \pm 2.7}$ & -- \\
PaT   & $95.4 {\scriptstyle \pm 1.8}$ & $95.6 {\scriptstyle \pm 1.7}$ & $26.4 {\scriptstyle \pm 2.4}$ & $21.4 {\scriptstyle \pm 2.2}$ \\
C     & $98.0 {\scriptstyle \pm 1.3}$ & $97.8 {\scriptstyle \pm 1.3}$ & $35.7 {\scriptstyle \pm 2.5}$ & $7.7  {\scriptstyle \pm 1.4}$ \\
C$_u$ & $96.0 {\scriptstyle \pm 1.7}$ & $91.8 {\scriptstyle \pm 2.4}$ & $30.8 {\scriptstyle \pm 2.5}$ & $39.1 {\scriptstyle \pm 2.6}$ \\
CODI  & $80.0 {\scriptstyle \pm 3.5}$ & $80.0 {\scriptstyle \pm 3.5}$ & $41.8 {\scriptstyle \pm 2.7}$ & $25.6 {\scriptstyle \pm 2.4}$ \\
\bottomrule
\end{tabular}
\caption{Accuracy under thought-ablation at test-time. Performance degrades on arithmetic-reasoning; graph-hopping is unaffected. \textbf{\textit{Takeaway:}} Influence of latent thoughts on model performance is task-varying.}
\label{tab:interventions}
\end{table}

\paragraph{Latent Thought Ablation at Test-Time.} First, we evaluate whether models require latent thoughts at all by removing them at test-time: $K{:=}K_{\max}\!\to\! 0$ (recurrence skipped for $C, C_u,\mathrm{CODI}$; the $k$ parallel tokens dropped for PaT). Results in Table~\ref{tab:interventions}.

On \textbf{graph-hopping}, only $C_u$ is affected meaningfully.\inlinecomment{\yongxin[]{C slight drop}} On \textbf{arithmetic-reasoning}, $C_u$ \textit{improves} but other models suffer major drops. Causal \textit{necessity} of latent thoughts is task-dependent, but the shortcut-learning dichotomy (used or bypassed) warrants further checks due to C$_u$'s behavior.
%
%
\paragraph{Per-Position Causal Tracing.}
Does maintaining performance after removing thoughts imply they weren't causally influencing computation when present? LLMs are known to be causally over-determined~\cite{mcgrath2023hydraeffectemergentselfrepair,meloux2026the}: multiple circuits computing the same behavior~\cite{lan2024towards}; when present, thoughts are still part of the computational paths and can be expected to causally impact behavior. To test this, we extend~\citet{meng2022locating}'s causal tracing methodology to latent thoughts. 


Define site $s=(\ell,p)$ as a layer $\ell$ and position $p$ in the residual stream~\cite{elhage2021mathematical}~\footnote{Per component (attention, MLP) decompositions appear in Appendix~\ref{app:extended_analyses}.}. We run a \emph{clean} pass, a \emph{corrupted} pass (cp) on a partner prompt (dataset instance with different answer), and a \emph{patched} pass (pp) that injects site $s$'s clean activation into the corrupted forward pass. At the answer boundary $A_b$, we greedily decode $N$ tokens, recording at each step $j$ the full next-token distribution $P^{(j)}_\bullet = \text{softmax}(z_j)$ for each pass. We compute the forward KL from the clean distribution $\mathrm{KL} \big(P^{(j)}_\mathrm{clean}\|P^{(j)}_\bullet \big)$, and average over the content window $W = [n_\mathrm{fmt}, e)$ spanning the clean pass output, where $n_\mathrm{fmt}$ skips the fixed format prefix (e.g.,\ \texttt{\#\#\#}, \texttt{The answer is:}), and $e$ is the first end-of-text step (or $N$). Patching thus measures recovery of the model's clean behavior:%
\[
\resizebox{\columnwidth}{!}{$\displaystyle \overline{\mathrm{KL}}_\bullet=\tfrac{1}{|W|}\sum_{j\in W}\mathrm{KL}\big(P^{(j)}_{\mathrm{clean}}\|P^{(j)}_\bullet\big),\quad \mathrm{IE}(s)=1-\overline{\mathrm{KL}}_{\mathrm{pp}}/\overline{\mathrm{KL}}_{\mathrm{cp}}$} \]
The indirect effect $\mathrm{IE}(s)\le 1$ measures how far the patch restores the clean output: $1$ is full recovery, $0$ none, and $<0$ is worsening. We report the $\arg\max_\ell$-IE $(s)$ over the position bucket $\{P_{\mathrm{full}},P_{\max},P_b,T_1,\dots,T_K,T_{\mathrm{full}},A_b\}$, where $T_t$ is thought $t$; $P_b,A_b$ the latent-block delimiters; $P_{\max}$ the strongest single prompt token; and $P_{\mathrm{full}}/T_{\mathrm{full}}$ the jointly patched prompt \& thought positions. Results in Figure~\ref{fig:causal_trace}.


\begin{figure*}
    \centering
    \includegraphics[width=\linewidth]{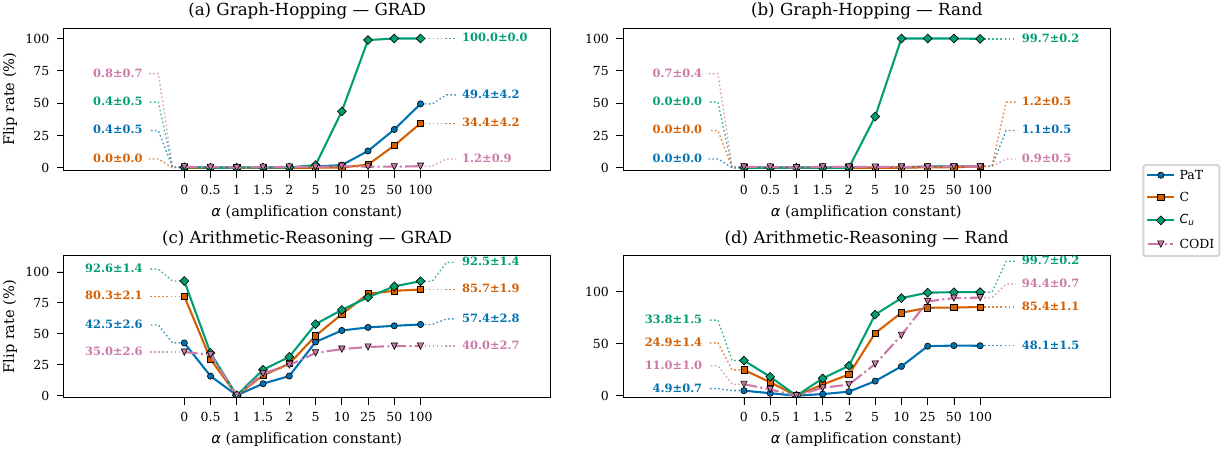}
    \caption{Gradient-subspace intervention flip rates (\%) across amplification strength $\alpha$ on the gradient-derived vs random control directions. \textbf{Graph-Hopping:} Robust to ablation ($\alpha=0$) but flips at high-$\alpha$ on GRAD. \textbf{Arithmetic-Reasoning:} Degradation under GRAD ablation; GRAD and Rand amplifications cause comparable flip-rates. \textbf{\textit{Takeaway:}} Thought utilization is \textit{graded} and depends on their influence power over model behavior.}
    \label{fig:amplification_heatmap}
\end{figure*}

Prompt positions (especially $P_\text{full}$) strongly recover performance across models on \textbf{graph-hopping}. Across all models, thoughts show zero recoverability. Even C$_u$ shows none despite its minor (4.8\%) thought necessity in the ablation experiment (possibly due to it being averaged over the full dataset). On \textbf{arithmetic-reasoning}, prompt-positions of PaT and CODI still show very high recoverability throughout layers. Individual thought positions also show recoverability rates albeit small. Interestingly, patching all thought positions while other positions are corrupted shows the best recoverability through all layers on C, C$_u$ and CODI and through the later layers on PaT.


\paragraph{Gradient-Subspace Interventions.}
The previous analyses test whether latent thoughts influence model behavior at test-time. Prior work typically frames this influence as binary: unused thoughts are simply bypassed shortcuts~\cite{zhang2025latenttokensthinkcausal, cui2026how}. But \textit{why?} Do these thoughts lack the capacity to influence behavior, or is their effect simply too small to measure globally? To test this, we isolate the representation directions most sensitive to the loss, geometrically concentrating thought influence into a subspace for intervention.


For thought $h_{i,t}\in\mathbb{R}^{D}$ (instance $i$, position $t$), loss $\mathcal{L}_i$ and gradients $g_{i,t} = \nabla_{h_{i,t}}\mathcal{L}_i$, we define the gradient-subspace $B_t \in \mathbb{R}^{D\times k_t}$ using the top-$k_t$ right singular vectors of the gradient matrix $G_t = [g_{1,t};\dots;g_{N,t}]$, capturing $95\%$ cumulative energy. To intervene, we split $h_{i,t}$ as $h_{i,t}^{B} = B_t B_t^\top h_{i,t}$ (projection onto $B_t$) and $h_{i,t}^{\perp} = h_{i,t} - h_{i,t}^{B}$. The projection is scaled by $\alpha \in \{0, 0.5, 1, 1.5, 2, 5, 10, 25, 50, 100\}$ via the update $h_{i,t} \leftarrow h_{i,t}^{\perp} + \alpha h_{i,t}^{B} = h_{i,t} + (\alpha-1) B_t B_t^\top h_{i,t}$ which ablates ($\alpha=0$) or amplifies ($\alpha>1$) the targeted component. A rank-matched orthonormal basis $B_t^{\mathrm{rand}}$ via QR decomposition on Gaussian noise serves as the control. Except for PaT, models skip $t=K$ as the gold answer token sits at step $K$ (meaning $G_K$ is empty). Results in Figure~\ref{fig:amplification_heatmap}.


On \textbf{graph-hopping}, ablation preserves accuracy, but PaT and C exhibit high-$\alpha$ GRAD flip-rates significantly exceeding Rand, indicating thoughts retain a \textit{graded} causal influence rather than being strictly \textbf{bypassed}. $C_u$ is an exception: despite degrading under thought ablation, the gradient directions fail to localize its influence to a concentrated subspace. On \textbf{arithmetic-reasoning}, GRAD ablation degrades all models but Rand is inert; both amplifications produce comparable flip-rates~\footnote{Subspace ranks in Appendix~\ref{app:extended_analyses} (Table~\ref{tab:subspace_dim_full_gpt2}).}.


\begin{Summary}
    \paragraph{Summary.}
    Thought utilization is \textit{graded} rather than binary, and depends on how much influence the latent thoughts carry on model behavior. Together with \S~\ref{sec:epiphenomena}, this motivates moving from finding low-level observable latent patterns to studying the dynamics of behavior-influential directions across latent steps.
\end{Summary}

\section{The Dynamics and Geometry of Latent Thoughts}
\label{sec:geometry}


Observable patterns such as BFS and scratchpad-thinking (\S~\ref{sec:epiphenomena}) are not sufficient, on their own, to establish causal contribution to model behavior.
In this section, we argue that rather than drawing general conclusions about any latent reasoning paradigm, the dynamic evolution of latent thoughts (and the directions where their causal influence concentrates) can offer more meaningful insights into how a specific $\{\text{model, task, latent--training--recipe}\}$ configuration behaves. Using the behavior-influential directions identified in \S~\ref{sec:thought_use}, we now apply geometric tools to assess whether the dynamics of highly causally influential latent thoughts differ from those of less causally influential ones.

\begin{figure}
    \centering
    \includegraphics[width=1\linewidth]{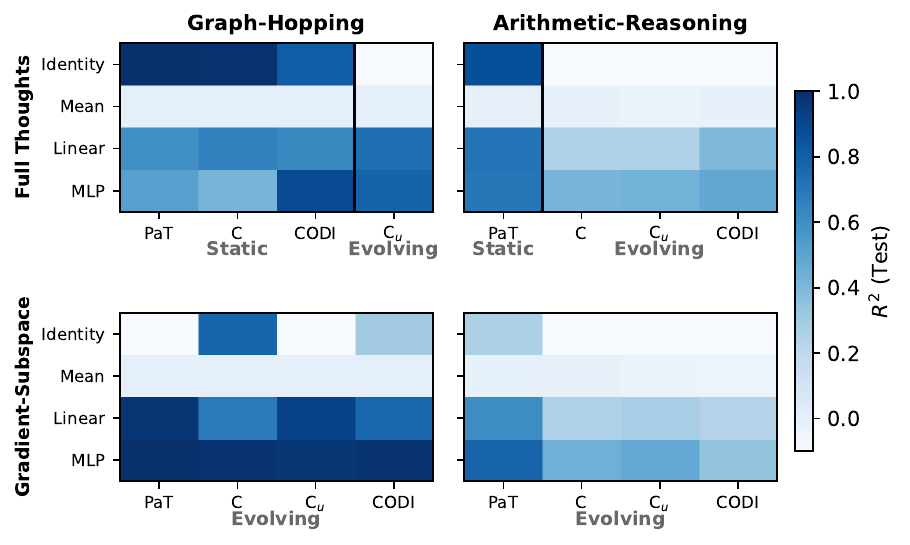}
    \caption{Markovianity of latent thoughts. \textbf{Graph-Hopping:} Static dynamics (except C$_u$); \textbf{Arithmetic-Reasoning:} Evolving dynamics. \textbf{\textit{Takeaway:}} Even static appearing full-thoughts show evolving dynamics within the gradient-subspace where influence concentrates.}
    \label{fig:markov}
\end{figure}

\paragraph{Markovianity of Thought Trajectories.} 
We first investigate the evolution of thought trajectories, testing whether future states can be predicted from preceding ones. By design, the transformer architecture makes every thought a function of all previous thoughts and prompt tokens, but the trajectories may encode a simpler approximate structure.

For the thoughts $h_{i,1}, \ldots, h_{i,K}$ for every test instance~$i$, we form pairs $X_{i,t} = [h_{i,t-1};\ldots;h_{i,t-n}]$ and $Y_{i,t} = h_{i,t}$ for a Markov order $n$. We fit a shared map $f:\mathbb{R}^{nD}\!\to\!\mathbb{R}^D$ on the train split and report uniform-average $R^2$ on the test split under two function classes: ridge-regularized linear regression and a two-layer MLP : $h_t = W_2\,\sigma(W_1 X + b_1) + b_2$ with $\sigma{=}\mathrm{GELU}$, $W_1 \in \mathbb{R}^{256 \times nD}$ (early-stopped on a 10\% train slice, $R^2$ averaged over 3 seeds). We compare against the \emph{mean baseline} $\hat h_t = \bar h_{\text{train}}$ ($R^2 = 0$ on train split), and the \emph{identity baseline} $\hat h_t = h_{t-1}$. Order $n=1$ is the strict first-order claim $h_t = f(h_{t-1})$; we sweep $n \in \{1,\ldots,5\}$. The same analysis on the gradient-subspace (\S~\ref{sec:thought_use}) projected thoughts $h^B_{i,t} = B_t B_t^\top h_{i,t}$ tests thought dynamics where influence concentrates. Results (averaged over $n$) in Figure~\ref{fig:markov}.

On \textbf{graph-hopping}, the identity baseline dominates for PaT, C, and CODI: no fitted transition meaningfully improves over simply copying the previous thought, but collapses on C$_u$~\footnote{Static requires $\max(R^2_{\text{id}}, R^2_{\text{mean}}) \ge \max(R^2_{\text{lin}}, R^2_{\text{MLP}}) - 0.1$ to avoid overfitting.} for which linear and MLP are the best approximates. On \textbf{arithmetic-reasoning}, identity dominates on PaT but collapses on C, C$_u$, and CODI, all best characterized by the MLP while linear closely follows. The dynamics change in the gradient-subspace; the MLP map nearly captures the full \textbf{graph-hopping} dynamics and dominates on \textbf{arithmetic-reasoning}. 

\paragraph{Geometric Stability of Gradient-Subspaces.} 

Next, we ask whether the gradient-subspace (\S~\ref{sec:thought_use}) itself remains stable throughout the reasoning trajectory. For adjacent time-steps, we measure the mean squared principal-angle similarity between subspaces: $s_t = \mathrm{mean}\!\left(\sigma(B_t^\top B_{t+1})^2\right),$ where $\sigma(\cdot)$ denotes the singular values of the inter-basis projection matrix. Values of $s_t \to 1$ indicate stable causal subspaces, while $s_t \to 0$ indicate orthogonality.


\begin{figure}[t]
    \centering
    \includegraphics[width=1.0\linewidth]{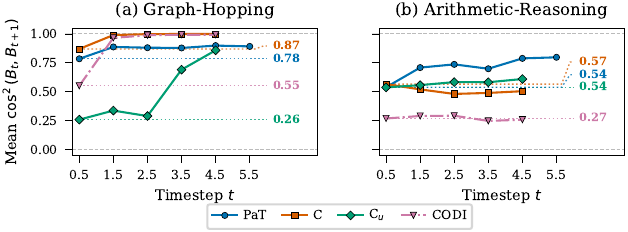}
    \caption{Geometric stability of the gradient-subspaces. \textbf{Graph-Hopping:} Stable subspaces (except C$_u$); \textbf{Arithmetic-Reasoning:} Rotating subspaces. \textbf{\textit{Takeaway:}} Thoughts with weak influence over model behavior exhibit highly stable gradient-subspace geometries. Points plotted at half-integers as each measures an adjacent-step pair $\cos^2(B_t, B_{t+1})$.}
    \label{fig:geometric_stability}
\end{figure}

Figure~\ref{fig:geometric_stability} further localizes this distinction to the geometry of the gradient-subspace. On \textbf{graph-hopping}, PaT, C and CODI maintain highly stable subspaces while $C_u$ shows near-orthogonal early rotations before stabilizing. On \textbf{arithmetic-reasoning}, stability is lower overall: PaT remains most stable, C and $C_u$ show moderate alignment, and CODI undergoes large step-to-step rotations.

\begin{Summary}
    \paragraph{Summary.}
    Beyond their causal necessity and sufficiency, the causally influential directions in latent thoughts are consistently subspace-concentrated, and their \textit{dynamics} track influence. Where causal effect is weak (e.g, graph-hopping), full trajectories barely evolve past the first thought, though the causally active subspace supports non-trivial computation; where stronger, trajectory dynamics are more structured, and the causally active subspace supports more diverse computation. 
\end{Summary}

\section{Discussion}
\label{sec:discussions}

\paragraph{A causal-first view of latent thoughts.}
LRM interpretability is especially vulnerable to false explanations: decodable intermediates, attention mass, and visually salient geometries can appear even when the proposed mechanism is absent. Observational tools are most natural when computation is partly exposed in token space; in LRMs, they can recover structure from continuous states that need not coincide with what the model uses to answer. This makes an observational-first workflow risky. We therefore reverse the order of analysis: rather than starting from a readable latent pattern, we first test whether latent thoughts causally affect behavior and where this influence is concentrated. Only then do we analyze the geometry and dynamics of these behavior-influential regions. Geometry is informative when it describes the part of latent computation that affects predictions.

\paragraph{From thought use to causal localization.}
This causal-first view helps reframe the shortcut-vs-reasoning dichotomy. The relevant question is not whether latent thoughts are simply ``used'' or ``bypassed'', but how much they influence behavior and in which directions of the representation. Concurrent work similarly finds that causal influence can vary substantially across latent steps and latent states can be steered through intervention \citep{li2026dynamics,unlockingblackboxlatent2026}. Our results further show that this influence varies across tasks with different complexity and structure, and can concentrate in specific directions within a latent state. Moreover, a model may solve a task without requiring latent thoughts under ablation while still containing latent directions that can steer its predictions when intervened on. Such cases do not show that the model is incapable of using latent thoughts; they may instead reflect alternative circuits that solve the task when latent computation is removed. Conversely, forcing stronger dependence on latent thoughts (for instance by stronger supervision) does not by itself show that these thoughts implement faithful reasoning. The more informative unit of analysis is therefore the strength and localization of latent-thought influence for a given model and task.

\paragraph{Implications for LRM interpretation and design.}
Causal localization also gives interpretability a constructive role. Once behavior-influential regions are identified, they provide concrete targets for geometric analysis, monitoring, and downstream interventions. For example, if influence is consistently concentrated in low-dimensional, loss-sensitive subspaces, compression could focus on preserving the task-relevant component of the latent state rather than the full representation. Along these lines, concurrent work already exploits low-rank CoT geometry to distill reasoning into compact latent trajectories \citep{solgi2026lorilowrankdistillationimplicit}. Similarly, if behavior-influential directions are shown to follow stable or predictable step-to-step dynamics, they could inform simplified transition models or more targeted recurrent computation, which is particularly relevant because the recurrent mechanism described in \S~\ref{subsec:lat_reasoning_definition} can impose sequential dependencies that are difficult to parallelize during training. 


\paragraph{Semantic readability and causal influence are complementary axes.}
Latent computation can also be interpreted through semantically readable coordinates, but readability and behavioral influence are distinct: a direction can be readable without being causal. The Jacobian-lens work is one instance---a sparse, vocabulary-indexed J-space read as a ``language-model global workspace''~\cite{gurnee2026GlobalWorkspaceJacobian}, extended to LRMs as recurrent J-thoughts~\cite{wu2026jcotchainofthoughtjspace}. These semantic directions complement our loss-gradient subspaces, which instead localize causal influence; yet~\citet{SurfacingModelAlgorithms} find J-lens readouts that suggested a strategy the directions did not causally support. This reinforces our causal-first view: readable coordinates can propose computational variables, but only interventions establish their mechanistic relevance.

\paragraph{Relation to CoT faithfulness concerns.}
Our findings extend documented concerns about explicit CoT: verbalized rationales can misrepresent a model's internal reasons~\cite{turpin2023language, arcuschin2026chainofthought}, and intermediate steps (including ``aha'' or self-verification moments) may exert little causal influence on the final answer~\cite{boppana2026reasoning, zhao2025can}. LRMs inherit this concern while removing the surface artifact that makes CoT at least partially inspectable. The risk is therefore not only opacity, but false interpretability: latent states may contain decodable or geometrically structured information that appears explanatory without being behaviorally relevant, strengthening the case for causal-first monitoring.

Overall, latent thoughts should be treated as hidden computation, not hidden explanation. Instead of readable latent patterns, LRM interpretability should target the regions of latent computation that causally shape behavior and analyze their dynamics. This causal-first view provides a more grounded and principled basis for designing and auditing latent reasoning systems.

\section*{Limitations}
\label{sec:limitations}
First, we estimate the gradient subspace linearly via SVD; a nonlinear estimator (e.g., autoencoder) could capture causal structure a linear basis flattens. Relatedly, our loss-gradient subspaces localize behavioral sensitivity but do not identify the semantic content of those directions. A direct next step is to test whether behavior-influential latent directions admit sparse J-space~\cite{gurnee2026GlobalWorkspaceJacobian} descriptions. Second, our causal evidence relies on local intervention-based methods with known caveats: causal tracing and gradient-subspace interventions probe behavior under specific perturbations that can shift the model off-distribution, and interventions admit multiple compatible mechanistic explanations, so ``behavior-influential'' is a necessary-but-not-sufficient signal of mechanism, \textbf{not} proof. Third, our conclusions draw on relatively small models at modest latent steps (K=6), and our LRM coverage is partial (PaT, COCONUT, CODI); future work should extend to larger model families and more recent reasoning paradigms. Fourth, our models are trained separately for each task, so task differences may partly reflect task-specific training; multi-task models would allow within-model comparisons. We also analyze only final checkpoints, leaving how observable patterns and causal subspaces develop during training unexplored. To facilitate such training dynamics analyses, we release all intermediate checkpoints in our \href{https://huggingface.co/collections/darpanaswal/interpretinglatents}{model collection}.



\section*{Ethical Considerations and Societal Impact}
\label{sec:ethics_societal_impact}
Latent reasoning models may have substantial societal impact if they make multi-step reasoning more efficient, scalable, and easier to integrate into agentic systems. Yet, by replacing explicit chain-of-thought traces with continuous hidden-state computation, LRMs remove a surface artifact that can be inspected, filtered, or audited. This creates two related risks. First, without reliable tools for interpreting LRMs, users, developers, and auditors may have fewer signals for detecting harmful, deceptive, or misaligned behavior, especially in high-stakes domains such as education, healthcare triage, legal assistance, scientific automation, or autonomous agents. Second, unreliable interpretability tools can create false interpretations, which may lead to monitoring the wrong features and missing the computations that actually drive harmful or misaligned behavior. Our work contributes towards addressing this risk by arguing that latent thoughts should be treated as hidden computational states, not hidden explanations, and that mechanistic claims require controls and causal tests before latent structure is interpreted. This causal-first view is not a bullet-proof procedure that solves this completely, since interventions can still be local, distribution-shifting, or compatible with multiple explanations. Rather, it provides a stricter evidential standard that should be combined with stress testing, adversarial evaluation, uncertainty reporting, and domain-specific oversight.

\makeatletter
\ifacl@anonymize\else
\section*{Acknowledgements}

This work was conducted within the French research unit UMR 5217 and was supported by CNRS (grant ANR-22-CPJ2-0036-01 and ANR-25-CE23-2059-01) and by MIAI@Grenoble-Alpes (grant ANR-19-P3IA-0003 and ANR-23-IACL-0006). This project was provided with HPC computing and storage resources by GENCI at IDRIS thanks to the grant \texttt{2026-AD011017855} on the supercomputer Jean Zay's V100 partition. Thomas' research is also partially supported by the French Ministry of Higher Education, Research and Innovation under CIFRE PhD Convention \textit{``Reconcile Efficiency and Interpretability in Structured Planning and Reasoning''} (no. \texttt{2025/0487}).

\paragraph{Use of AI Assistants.} AI tools were used during this research for manuscript language editing and polishing, and assistance with editing software code. All AI-generated outputs were reviewed and verified by the authors, who take full responsibility for the scientific content, experimental results, interpretations, and conclusions.
\fi
\makeatother

\bibliography{references}

@inproceedings{
  coconut,
  title={{Training Large Language Models to Reason in a Continuous Latent Space}},
  author={Shibo Hao and Sainbayar Sukhbaatar and DiJia Su and Xian Li and Zhiting Hu and Jason E Weston and Yuandong Tian},
  booktitle={Second Conference on Language Modeling},
  year={2025},
  url={https://openreview.net/forum?id=Itxz7S4Ip3}
}

@inproceedings{ravichander-etal-2021-probing,
    title = "Probing the Probing Paradigm: Does Probing Accuracy Entail Task Relevance?",
    author = {Ravichander, Abhilasha  and
      Belinkov, Yonatan  and
      Hovy, Eduard},
    editor = {Merlo, Paola  and
      Tiedemann, Jorg  and
      Tsarfaty, Reut},
    booktitle = "Proceedings of the 16th Conference of the European Chapter of the Association for Computational Linguistics: Main Volume",
    month = apr,
    year = "2021",
    address = "Online",
    publisher = "Association for Computational Linguistics",
    url = "https://aclanthology.org/2021.eacl-main.295",
    doi = "10.18653/v1/2021.eacl-main.295",
    pages = "3363--3377",
}

@inproceedings{hewitt-liang-2019-designing,
    title = "Designing and Interpreting Probes with Control Tasks",
    author = "Hewitt, John  and
      Liang, Percy",
    editor = "Inui, Kentaro  and
      Jiang, Jing  and
      Ng, Vincent  and
      Wan, Xiaojun",
    booktitle = "Proceedings of the 2019 Conference on Empirical Methods in Natural Language Processing and the 9th International Joint Conference on Natural Language Processing (EMNLP-IJCNLP)",
    month = nov,
    year = "2019",
    address = "Hong Kong, China",
    publisher = "Association for Computational Linguistics",
    url = "https://aclanthology.org/D19-1275/",
    doi = "10.18653/v1/D19-1275",
    pages = "2733--2743"
}

@inproceedings{
kantamneni2025are,
title={Are Sparse Autoencoders Useful? A Case Study in Sparse Probing},
author={Subhash Kantamneni and Joshua Engels and Senthooran Rajamanoharan and Max Tegmark and Neel Nanda},
booktitle={Forty-second International Conference on Machine Learning},
year={2025},
url={https://openreview.net/forum?id=rNfzT8YkgO}
}

@inproceedings{exp_unders,
  author = {Teney, Damien and Peyrard, Maxime and Abbasnejad, Ehsan},
  title = {Predicting Is Not Understanding: Recognizing And Addressing Underspecification In Machine Learning},
  year = {2022},
  isbn = {978-3-031-20049-6},
  publisher = {Springer-Verlag},
  address = {Berlin, Heidelberg},
  url = {https://doi.org/10.1007/978-3-031-20050-2_27},
  doi = {10.1007/978-3-031-20050-2_27},
  booktitle = {Computer Vision – ECCV 2022: 17th European Conference, Tel Aviv, Israel, October 23–27, 2022, Proceedings, Part XXIII},
  pages = {458–476},
  numpages = {19},
  location = {Tel Aviv, Israel}
}

@article{elazar-etal-2021-amnesic,
    title = "Amnesic Probing: Behavioral Explanation with Amnesic Counterfactuals",
    author = "Elazar, Yanai  and
      Ravfogel, Shauli  and
      Jacovi, Alon  and
      Goldberg, Yoav",
    editor = "Roark, Brian  and
      Nenkova, Ani",
    journal = "Transactions of the Association for Computational Linguistics",
    volume = "9",
    year = "2021",
    address = "Cambridge, MA",
    publisher = "MIT Press",
    url = "https://aclanthology.org/2021.tacl-1.10/",
    doi = "10.1162/tacl_a_00359",
    pages = "160--175",
}

@inproceedings{lasri-etal-2022-probing,
    title = "Probing for the Usage of Grammatical Number",
    author = "Lasri, Karim  and
      Pimentel, Tiago  and
      Lenci, Alessandro  and
      Poibeau, Thierry  and
      Cotterell, Ryan",
    editor = "Muresan, Smaranda  and
      Nakov, Preslav  and
      Villavicencio, Aline",
    booktitle = "Proceedings of the 60th Annual Meeting of the Association for Computational Linguistics (Volume 1: Long Papers)",
    month = may,
    year = "2022",
    address = "Dublin, Ireland",
    publisher = "Association for Computational Linguistics",
    url = "https://aclanthology.org/2022.acl-long.603/",
    doi = "10.18653/v1/2022.acl-long.603",
    pages = "8818--8831",
}

@article{cobbe2021training,
  title={Training verifiers to solve math word problems},
  author={Cobbe, Karl and Kosaraju, Vineet and Bavarian, Mohammad and Chen, Mark and Jun, Heewoo and Kaiser, Lukasz and Plappert, Matthias and Tworek, Jerry and Hilton, Jacob and Nakano, Reiichiro and others},
  journal={arXiv preprint arXiv:2110.14168},
  year={2021},
  url={https://arxiv.org/abs/2110.14168}
}

@inproceedings{shen2025codi,
    title = "{CODI}: Compressing Chain-of-Thought into Continuous Space via Self-Distillation",
    author = "Shen, Zhenyi  and
      Yan, Hanqi  and
      Zhang, Linhai  and
      Hu, Zhanghao  and
      Du, Yali  and
      He, Yulan",
    editor = "Christodoulopoulos, Christos  and
      Chakraborty, Tanmoy  and
      Rose, Carolyn  and
      Peng, Violet",
    booktitle = "Proceedings of the 2025 Conference on Empirical Methods in Natural Language Processing",
    month = nov,
    year = "2025",
    address = "Suzhou, China",
    publisher = "Association for Computational Linguistics",
    url = "https://aclanthology.org/2025.emnlp-main.36/",
    doi = "10.18653/v1/2025.emnlp-main.36",
    pages = "677--693",
    ISBN = "979-8-89176-332-6"
}

@inproceedings{goyal2024think,
  title={Think before you speak: Training Language Models With Pause Tokens},
  author={Sachin Goyal and Ziwei Ji and Ankit Singh Rawat and Aditya Krishna Menon and Sanjiv Kumar and Vaishnavh Nagarajan},
  booktitle={The Twelfth International Conference on Learning Representations},
  year={2024},
  url={https://openreview.net/forum?id=ph04CRkPdC}
}

@article{radford2019language,
  title={{Language Models are Unsupervised Multitask Learners}},
  author={Radford, Alec and Wu, Jeff and Child, Rewon and Luan, David and Amodei, Dario and Sutskever, Ilya},
  journal={OpenAI blog},
  volume={1},
  number={8},
  pages={9},
  year={2019},
  url={https://cdn.openai.com/better-language-models/language_models_are_unsupervised_multitask_learners.pdf}
}

@inproceedings{
meloux2026the,
title={The Dead Salmons of {AI} Interpretability},
author={Maxime M{\'e}loux and Giada Dirupo and Fran{\c{c}}ois Portet and Maxime Peyrard},
booktitle={Mechanistic Interpretability Workshop at ICML 2026},
year={2026},
url={https://openreview.net/forum?id=6ijbFHrAbY}
}

@misc{deng2023implicitchainthoughtreasoning,
      title={Implicit Chain of Thought Reasoning via Knowledge Distillation}, 
      author={Yuntian Deng and Kiran Prasad and Roland Fernandez and Paul Smolensky and Vishrav Chaudhary and Stuart Shieber},
      year={2023},
      eprint={2311.01460},
      archivePrefix={arXiv},
      primaryClass={cs.CL},
      url={https://arxiv.org/abs/2311.01460}, 
}

@InProceedings{pmlr-v202-giannou23a,
  title = 	 {Looped Transformers as Programmable Computers},
  author =       {Giannou, Angeliki and Rajput, Shashank and Sohn, Jy-Yong and Lee, Kangwook and Lee, Jason D. and Papailiopoulos, Dimitris},
  booktitle = 	 {Proceedings of the 40th International Conference on Machine Learning},
  pages = 	 {11398--11442},
  year = 	 {2023},
  editor = 	 {Krause, Andreas and Brunskill, Emma and Cho, Kyunghyun and Engelhardt, Barbara and Sabato, Sivan and Scarlett, Jonathan},
  volume = 	 {202},
  series = 	 {Proceedings of Machine Learning Research},
  month = 	 {23--29 Jul},
  publisher =    {PMLR},
  url = 	 {https://proceedings.mlr.press/v202/giannou23a.html}
}

@inproceedings{
dehghani2018universal,
title={Universal Transformers},
author={Mostafa Dehghani and Stephan Gouws and Oriol Vinyals and Jakob Uszkoreit and Lukasz Kaiser},
booktitle={International Conference on Learning Representations},
year={2019},
url={https://openreview.net/forum?id=HyzdRiR9Y7},
}

@article{CODImechinterp25,
  title = {{Can we interpret latent reasoning using current mechanistic interpretability tools?}},
  author = {Cywinski, Bartosz and Bussmann, Bart and Conmy, Arthur and Engels, Joshua and Nanda, Neel and Rajamanoharan, Senthooran},
  journal={LessWrong. AI Aligment Forum},
  date = {2025-12-22},
  year = {2025},
  url={https://www.lesswrong.com/posts/YGAimivLxycZcqRFR/}
}

@inproceedings{
  scratchpadthinking,
  title={{Scratchpad Thinking: Alternation Between Storage and Computation in Latent Reasoning Models}},
  author={Brad Peters and Sayam Goyal and Mar{\'\i}a Emilia Granda and Akshath Vijayakumar Narmadha and Dharunish Yugeswardeenoo and Callum Stuart McDougall and Sean O'Brien and Ashwinee Panda and Kevin Zhu and Cole Blondin},
  booktitle={Mechanistic Interpretability Workshop at NeurIPS 2025},
  year={2025},
  url={https://openreview.net/forum?id=EV30qkZXrR}
}

@inproceedings{cui2026how,
    title={{How Do Latent Reasoning Methods Perform Under Weak and Strong Supervision?}},
    author={Yingqian Cui and Zhenwei Dai and Bing He and Zhan Shi and Hui Liu and Rui Sun and Zhiji Liu and Yue Xing and Jiliang Tang and Benoit Dumoulin},
    booktitle={Workshop on Latent {\&} Implicit Thinking {\textendash} Going Beyond CoT Reasoning},
    year={2026},
    url={https://openreview.net/forum?id=F2KA5IkONu}
}

@misc{zhang2025latenttokensthinkcausal,
      title={{Do Latent Tokens Think? A Causal and Adversarial Analysis of Chain-of-Continuous-Thought}}, 
      author={Yuyi Zhang and Boyu Tang and Tianjie Ju and Sufeng Duan and Gongshen Liu},
      year={2025},
      eprint={2512.21711},
      archivePrefix={arXiv},
      primaryClass={cs.CL},
      url={https://arxiv.org/abs/2512.21711}, 
}

@inproceedings{
rizvi-martel2026the,
title={The Illusion of Superposition? A Principled Analysis of Latent Thinking in Language Models},
author={Michael Rizvi-Martel and Guillaume Rabusseau and Marius Mosbach},
booktitle={Third Conference on Language Modeling},
year={2026},
url={https://openreview.net/forum?id=49kLxSRDzL}
}

@inproceedings{reasoningbysuperposition25,
  author = {Zhu, Hanlin and Hao, Shibo and Hu, Zhiting and Jiao, Jiantao and Russell, Stuart J and Tian, Yuandong},
  booktitle = {Advances in Neural Information Processing Systems},
  editor = {D. Belgrave and C. Zhang and H. Lin and R. Pascanu and P. Koniusz and M. Ghassemi and N. Chen},
  pages = {79931--79963},
  publisher = {Curran Associates, Inc.},
  title = {Reasoning by Superposition: A Theoretical Perspective on Chain of Continuous Thought},
  url = {https://proceedings.neurips.cc/paper_files/paper/2025/file/72c363c2a573ca2128bd176d3317696b-Paper-Conference.pdf},
  volume = {38},
  year = {2025}
}

@misc{liang2026latentcotmodelsthinkstepbystep,
      title={Do Latent-CoT Models Think Step-by-Step? A Mechanistic Study on Sequential Reasoning Tasks}, 
      author={Jia Liang and Liangming Pan},
      year={2026},
      eprint={2602.00449},
      archivePrefix={arXiv},
      primaryClass={cs.AI},
      url={https://arxiv.org/abs/2602.00449}, 
}

@article{wei2022chain,
  title={Chain-of-thought prompting elicits reasoning in large language models},
  author={Wei, Jason and Wang, Xuezhi and Schuurmans, Dale and Bosma, Maarten and Xia, Fei and Chi, Ed and Le, Quoc V and Zhou, Denny and others},
  journal={Advances in neural information processing systems},
  volume={35},
  pages={24824--24837},
  year={2022},
  url={https://proceedings.neurips.cc/paper_files/paper/2022/file/9d5609613524ecf4f15af0f7b31abca4-Paper-Conference.pdf}
}

@inproceedings{zhang2026soft,
 title = {{Soft Thinking: Unlocking the Reasoning Potential of LLMs in Continuous Concept Space}},
 author = {Zhang, Zhen and He, Xuehai and Yan, Weixiang and Shen, Ao and Zhao, Chenyang and Wang, Xin},
 booktitle = {Advances in Neural Information Processing Systems},
 doi = {10.52202/085713-5629},
 editor = {D. Belgrave and C. Zhang and H. Lin and R. Pascanu and P. Koniusz and M. Ghassemi and N. Chen},
 pages = {168990--169012},
 publisher = {Curran Associates, Inc.},
 url = {https://proceedings.neurips.cc/paper_files/paper/2025/file/f7396d1c54d51416958d63e285377103-Paper-Conference.pdf},
 volume = {38, Main Conference},
 year = {2025}
}

@article{chen2025reasoning,
  title={Reasoning models don't always say what they think},
  author={Chen, Yanda and Benton, Joe and Radhakrishnan, Ansh and Uesato, Jonathan and Denison, Carson and Schulman, John and Somani, Arushi and Hase, Peter and Wagner, Misha and Roger, Fabien and others},
  journal={arXiv preprint arXiv:2505.05410},
  year={2025},
  url={https://arxiv.org/abs/2505.05410}
}

@inproceedings{chan2023harms,
  title={Harms from increasingly agentic algorithmic systems},
  author={Chan, Alan and Salganik, Rebecca and Markelius, Alva and Pang, Chris and Rajkumar, Nitarshan and Krasheninnikov, Dmitrii and Langosco, Lauro and He, Zhonghao and Duan, Yawen and Carroll, Micah and others},
  booktitle={Proceedings of the 2023 ACM conference on fairness, accountability, and transparency},
  pages={651--666},
  year={2023},
  url={https://dl.acm.org/doi/10.1145/3593013.3594033}
}

@article{turpin2023language,
  title={Language models don't always say what they think: Unfaithful explanations in chain-of-thought prompting},
  author={Turpin, Miles and Michael, Julian and Perez, Ethan and Bowman, Samuel},
  journal={Advances in Neural Information Processing Systems},
  volume={36},
  pages={74952--74965},
  year={2023},
  url={https://proceedings.neurips.cc/paper_files/paper/2023/file/ed3fea9033a80fea1376299fa7863f4a-Paper-Conference.pdf}
}

@article{
bereska2024mechanistic,
title={Mechanistic Interpretability for {AI} Safety - A Review},
author={Leonard Bereska and Stratis Gavves},
journal={Transactions on Machine Learning Research},
issn={2835-8856},
year={2024},
url={https://openreview.net/forum?id=ePUVetPKu6},
note={Survey Certification, Expert Certification}
}

@inproceedings{wei2025sim,
  title={{SIM-CoT: Supervised Implicit Chain-of-Thought}},
  author={Xilin Wei and Xiaoran Liu and Yuhang Zang and Xiaoyi Dong and Yuhang Cao and Jiaqi Wang and Xipeng Qiu and Dahua Lin},
  booktitle={The Fourteenth International Conference on Learning Representations},
  year={2026},
  url={https://openreview.net/forum?id=6YRJ4jmVQl}
}

@inproceedings{LatentReasoningInterpretable2026,
  title={{Are Latent Reasoning Models Easily Interpretable?}},
  author={Connor Dilgren and Sarah Wiegreffe},
  booktitle={Workshop on Latent {\&} Implicit Thinking {\textendash} Going Beyond CoT Reasoning},
  year={2026},
  url={https://openreview.net/forum?id=L4k8rbmwrr}
}

@inproceedings{MarkovCoTHelp2026,
  title={{When does Chain-of-Thought Help: A Markovian Perspective}},
  author={Zihan Wang and Yijun Dong and Qi Lei},
  booktitle={Workshop on Latent {\&} Implicit Thinking {\textendash} Going Beyond CoT Reasoning},
  year={2026},
  url={https://openreview.net/forum?id=fz5BC8VJ6X}
}

@article{elhage2022toy,
  title={Toy models of superposition},
  author={Elhage, Nelson and Hume, Tristan and Olsson, Catherine and Schiefer, Nicholas and Henighan, Tom and Kravec, Shauna and Hatfield-Dodds, Zac and Lasenby, Robert and Drain, Dawn and Chen, Carol and others},
  journal={arXiv preprint arXiv:2209.10652},
  year={2022},
  url={https://arxiv.org/abs/2209.10652}
}

@inproceedings{
bernas2026revisiting,
title={Revisiting Anisotropy in Language Transformers: The Geometry of Learning Dynamics},
author={Raphael Bernas and Fanny Jourdan and Antonin Poch{\'e} and Celine Hudelot},
booktitle={Forty-third International Conference on Machine Learning},
year={2026},
url={https://openreview.net/forum?id=BxSubxbbNc}
}

@article{gur2018gradient,
  title={Gradient descent happens in a tiny subspace},
  author={Gur-Ari, Guy and Roberts, Daniel A and Dyer, Ethan},
  journal={arXiv preprint arXiv:1812.04754},
  year={2018},
  url={https://arxiv.org/abs/1812.04754}
}

@inproceedings{
park2024the,
title={The Linear Representation Hypothesis and the Geometry of Large Language Models},
author={Kiho Park and Yo Joong Choe and Victor Veitch},
booktitle={Forty-first International Conference on Machine Learning},
year={2024},
url={https://openreview.net/forum?id=UGpGkLzwpP}
}

@inproceedings{
gurnee2024language,
title={Language Models Represent Space and Time},
author={Wes Gurnee and Max Tegmark},
booktitle={The Twelfth International Conference on Learning Representations},
year={2024},
url={https://openreview.net/forum?id=jE8xbmvFin}
}

@inproceedings{geva2023dissecting,
    title = {{Dissecting Recall of Factual Associations in Auto-Regressive Language Models}},
    author = "Geva, Mor  and
      Bastings, Jasmijn  and
      Filippova, Katja  and
      Globerson, Amir",
    editor = "Bouamor, Houda  and
      Pino, Juan  and
      Bali, Kalika",
    booktitle = "Proceedings of the 2023 Conference on Empirical Methods in Natural Language Processing",
    month = dec,
    year = "2023",
    address = "Singapore",
    publisher = "Association for Computational Linguistics",
    url = "https://aclanthology.org/2023.emnlp-main.751/",
    doi = "10.18653/v1/2023.emnlp-main.751",
    pages = "12216--12235"
}

@inproceedings{li2026thinking,
 author = {Li, Xiaomin and Yu, Zhou and Zhang, Zhiwei and Chen, Xupeng and Zhang, Ziji and Zhuang, Yingying and Sadagopan, Narayanan and Beniwal, Anurag},
 booktitle = {Advances in Neural Information Processing Systems},
 doi = {10.52202/085713-2612},
 editor = {D. Belgrave and C. Zhang and H. Lin and R. Pascanu and P. Koniusz and M. Ghassemi and N. Chen},
 pages = {77925--77962},
 publisher = {Curran Associates, Inc.},
 title = {{When Thinking Fails: The Pitfalls of Reasoning for Instruction-Following in LLMs}},
 url = {https://proceedings.neurips.cc/paper_files/paper/2025/file/706338a08f9378b708f21cbf5686e617-Paper-Conference.pdf},
 volume = {38, Main Conference},
 year = {2025}
}

@inproceedings{lyu2023faithful,
    title = {{Faithful Chain-of-Thought Reasoning}},
    author = "Lyu, Qing  and
      Havaldar, Shreya  and
      Stein, Adam  and
      Zhang, Li  and
      Rao, Delip  and
      Wong, Eric  and
      Apidianaki, Marianna  and
      Callison-Burch, Chris",
    editor = "Park, Jong C.  and
      Arase, Yuki  and
      Hu, Baotian  and
      Lu, Wei  and
      Wijaya, Derry  and
      Purwarianti, Ayu  and
      Krisnadhi, Adila Alfa",
    booktitle = "Proceedings of the 13th International Joint Conference on Natural Language Processing and the 3rd Conference of the Asia-Pacific Chapter of the Association for Computational Linguistics (Volume 1: Long Papers)",
    month = nov,
    year = "2023",
    address = "Nusa Dua, Bali",
    publisher = "Association for Computational Linguistics",
    url = "https://aclanthology.org/2023.ijcnlp-main.20/",
    doi = "10.18653/v1/2023.ijcnlp-main.20",
    pages = "305--329"
}

@misc{korbak2025cotmonitorability,
      title={{Chain of Thought Monitorability: A New and Fragile Opportunity for AI Safety}}, 
      author={Tomek Korbak and Mikita Balesni and Elizabeth Barnes and Yoshua Bengio and Joe Benton and Joseph Bloom and Mark Chen and Alan Cooney and Allan Dafoe and Anca Dragan and Scott Emmons and Owain Evans and David Farhi and Ryan Greenblatt and Dan Hendrycks and Marius Hobbhahn and Evan Hubinger and Geoffrey Irving and Erik Jenner and Daniel Kokotajlo and Victoria Krakovna and Shane Legg and David Lindner and David Luan and Aleksander Mądry and Julian Michael and Neel Nanda and Dave Orr and Jakub Pachocki and Ethan Perez and Mary Phuong and Fabien Roger and Joshua Saxe and Buck Shlegeris and Martín Soto and Eric Steinberger and Jasmine Wang and Wojciech Zaremba and Bowen Baker and Rohin Shah and Vlad Mikulik},
      year={2025},
      eprint={2507.11473},
      archivePrefix={arXiv},
      primaryClass={cs.AI},
      url={https://arxiv.org/abs/2507.11473}
}

@article{geiger2025causal,
  author  = {Atticus Geiger and Duligur Ibeling and Amir Zur and Maheep Chaudhary and Sonakshi Chauhan and Jing Huang and Aryaman Arora and Zhengxuan Wu and Noah Goodman and Christopher Potts and Thomas Icard},
  title   = {{Causal Abstraction: A Theoretical Foundation for Mechanistic Interpretability}},
  journal = {Journal of Machine Learning Research},
  year    = {2025},
  volume  = {26},
  number  = {83},
  pages   = {1--64},
  url     = {http://jmlr.org/papers/v26/23-0058.html}
}

@article{belinkov2022probing,
    title = "Probing Classifiers: Promises, Shortcomings, and Advances",
    author = "Belinkov, Yonatan",
    journal = "Computational Linguistics",
    volume = "48",
    number = "1",
    month = mar,
    year = "2022",
    address = "Cambridge, MA",
    publisher = "MIT Press",
    url = "https://aclanthology.org/2022.cl-1.7/",
    doi = "10.1162/coli_a_00422",
    pages = "207--219"
}

@inproceedings{jain2019attention,
    title = "{A}ttention is not {E}xplanation",
    author = "Jain, Sarthak  and
      Wallace, Byron C.",
    editor = "Burstein, Jill  and
      Doran, Christy  and
      Solorio, Thamar",
    booktitle = "Proceedings of the 2019 Conference of the North {A}merican Chapter of the Association for Computational Linguistics: Human Language Technologies, Volume 1 (Long and Short Papers)",
    month = jun,
    year = "2019",
    address = "Minneapolis, Minnesota",
    publisher = "Association for Computational Linguistics",
    url = "https://aclanthology.org/N19-1357/",
    doi = "10.18653/v1/N19-1357",
    pages = "3543--3556"
}

@inproceedings{
  wang2023interpretability,
  title={Interpretability in the Wild: a Circuit for Indirect Object Identification in {GPT}-2 Small},
  author={Kevin Ro Wang and Alexandre Variengien and Arthur Conmy and Buck Shlegeris and Jacob Steinhardt},
  booktitle={The Eleventh International Conference on Learning Representations },
  year={2023},
  url={https://openreview.net/forum?id=NpsVSN6o4ul}
}

@article{chan2022causal, 
	title={Causal scrubbing, a method for rigorously testing interpretability hypotheses},
	author={Chan, Lawrence and Garriga-Alonso, Adrià and Goldwosky-Dill, Nicholas and Greenblatt, Ryan and Nitishinskaya, Jenny and Radhakrishnan, Ansh and Shlegeris, Buck and Thomas, Nate},
	year={2022},
	journal={AI Alignment Forum},
	url={https://www.alignmentforum.org/posts/JvZhhzycHu2Yd57RN}
}

@misc{heimersheim2024activation,
      title={How to use and interpret activation patching}, 
      author={Stefan Heimersheim and Neel Nanda},
      year={2024},
      eprint={2404.15255},
      archivePrefix={arXiv},
      primaryClass={cs.LG},
      url={https://arxiv.org/abs/2404.15255}, 
}

@inproceedings{meloux2025everything,
  title={Everything, Everywhere, All at Once: Is Mechanistic Interpretability Identifiable?},
  author={Maxime M{\'e}loux and Silviu Maniu and Fran{\c{c}}ois Portet and Maxime Peyrard},
  booktitle={The Thirteenth International Conference on Learning Representations},
  year={2025},
  url={https://openreview.net/forum?id=5IWJBStfU7}
}

@inproceedings{
arcuschin2026chainofthought,
title={Chain-of-Thought Reasoning In The Wild Is Not Always Faithful},
author={Iv{\'a}n Arcuschin and Jett Janiak and Robert Krzyzanowski and Senthooran Rajamanoharan and Neel Nanda and Arthur Conmy},
booktitle={Forty-third International Conference on Machine Learning},
year={2026},
url={https://openreview.net/forum?id=NUyt4uxzx0}
}

@inproceedings{
boppana2026reasoning,
title={Reasoning Theater: Disentangling Model Beliefs from Chain-of-Thought},
author={Siddharth Boppana and Annabel Ma and Max Loeffler and Rapha{\"e}l Sarfati and Eric Bigelow and Atticus Geiger and Owen Lewis and Jack Merullo},
booktitle={Forty-third International Conference on Machine Learning},
year={2026},
url={https://openreview.net/forum?id=1lyagkzogH}
}

@article{zhao2025can,
  title={Can Aha Moments Be Fake? Identifying True and Decorative Thinking Steps in Chain-of-Thought},
  author={Zhao, Jiachen and Sun, Yiyou and Shi, Weiyan and Song, Dawn},
  journal={arXiv preprint arXiv:2510.24941},
  year={2025},
  url={https://arxiv.org/abs/2510.24941}
}

@article{meng2022locating,
  title={Locating and editing factual associations in gpt},
  author={Meng, Kevin and Bau, David and Andonian, Alex and Belinkov, Yonatan},
  journal={Advances in neural information processing systems},
  volume={35},
  pages={17359--17372},
  year={2022},
  url={https://proceedings.neurips.cc/paper_files/paper/2022/file/6f1d43d5a82a37e89b0665b33bf3a182-Paper-Conference.pdf}
}

@inproceedings{lan2024towards,
    title = {{Towards Interpretable Sequence Continuation: Analyzing Shared Circuits in Large Language Models}},
    author = "Lan, Michael  and
      Torr, Philip  and
      Barez, Fazl",
    editor = "Al-Onaizan, Yaser  and
      Bansal, Mohit  and
      Chen, Yun-Nung",
    booktitle = "Proceedings of the 2024 Conference on Empirical Methods in Natural Language Processing",
    month = nov,
    year = "2024",
    address = "Miami, Florida, USA",
    publisher = "Association for Computational Linguistics",
    url = "https://aclanthology.org/2024.emnlp-main.699/",
    doi = "10.18653/v1/2024.emnlp-main.699",
    pages = "12576--12601"
}

@misc{mcgrath2023hydraeffectemergentselfrepair,
      title={The Hydra Effect: Emergent Self-repair in Language Model Computations}, 
      author={Thomas McGrath and Matthew Rahtz and Janos Kramar and Vladimir Mikulik and Shane Legg},
      year={2023},
      eprint={2307.15771},
      archivePrefix={arXiv},
      primaryClass={cs.LG},
      url={https://arxiv.org/abs/2307.15771}, 
}

@misc{bhatia2026reallycontrolstemporalreasoning,
      title={What Really Controls Temporal Reasoning in Large Language Models: Tokenisation or Representation of Time?}, 
      author={Gagan Bhatia and Ahmad Muhammad Isa and Maxime Peyrard and Wei Zhao},
      year={2026},
      eprint={2603.19017},
      archivePrefix={arXiv},
      primaryClass={cs.CL},
      url={https://arxiv.org/abs/2603.19017}, 
}

@inproceedings{bhatia-etal-2025-date,
    title = "Date Fragments: A Hidden Bottleneck of Tokenization for Temporal Reasoning",
    author = "Bhatia, Gagan  and
      Peyrard, Maxime  and
      Zhao, Wei",
    editor = "Christodoulopoulos, Christos  and
      Chakraborty, Tanmoy  and
      Rose, Carolyn  and
      Peng, Violet",
    booktitle = "Proceedings of the 2025 Conference on Empirical Methods in Natural Language Processing",
    month = nov,
    year = "2025",
    address = "Suzhou, China",
    publisher = "Association for Computational Linguistics",
    url = "https://aclanthology.org/2025.emnlp-main.159/",
    doi = "10.18653/v1/2025.emnlp-main.159",
    pages = "3201--3219",
    ISBN = "979-8-89176-332-6"
}

@inproceedings{monea-etal-2024-glitch,
    title = "A Glitch in the Matrix? Locating and Detecting Language Model Grounding with Fakepedia",
    author = "Monea, Giovanni  and
      Peyrard, Maxime  and
      Josifoski, Martin  and
      Chaudhary, Vishrav  and
      Eisner, Jason  and
      Kiciman, Emre  and
      Palangi, Hamid  and
      Patra, Barun  and
      West, Robert",
    editor = "Ku, Lun-Wei  and
      Martins, Andre  and
      Srikumar, Vivek",
    booktitle = "Proceedings of the 62nd Annual Meeting of the Association for Computational Linguistics (Volume 1: Long Papers)",
    month = aug,
    year = "2024",
    address = "Bangkok, Thailand",
    publisher = "Association for Computational Linguistics",
    url = "https://aclanthology.org/2024.acl-long.369/",
    doi = "10.18653/v1/2024.acl-long.369",
    pages = "6828--6844"
}

@inproceedings{
    zhou2026the,
    title={The Geometry of Reasoning: Flowing Logics in Representation Space},
    author={Yufa Zhou and Yixiao Wang and Xunjian Yin and Shuyan Zhou and Anru Zhang},
    booktitle={The Fourteenth International Conference on Learning Representations},
    year={2026},
    url={https://openreview.net/forum?id=ixr5Pcabq7}
}

@misc{meta_llama32,
  author       = {{Meta AI}},
  title        = {{Llama 3.2: Revolutionizing edge AI and vision with open, customizable models}},
  year         = {2024},
  month        = sep,
  url      = {https://ai.meta.com/blog/llama-3-2-connect-2024-vision-edge-mobile-devices/},
  note         = {Accessed: 2026-02-16}
}

@misc{llama3herdmodels2024,
      title={{The Llama 3 Herd of Models}}, 
      author={Grattafiori, Aaron and Dubey, Abhimanyu and Jauhri, Abhinav and Pandey, Abhinav and Kadian, Abhishek and Al-Dahle, Ahmad and Letman, Aiesha and Mathur, Akhil and Schelten, Alan and Vaughan, Alex and others},
      year={2024},
      eprint={2407.21783},
      archivePrefix={arXiv},
      primaryClass={cs.AI},
      url={https://arxiv.org/abs/2407.21783}, 
}

@article{elhage2021mathematical,
  title={A mathematical framework for transformer circuits},
  author={Elhage, Nelson and Nanda, Neel and Olsson, Catherine and Henighan, Tom and Joseph, Nicholas and Mann, Ben and Askell, Amanda and Bai, Yuntao and Chen, Anna and Conerly, Tom and others},
  journal={Transformer Circuits Thread},
  volume={1},
  number={1},
  pages={12},
  year={2021},
  url={https://transformer-circuits.pub/2021/framework/index.html}
}

@article{gurnee2026GlobalWorkspaceJacobian,
  title = {Verbalizable {{Representations Form}} a {{Global Workspace}} in {{Language Models}}},
  author = {Gurnee, Wes and Sofroniew, Nicholas and Pearce, Adam and Piotrowski, Mateusz and Kauvar, Isaac and Chen, Runjin and Soligo, Anna and Bogdan, Paul and Ong, Euan and Wang, Rowan and Thompson, Ben and Abrahams, David and Kantamneni, Subhash and Ameisen, Emmanuel and Batson, Joshua and Lindsey, Jack},
  year = 2026,
  month = jul,
  journal = {Transformer Circuits Thread},
  eprint = {2607.15495},
  primaryclass = {cs.CL},
  publisher = {arXiv},
  doi = {10.48550/arXiv.2607.15495},
  urldate = {2026-07-27},
  archiveprefix = {arXiv},
}

@article{SurfacingModelAlgorithms,
  title        = {Towards Surfacing Model Algorithms with Meta-Tokens in the {J-Space}},
  author       = {Bhatia, Agam and Blank, Camila and Nanda, Neel},
  year         = {2026},
  month        = jul,
  urldate      = {2026-07-27},
  url = {https://www.lesswrong.com/posts/6ek6n7yZ5DzfarJHy},
  journal =  {LessWrong. AI Alignment Forum},
}

@misc{wu2026jcotchainofthoughtjspace,
      title={J-CoT: Chain-of-Thought in J-Space}, 
      author={Junde Wu and Jiayuan Zhu and Fengling Liu and Minhao Hu and Jiazhen Pan},
      year={2026},
      eprint={2607.21981},
      archivePrefix={arXiv},
      primaryClass={cs.CL},
      url={https://arxiv.org/abs/2607.21981}, 
}

@inproceedings{grant2026addressing,
  title={Addressing divergent representations from causal interventions on neural networks},
  author={Grant, Satchel and Han, Simon Jerome and Tartaglini, Alexa and Potts, Christopher},
  booktitle={International Conference on Learning Representations},
  volume={2026},
  pages={11330--11356},
  year={2026},
  url={https://openreview.net/forum?id=cZrTMqYVL6}
}

@inproceedings{unlockingblackboxlatent2026,
    title = {{Unlocking the Black Box of Latent Reasoning: An Interpretability-Guided Approach to Intervention}},
    author = "Chang, Shuochen  and
      Bai, Tong  and
      Zhang, Xiaofeng  and
      Ma, Qianli  and
      Liu, Qingyang  and
      Liao, Zhaohe  and
      Miao, Yibo  and
      Niu, Li",
    editor = "Liakata, Maria  and
      Moreira, Viviane P.  and
      Zhang, Jiajun  and
      Jurgens, David",
    booktitle = "Proceedings of the 64th Annual Meeting of the {A}ssociation for {C}omputational {L}inguistics (Volume 1: Long Papers)",
    month = jul,
    year = "2026",
    address = "San Diego, California, United States",
    publisher = "Association for Computational Linguistics",
    url = "https://aclanthology.org/2026.acl-long.1568/",
    doi = "10.18653/v1/2026.acl-long.1568",
    pages = "34019--34032",
    ISBN = "979-8-89176-390-6"
}

@inproceedings{li2026dynamics,
    title={{Dynamics Within Latent Chain-of-Thought: An Empirical Study of Causal Structure}},
    author={Zirui Li and Xuefeng Bai and Kehai Chen and Yizhi LI and Jian Yang and Chenghua Lin and Min Zhang},
    booktitle={Forty-third International Conference on Machine Learning},
    year={2026},
    url={https://openreview.net/forum?id=kHB8m3ojGe}
}

@misc{seddik2026formalizinglatentthoughtsaxioms,
      title={Formalizing Latent Thoughts: Four Axioms of Thought Representation in LLMs}, 
      author={Fahd Seddik and Fatemeh Fard},
      year={2026},
      eprint={2606.27378},
      archivePrefix={arXiv},
      primaryClass={cs.CL},
      url={https://arxiv.org/abs/2606.27378}, 
}

@misc{kumar2026geometriclatentreasoninginduces,
      title={Geometric Latent Reasoning Induces Shorter Generations in LLMs}, 
      author={Shashi Kumar and Yacouba Kaloga and Petr Motlicek and Ina Kodrasi and Andrea Cavallaro},
      year={2026},
      eprint={2606.02248},
      archivePrefix={arXiv},
      primaryClass={cs.CL},
      url={https://arxiv.org/abs/2606.02248}, 
}

@misc{solgi2026lorilowrankdistillationimplicit,
      title={{LoRi: Low-Rank Distillation for Implicit Reasoning}}, 
      author={Ryan Solgi and Jiayi Tian and Zheng Zhang},
      year={2026},
      eprint={2606.05315},
      archivePrefix={arXiv},
      primaryClass={cs.CL},
      url={https://arxiv.org/abs/2606.05315}, 
}
\appendix

\clearpage
\twocolumn[
\noindent\begin{minipage}{\textwidth}
\centering
\small
\setlength{\tabcolsep}{5pt}
\begin{tabular}{lll}
\toprule
\textbf{Model} &
\textbf{GPT-2} &
\textbf{Llama-3.2-1B-Instruct} \\
\midrule
\multicolumn{3}{l}{\textbf{\qquad GSM8K-Aug}} \\
CoT &
\href{https://huggingface.co/darpanaswal/cot-gpt2-gsm}
{\texttt{darpanaswal/cot-gpt2-gsm}} &
\href{https://huggingface.co/thomas-ferraz/CoT-Llama3.2-1B-GSM8kAug}
{\texttt{thomas-ferraz/CoT-Llama3.2-1B-GSM8kAug}} \\
PaT &
\href{https://huggingface.co/darpanaswal/pause-gpt2-gsm}
{\texttt{darpanaswal/pause-gpt2-gsm}} &
\href{https://huggingface.co/thomas-ferraz/Pause-Llama3.2-1B-GSM8kAug}
{\texttt{thomas-ferraz/Pause-Llama3.2-1B-GSM8kAug}} \\
Coconut &
\href{https://huggingface.co/darpanaswal/coconut-gpt2-gsm}
{\texttt{darpanaswal/coconut-gpt2-gsm}} &
\href{https://huggingface.co/thomas-ferraz/Coconut-Llama3.2-1B-GSM8kAug}
{\texttt{thomas-ferraz/Coconut-Llama3.2-1B-GSM8kAug}} \\
Coconut$_u$ &
\href{darpanaswal/coconut_u0.3-gpt2-gsm}
{\texttt{darpanaswal/coconut\_u0.3-gpt2-gsm}} &
\href{https://huggingface.co/thomas-ferraz/Coconut-u0.3-Llama3.2-1B-GSM8kAug}
{\texttt{thomas-ferraz/Coconut-u0.3-Llama3.2-1B-GSM8kAug}} \\
CODI &
\href{https://huggingface.co/darpanaswal/codi-gpt2-gsm}
{\texttt{darpanaswal/codi-gpt2-gsm}} &
\href{https://huggingface.co/thomas-ferraz/CODI-Llama3.2-1B-GSM8kAug}
{\texttt{thomas-ferraz/CODI-Llama3.2-1B-GSM8kAug}} \\
\midrule
\multicolumn{3}{l}{\textbf{\qquad ProsQA}} \\
CoT &
\href{https://huggingface.co/darpanaswal/cot-gpt2-prosqa}
{\texttt{darpanaswal/cot-gpt2-prosqa}} &
\href{https://huggingface.co/thomas-ferraz/CoT-Llama3.2-1B-ProsQA}
{\texttt{thomas-ferraz/CoT-Llama3.2-1B-ProsQA}} \\
PaT &
\href{https://huggingface.co/darpanaswal/pause-gpt2-prosqa}
{\texttt{darpanaswal/pause-gpt2-prosqa}} &
\href{https://huggingface.co/thomas-ferraz/Pause-Llama3.2-1B-ProsQA}
{\texttt{thomas-ferraz/Pause-Llama3.2-1B-ProsQA}} \\
Coconut &
\href{https://huggingface.co/darpanaswal/coconut-gpt2-prosqa}
{\texttt{darpanaswal/coconut-gpt2-prosqa}} &
\href{https://huggingface.co/thomas-ferraz/Coconut-Llama3.2-1B-ProsQA}
{\texttt{thomas-ferraz/Coconut-Llama3.2-1B-ProsQA}} \\
Coconut$_u$ &
\href{https://huggingface.co/darpanaswal/coconut_u0.3-gpt2-prosqa}
{\texttt{darpanaswal/coconut\_u0.3-gpt2-prosqa}} &
\href{https://huggingface.co/thomas-ferraz/Coconut-u0.3-Llama3.2-1B-ProsQA}
{\texttt{thomas-ferraz/Coconut-u0.3-Llama3.2-1B-ProsQA}} \\
CODI &
\href{https://huggingface.co/darpanaswal/codi-gpt2-prosqa}
{\texttt{darpanaswal/codi-gpt2-prosqa}} &
\href{https://huggingface.co/thomas-ferraz/CODI-Llama3.2-1B-ProsQA}
{\texttt{thomas-ferraz/CODI-Llama3.2-1B-ProsQA}} \\
\bottomrule
\end{tabular}

\captionof{table}{Model checkpoints trained for this work and publicly released.
We release all model variants for both GPT-2 and
Llama-3.2-1B-Instruct backbones on GSM8K-Aug and ProsQA.}
\label{tab:released_checkpoints}
\end{minipage}

\vspace{0.75em}
]

\section{Reproducibility Statement}
To facilitate full reproducibility, we release the source code required to reproduce our training and analyses. The main analysis code is available ~\href{https://github.com/darpanaswal/InterpretingLatents}{here}. Training code for the Coconut family and its controls and variants is available~\href{https://github.com/darpanaswal/coconut-pause-enabled}{here}, while the CODI training implementation and variants are available~\href{https://github.com/darpanaswal/codi-prosqa}{here}. We additionally release all trained model checkpoints used in our experiments, covering the model variants considered across GPT-2 and Llama-3.2-1B-Instruct on GSM8K-Aug and ProsQA, and including intermediate training checkpoints, at the Hugging Face collection~\href{https://huggingface.co/collections/darpanaswal/latent-reasoning-models}{here}; direct links to each checkpoint are provided in Table~\ref{tab:released_checkpoints}.

Dataset sources, splits, and statistics are documented in Appendix~\ref{app:dataset_motivation}, while model and control definitions, training procedures, and implementation details are provided in Appendix~\ref{app:model_roles}--\ref{app:training_details}, including the complete training configurations and hyperparameters in Tables~\ref{tab:baseline_training_hyperparameters}--\ref{tab:codi_training_hyperparameters}. The analysis procedures are described in \S~\ref{sec:epiphenomena}--\S~\ref{sec:geometry}, with the Llama-3.2 replication and additional analyses reported in Appendices~\ref{app:llama_results}--\ref{app:extended_analyses}. Statistical evaluation procedures and full significance results are provided in \S~\ref{sec:experimental_setup} and Appendix~\ref{app:statistical_tests}, including bootstrap confidence intervals, paired significance tests, and multiple-testing correction. Together, these resources provide a reproducible foundation for verifying our results and extending the causal and geometric analysis of latent reasoning models.


\begin{figure*}[!ht]
    \centering
    \begin{subfigure}[t]{\linewidth}
        \centering
        \includegraphics[width=\linewidth]{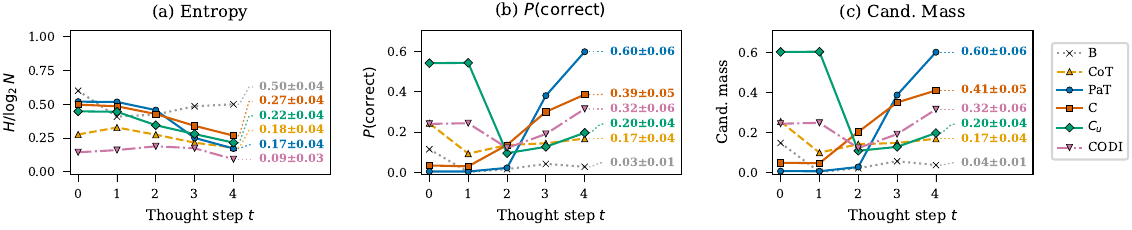}
        \caption{Case Study 1: Superposition and BFS-like search on Graph-Hopping}
        \label{fig:epiphenomena_llama_superposition}
    \end{subfigure}
    
    \vspace{0.5em}
    
    \begin{subfigure}[t]{\linewidth}
        \centering
        \includegraphics[width=\linewidth]{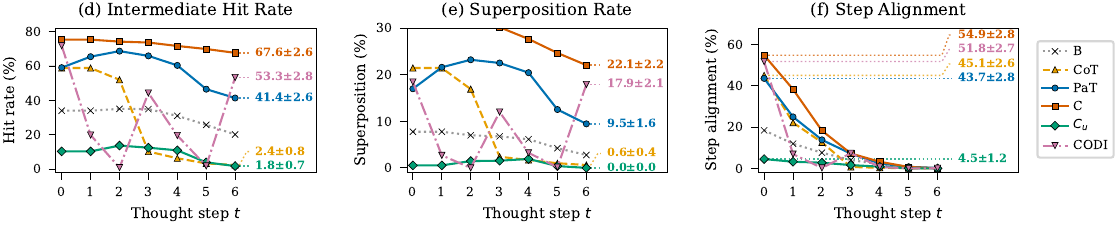}
        \caption{Case Study 2: Scratchpad Thinking on Arithmetic-Reasoning via Logit-Lens}
        \label{fig:epiphenomena_llama_scratchpad}
    \end{subfigure}
    
    \caption{Probing for breadth-first-search (BFS) patterns on graph-hopping (top) and scratchpad-thinking on arithmetic-reasoning (bottom) on the Llama-3.2 model family (\S~\ref{sec:epiphenomena}). \textbf{\textit{Takeaway:}} Matched controls reproduce or invert these observable patterns, showing they are not specific to the proposed recurrence or curriculum mechanisms and, thus, are insufficient alone for mechanistic attribution.}
    \label{fig:epiphenomena_llama}
\end{figure*}%

\section{Extension of Main Results to Llama-3.2-1B-Instruct}
\label{app:llama_results}
In this section, we extend the experiments in \S~\ref{sec:epiphenomena}, \S~\ref{sec:thought_use} and \S~\ref{sec:geometry} to Llama-3.2-1B-Instruct~\footnote{Available at the HuggingFace's Hub at: \url{https://huggingface.co/meta-llama/Llama-3.2-1B-Instruct}.}~\citep{meta_llama32,llama3herdmodels2024} to analyze how latent training impacts larger models.

\paragraph{Epiphenomenal Patterns in LRM Interpretability (\S~\ref{sec:epiphenomena}).}
Figure~\ref{fig:epiphenomena_llama_superposition} reports the results for case study 1 (superposition and BFS-like search on graph-hopping). Like in the GPT-2 case, we find PaT replicating the superposition and BFS dynamics of C, arguably better than C itself. C$_u$ does not create the flipped profile with Llama as it did in the GPT-2 case, but we see its confidence peak at an intermediate step instead of the final step. B, CoT and CODI show similar profiles as in their GPT-2 counterpart. Overall, the observable patterns of superposition and BFS remain uninformative of where they truly emerge and whether they impact model performance in any way.

Figure~\ref{fig:epiphenomena_llama_scratchpad} reports the results for case study 2 (scratchpad reasoning on arithmetic-reasoning via logit-lens). CODI's own strict alternation pattern disappears, and no other model recreates anything resembling it. C$_u$ completely collapses, showing flat profiles across metrics under logit-lens decodability (even below control B). This provides further evidence for CODI's scratchpad thinking mechanism being attributable more to the model-family than to the latent-reasoning recipe itself.

\paragraph{When and How do LRMs use Latent Thoughts (\S~\ref{sec:thought_use})?}

\begin{table}[!ht]
\centering
\small
\setlength{\tabcolsep}{3pt}
\begin{tabular}{lcc|cc}
\toprule
Model & \makecell{Graph- \\ Hopping} & \makecell{Thoughts \\ Removed} & \makecell{Arithmetic- \\ Reasoning} & \makecell{Thoughts \\ Removed} \\
\midrule
B & $2.2 {\scriptstyle \pm 1.3}$ & -- & $7.7 {\scriptstyle \pm 1.5}$ & -- \\
CoT & $96.8 {\scriptstyle \pm 1.5}$ & -- & $53.3 {\scriptstyle \pm 2.6}$ & -- \\
PaT & $99.8 {\scriptstyle \pm 0.3}$ & $99.8 {\scriptstyle \pm 0.3}$ & $26.8 {\scriptstyle \pm 2.4}$ & $3.9 {\scriptstyle \pm 1.0}$ \\
C & $99.6 {\scriptstyle \pm 0.5}$ & $99.6 {\scriptstyle \pm 0.5}$ & $33.2 {\scriptstyle \pm 2.4}$ & $12.1 {\scriptstyle \pm 1.7}$ \\
C$_u$ & $99.8 {\scriptstyle \pm 0.3}$ & $99.8 {\scriptstyle \pm 0.3}$ & $51.7 {\scriptstyle \pm 2.6}$ & $49.4 {\scriptstyle \pm 2.7}$ \\
CODI & $99.8 {\scriptstyle \pm 0.3}$ & $99.8 {\scriptstyle \pm 0.3}$ & $44.6 {\scriptstyle \pm 2.6}$ & $23.5 {\scriptstyle \pm 2.3}$ \\
\bottomrule
\end{tabular}
\caption{Accuracy under thought-ablation at test-time on the Llama-3.2 model family (\S~\ref{sec:thought_use}). Performance degrades on arithmetic-reasoning; graph-hopping is unaffected. \textbf{\textit{Takeaway:}} Influence of latent thoughts on model performance is task-varying.}
\label{tab:ablation_main_llama}
\end{table}

Table~\ref{tab:ablation_main_llama} reports the results for the latent thought ablation at test-time experiment. Results are comparable to the GPT-2 family except for C$_u$: Unlike C$_u$ GPT-2 (Table~\ref{tab:interventions}), it shows (i) negligible thought necessity in graph-hopping, and (ii) a minor drop (instead of improvement) in arithmetic-reasoning under thought ablation. This shows that latent thought necessity varies not just across tasks but also model families. Additionally, C$_u$'s flat hit-rate on arithmetic-reasoning despite clear (albeit small) thought necessity further demonstrates how observable patterns fall short of explaining true model behavior.

\begin{figure*}[!t]
  \centering
  \includegraphics[width=\textwidth]{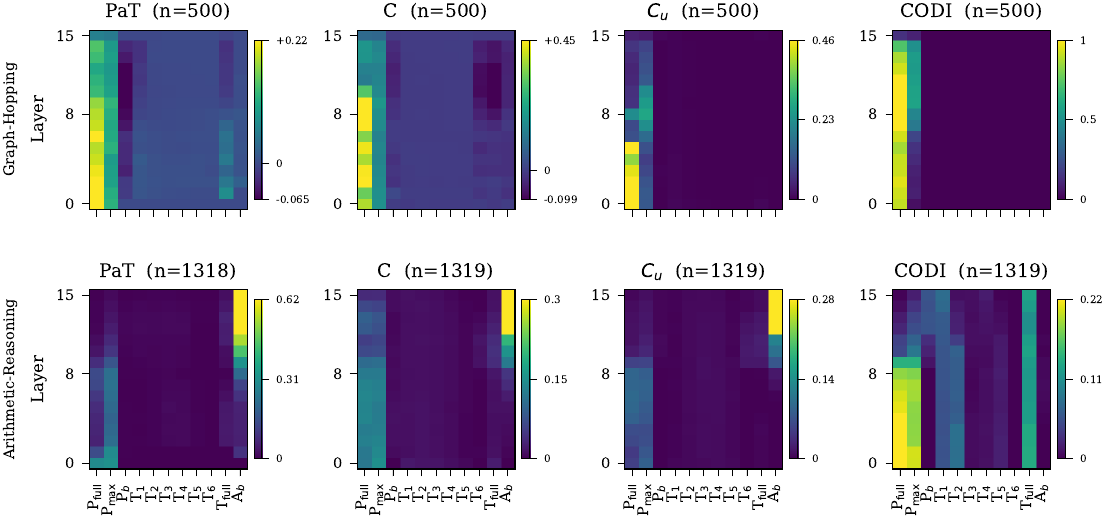}
  \caption{Per-layer $\mathrm{IE}_{\mathrm{KL}}$ on the residual stream under partner-prompt corruption across buckets $\{P_{\text{full}}, P_{\max}, P_b, T_1,\dots,T_K, T_{\text{full}}, A_b\}$ on the Llama-3.2 model family. Prompt positions recover best across models and tasks. Thought positions $T_t$ contribute near zero on graph-hopping; on arithmetic-reasoning, $T_{\text{full}}$ yields the strongest recovery for C, $C_u$, and CODI. \textbf{\textit{Takeaway:}} Latent-thought influence is graded---when large, it can override corrupted prompt positions and steer models toward correct paths. C and PaT show weaker thought recovery than on GPT-2, consistent with~\citet{coconut}'s latent-training recipe being less pronounced on Llama-3.2; CODI~\cite{shen2025codi}, whose recipe transfers across families still recovers strongly when all thoughts are restored (\S~\ref{sec:thought_use}).}
  \label{fig:causal_trace_llama}
\end{figure*}

Figure~\ref{fig:causal_trace_llama} reports the results on the causal-tracing experiment. All \textbf{graph-hopping} models and CODI in arithmetic-reasoning show near identical patterns to their GPT-2 counterparts, however while PaT shows weak but non-zero recoverability in the thought positions ($T_\text{full}$ specifically), C's thought-positions ($T_6$ \& $T_\text{full}$) show negative recoverability in some layers, showing that thoughts might actually hurt the model's reasoning. For PaT, C and C$_u$ in \textbf{arithmetic-reasoning}, even patching full thoughts has negligible restoration effect; however, patching the answer boundary at later layers shows high recoverability. CODI shows high recoverability when all thoughts are restored. This aligns with~\citet{coconut}'s account that their latent training recipe provides more pronounced results on GPT-2 as compared to Llama-3.2-1B-Instruct, possibly due to larger models' extensive language pre-training making latent training more challenging. However,~\citet{shen2025codi} demonstrate similar gains across model families.

\begin{figure*}[!t]
  \centering
  \includegraphics[width=\textwidth]{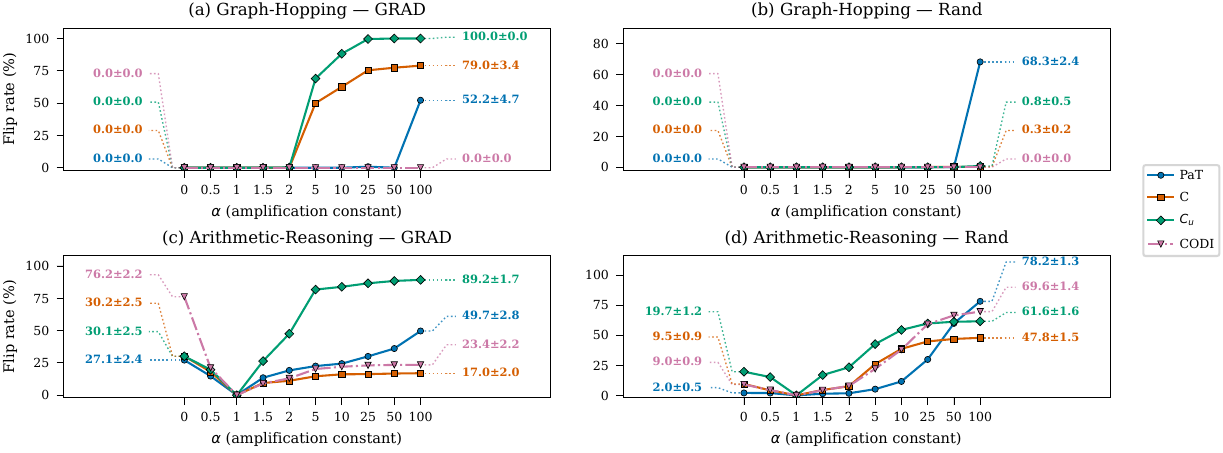}
  \caption{Gradient-subspace intervention flip rates (\%) across amplification strength $\alpha$ on the gradient-derived vs random control directions on the Llama-3.2 model family (\S~\ref{sec:thought_use}). \textbf{Graph-Hopping:} Robust to ablation ($\alpha=0$) but flips at high-$\alpha$ on GRAD. \textbf{Arithmetic-Reasoning:} Degradation under GRAD ablation; GRAD and Rand amplifications cause comparable flip-rates. \textbf{\textit{Takeaway:}} Thought utilization is \textit{graded} and depends on their influence power over model behavior.}
  \label{fig:amplification_heatmap_llama}
\end{figure*}

Figure~\ref{fig:amplification_heatmap_llama} reports the results on the gradient-subspace interventions experiment. On \textbf{graph-hopping}, the results remain comparable to the GPT-2 model family. C$_u$, now with negligible thought-necessity (Table~\ref{tab:ablation_main_llama}), shows comparable behavior to other models. PaT remains flat until a sudden spike at $\alpha=100$, possibly an off-manifold effect which is a known consequence of causal interventions~\cite{grant2026addressing}. \textbf{Arithmetic-reasoning} models show no noticeable deviations from their GPT-2 counterparts.

\paragraph{The Dynamics and Geometry of Latent Thoughts (\S~\ref{sec:geometry}).}

\begin{figure}[!t]
  \centering
  \includegraphics[width=\linewidth]{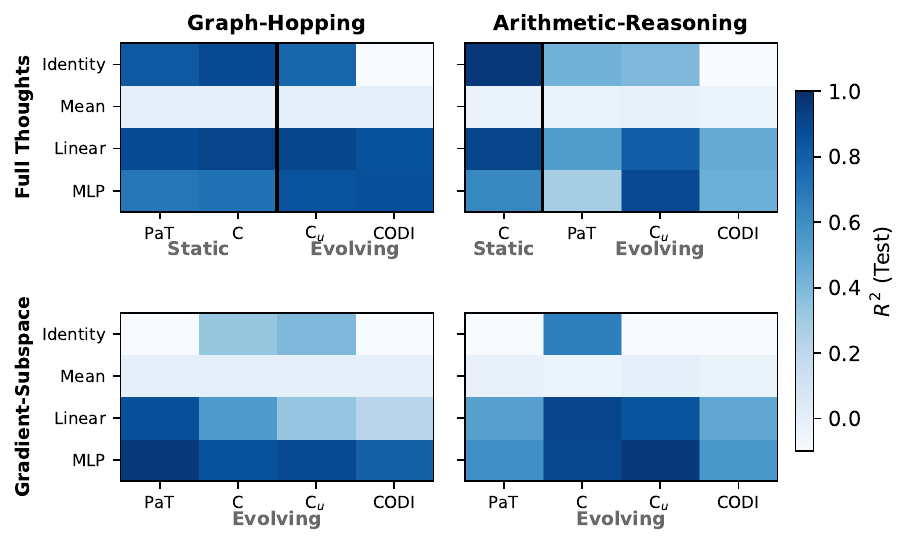}
  \caption{Markovianity of latent thoughts on the Llama-3.2 model family (\S~\ref{sec:geometry}). \textbf{Graph-Hopping:} Mixed dynamics; \textbf{Arithmetic-Reasoning:} Evolving dynamics (except C). \textbf{\textit{Takeaway:}} Unlike the clean task-split on GPT-2, full thought dynamics are mixed across tasks, yet static appearing full-thoughts still show evolving dynamics within the gradient-subspace where influence concentrates.}
  \label{fig:markov_llama}
\end{figure}

Figure~\ref{fig:markov_llama} reports the results on the Markovianity of thought trajectories experiment. Results remain comparable to the GPT-2 counterpart, especially in the gradient subspaces. Notably, C$_u$ on graph-hopping still sides in the evolving column despite negligible thought-necessity (Table~\ref{tab:ablation_main_llama}), demonstrating the impact of its stage-mixed curricula over the evolution of its latent thoughts. Identity collapses for CODI on \textbf{graph-hopping} which is best approximated by linear and MLP. All \textbf{arithmetic-reasoning} models show evolving dynamics except for C, which is overwhelmingly static.

\begin{figure}[!t]
  \centering
  \includegraphics[width=\linewidth]{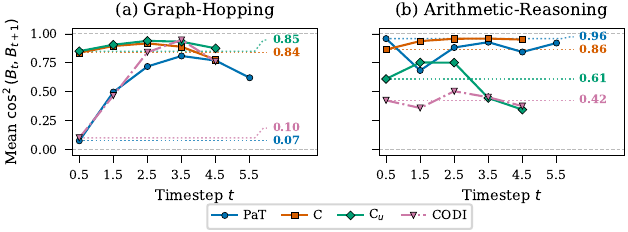}
  \caption{Geometric stability of the gradient-subspaces of the Llama-3.2 model family (\S~\ref{sec:geometry}). \textbf{Graph-Hopping:} C and C$_u$ stable, PaT and CODI show initial rotations; \textbf{Arithmetic-Reasoning:} PaT and C stable, C$_u$ and CODI rotating. \textbf{\textit{Takeaway:}} Unlike GPT-2, subspace geometry neither splits cleanly by task nor tracks thought influence.}
  \label{fig:geometric_stability_llama}
\end{figure}

Figure~\ref{fig:geometric_stability_llama} reports the results on the geometric stability of gradient-subspaces experiment. On \textbf{graph-hopping}, C and C$_u$ remain highly stable. PaT appears to show a rotating subspace; Table~\ref{tab:subspace_dim_full_gpt2} shows ranks collapsed to 1--2 at all positions. With such few dominant directions, any change produces a large swing. A similar story follows for CODI except that its rotations stabilize after two steps. On \textbf{arithmetic-reasoning}, PaT and C remain stable; notably, PaT's rank again collapses yet alignment stays high, the opposite behavior from graph-hopping. C$_u$'s and CODI's subspaces show high rotation across steps.

\section{Extended Analyses}
\label{app:extended_analyses}
This section reports additional results for experiments conducted in \S~\ref{sec:epiphenomena}, \S~\ref{sec:thought_use} and \S~\ref{sec:geometry}.

\begin{table}[h!]
\centering
\tiny
\begin{tabular}{lccccccc}
\toprule
$k$ & 0 & 1 & 2 & 3 & 4 & 5 & 6 \\
\midrule
Depth probed & 1 & 1 & 2 & 3 & 4 & 5 & 6 \\
\makecell{$n$ (instances \\ with $\geq$2 candidates)} & 461 & 461 & 486 & 439 & 182 & 26 & 3 \\
\bottomrule
\end{tabular}
\caption{Sample size at each probing depth. Results at $k \geq 5$ are computed over too few instances ($n{<}50$) to support aggregate conclusions and are omitted from the main analysis.}
\label{tab:prosqa_sample_sizes}
\end{table}

\paragraph{Superposition and BFS-like search on Graph-Hopping.}
The main text reports aggregate results for $k{=}0\ldots4$ only. \citet{coconut} illustrate the probing methodology on individual examples; Table~\ref{tab:prosqa_sample_sizes} shows that the aggregate version necessarily confronts the fact that most ProsQA graphs do not extend to depth 6.


\begin{figure*}[h!]
    \centering
    \begin{subfigure}{\linewidth}
        \centering
        \includegraphics[width=\linewidth]{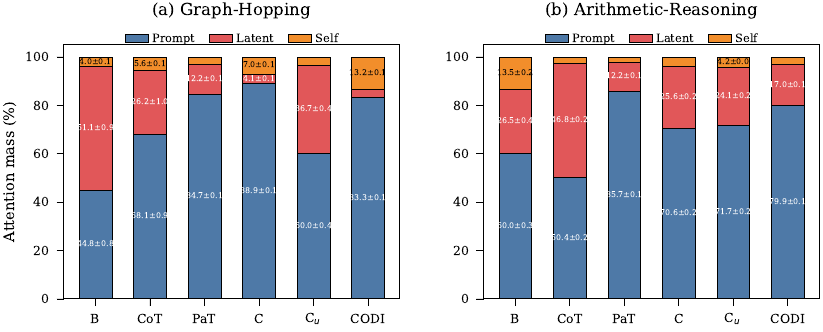}
        \caption{GPT-2.}
        \label{fig:attention_mass_gpt2}
    \end{subfigure}

    \vspace{0.5em}

    \begin{subfigure}{\linewidth}
        \centering
        \includegraphics[width=\linewidth]{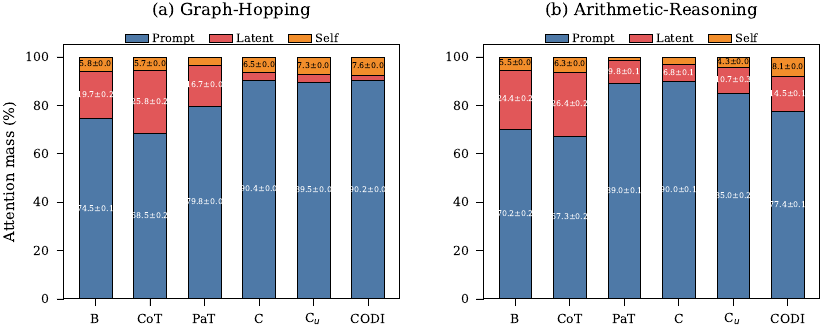}
        \caption{Llama-3.2.}
        \label{fig:attention_mass_llama}
    \end{subfigure}

    \caption{Attention mass distribution of the boundary token across prompt tokens, latent thoughts, and itself, averaged across all transformer layers and heads for GPT-2 and Llama-3.2 model families.}
    \label{fig:attention_mass}
\end{figure*}

\paragraph{Attention Mass Distribution.} 
As an additional demonstration of epiphenomenal patterns in interpretability, we evaluate how much attention the answer-generating boundary token directs at the latent thought tokens versus the prompt tokens. Let $t_{end}$ be the sequence position of this boundary token. We extract attention distribution over all preceding key positions $j$, averaged across all $L$ transformer layers and $H$ attention heads:
\[
    \bar{\alpha}_{j} = \frac{1}{L \cdot H} \sum_{l=1}^{L} \sum_{h=1}^{H} \alpha_{l,h}(t_{end}, j)
\]
where $\alpha_{l,h}$ is the attention weight at layer $l$ and head $h$. The total attention mass for a given sequence partition $\mathcal{S}$ (e.g., prompt or latent thoughts) is calculated as $\sum_{j \in \mathcal{S}} \bar{\alpha}_{j}$. Results in Figure~\ref{fig:attention_mass}.

\textbf{GPT-2:} On \textbf{graph-hopping}, PaT and C$_u$ attend to their latent thoughts while C and CODI largely ignore them. Despite no training, both controls B and CoT place significant mass on their (forced) thoughts. For \textbf{arithmetic-reasoning}, all models place considerable mass on their thought tokens.

\textbf{Llama-3.2:} On \textbf{graph-hopping}, B, CoT and PaT place substantial mass on the latent thoughts while C, C$_u$ and CODI largely ignore them. \textbf{Arithmetic-reasoning} has a similar story, except the attention of C, C$_u$ and CODI slightly increases.


\begin{figure*}[!h]
    \centering
    \begin{subfigure}{\linewidth}
        \centering
        \includegraphics[width=\linewidth]{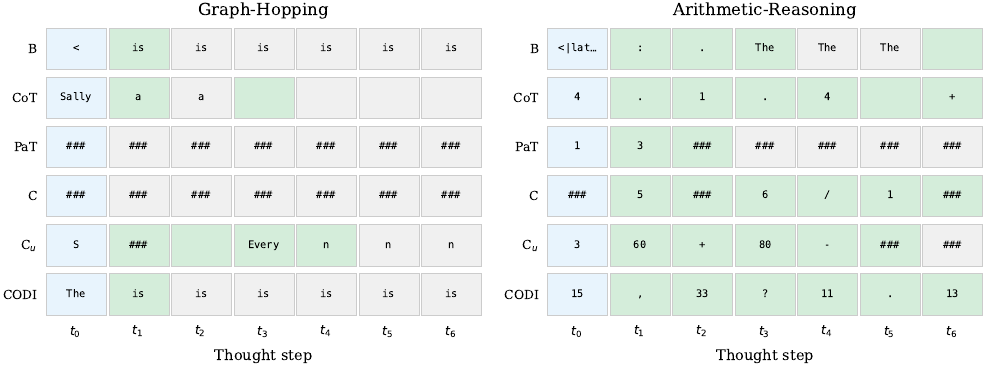}
        \caption{GPT-2.}
        \label{fig:logit_lens_decoded_example_gpt2}
    \end{subfigure}

    \vspace{0.5em}

    \begin{subfigure}{\linewidth}
        \centering
        \includegraphics[width=\linewidth]{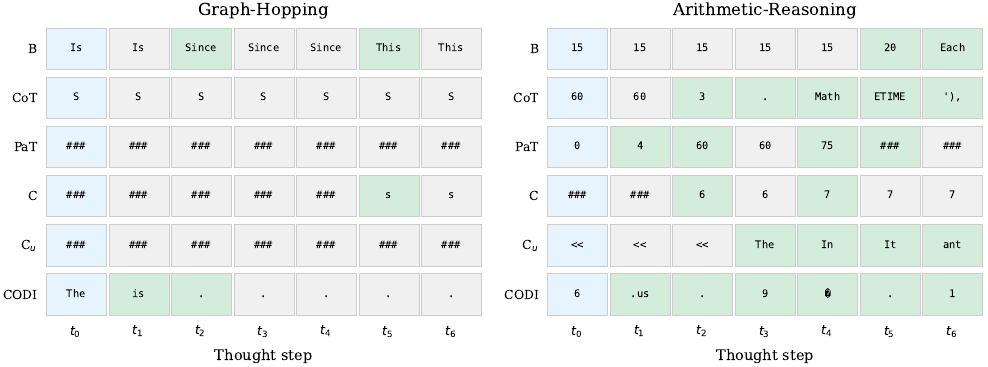}
        \caption{Llama-3.2.}
        \label{fig:logit_lens_decoded_example_llama}
    \end{subfigure}

    \caption{Logit-lens projection examples for both tasks (graph hopping and arithmetic reasoning) on the GPT-2 and Llama-3.2 model families. The final-layer hidden states at every thought position are decoded through the LM head to obtain the vocabulary projections.}
    \label{fig:logit_lens_decoded_example}
\end{figure*}

\paragraph{Logit-Lens Decoded Thought Trajectories.}
Figure~\ref{fig:logit_lens_decoded_example} illustrates representative logit-lens trajectories. 

\textbf{GPT-2:} On \textbf{graph-hopping}, PaT and C collapse to the answer delimiter at every position despite strong task performance; only C$_u$ produces somewhat evolving projections, while CODI and the controls remain semantically incoherent. On \textbf{arithmetic-reasoning}, B has no decodable structure while CoT shows moderate decodability despite no latent training---suggesting inherited structure from \textit{CoT} fine-tuning. PaT collapses to the answer delimiter at most positions again. CODI exhibits the clearest alternation between intermediate results and formatting tokens; C has lower decodability but similar alternation while C$_u$ shows alternation with operators at non-result positions.

\textbf{Llama-3.2:} On \textbf{graph-hopping}, PaT and C$_u$ collapse entirely to the answer delimiter at every position. C similarly decays into a mix of delimiters and static formatting tokens. CODI loops continuously on filler words. Controls B and CoT remain semantically incoherent. On \textbf{arithmetic-reasoning}, latent trajectories exhibit poorer overall decodability compared to the GPT-2 models. PaT decodes to a mix of numbers and delimiters, never laying out an ordered scratchpad. C initially decodes to answer delimiters, then degrades into repeating static numbers, whereas C$_u$ instead drifts across operators and random tokens without settling on coherent intermediates. CODI displays a mix of numerical tokens, text, and operators, but lacks the structured scratchpad alternation observed in its GPT-2 counterpart. Controls B and CoT output scattered numbers but fail to form coherent intermediate steps.

\begin{figure*}[!t]
  \centering

  \begin{subfigure}[t]{1\textwidth}
    \centering
    \includegraphics[width=\linewidth]{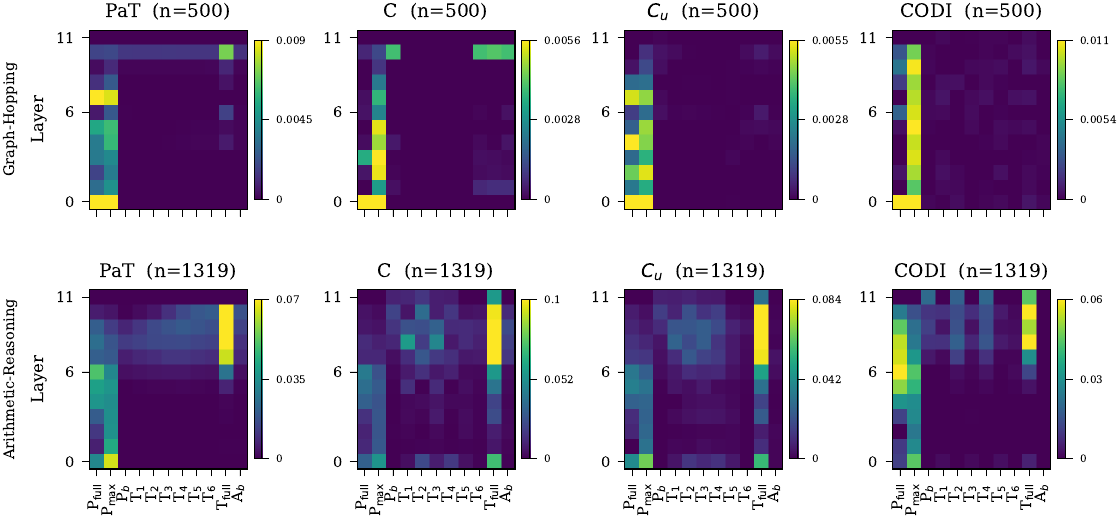}
    \caption{Attention outputs.}
    \label{fig:causal_trace_attn_gpt2}
  \end{subfigure}
  \hfill
  \begin{subfigure}[t]{1\textwidth}
    \centering
    \includegraphics[width=\linewidth]{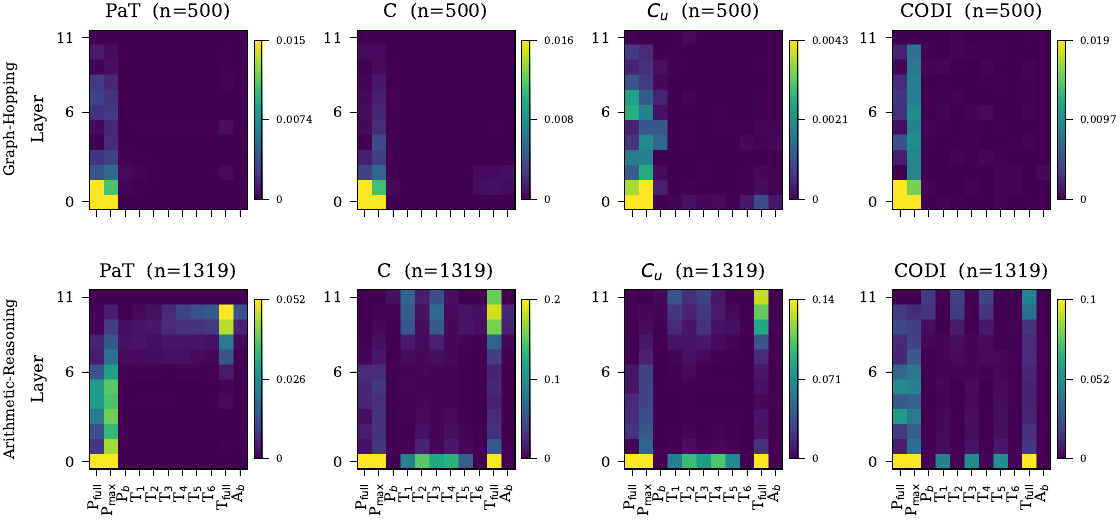}
    \caption{MLP outputs.}
    \label{fig:causal_trace_mlp_gpt2}
  \end{subfigure}

  \caption{Per-layer $\mathrm{IE_{KL}}$ under partner-prompt corruption across buckets $\{P_\text{full}, P_\text{max}, P_b, T_1, \ldots, T_K, T_\text{full}, A_b\}$, decomposed into (a) attention and (b) MLP outputs for the GPT-2 family. Unlike the full residual stream (Figure~\ref{fig:causal_trace}), thought positions $T_t$ on \textbf{graph-hopping} carry a \textit{small but nonzero} causal effect, most visible in the attention decomposition. \textbf{Arithmetic-reasoning} mirrors the full-residual pattern, with $T_\text{full}$ recovering most strongly for C, C$_u$, and CODI. \textit{Takeaway:} Latent-thought influence is graded rather than binary---even where it appears negligible, thought positions exert a measurable causal effect.}
  \label{fig:causal_trace_gpt2_extended}
\end{figure*}

\begin{figure*}[!t]
  \centering

  \begin{subfigure}[t]{1\textwidth}
    \centering
    \includegraphics[width=\linewidth]{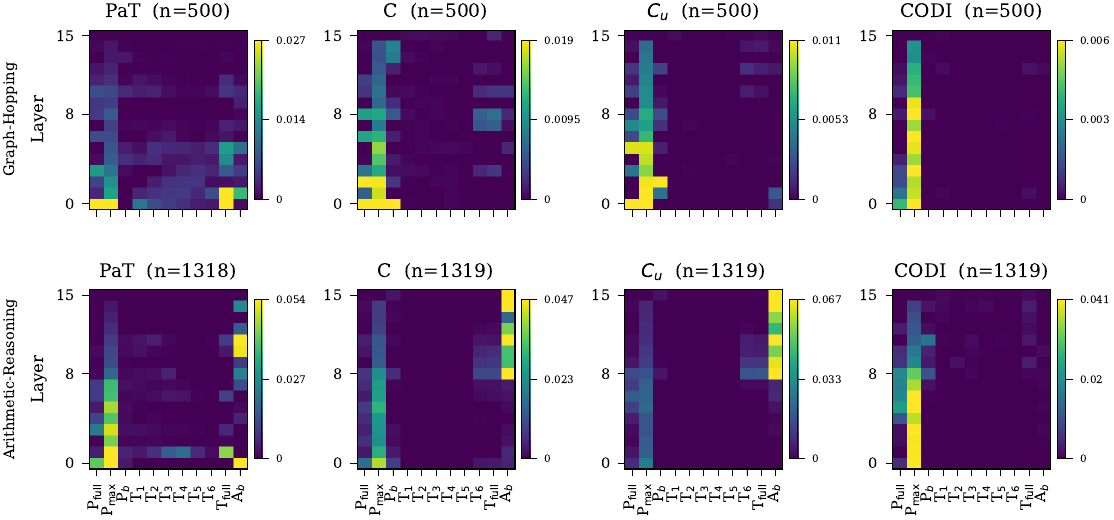}
    \caption{Attention outputs.}
    \label{fig:causal_trace_attn_llama}
  \end{subfigure}
  \hfill
  \begin{subfigure}[t]{1\textwidth}
    \centering
    \includegraphics[width=\linewidth]{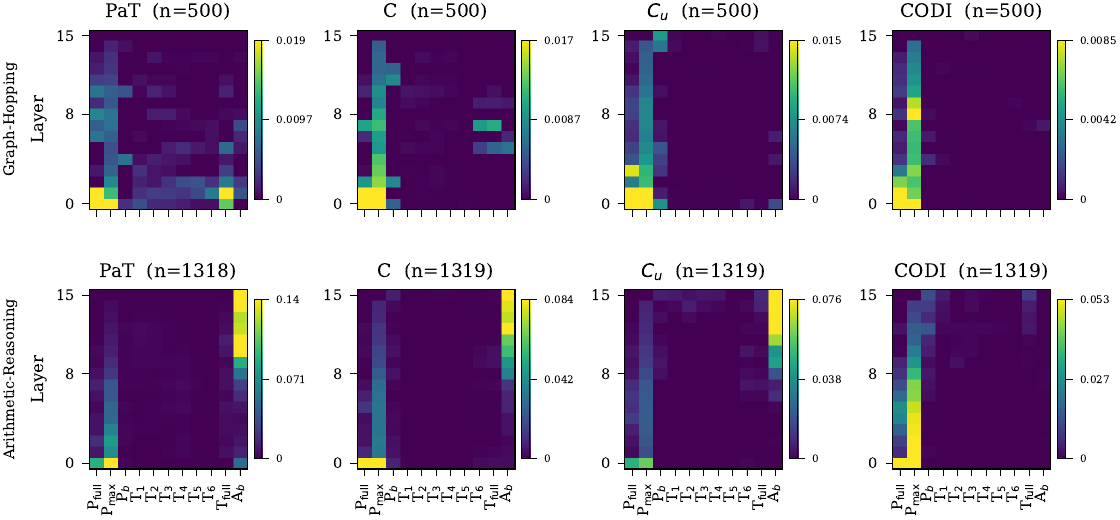}
    \caption{MLP outputs.}
    \label{fig:causal_trace_mlp_llama}
  \end{subfigure}

  \caption{Per-layer $\mathrm{IE_{KL}}$ under partner-prompt corruption across buckets $\{P_\text{full}, P_\text{max}, P_b, T_1, \ldots, T_K, T_\text{full}, A_b\}$, decomposed into (a) attention and (b) MLP outputs for the Llama-3.2 family. Unlike the full residual stream (Figure~\ref{fig:causal_trace_llama}), thought positions $T_t$ on \textbf{graph-hopping} carry a \textit{small but nonzero} causal effect in both attention and MLP output decompositions. \textbf{Arithmetic-reasoning} mirrors the full-residual pattern. \textit{Takeaway:} Latent-thought influence is graded rather than binary---even where it appears negligible, thought positions exert a measurable causal effect.}
  \label{fig:causal_trace_llama_extended}
\end{figure*}

\paragraph{Decomposition of Residual Stream Causal Tracing Recovery into Attention and MLP-Outputs.} Results for GPT-2 family in Figures~\ref{fig:causal_trace_attn_gpt2} \&~\ref{fig:causal_trace_mlp_gpt2} and for Llama-3.2 family in Figures~\ref{fig:causal_trace_attn_llama} \&~\ref{fig:causal_trace_mlp_llama}. 

\textbf{GPT-2:} In contrast to the full residual stream (Figure~\ref{fig:causal_trace}), the thought positions of \textbf{graph-hopping} models show \textit{small but real} causal effect (specially in attention decomposition). This further evidences that thought use isn't necessarily binary but depends on the influence carried by the thoughts. The \textbf{arithmetic-reasoning} models show patterns similar to the full residual stream (Figure~\ref{fig:causal_trace}).

\textbf{Llama-3.2:} Similarly to the GPT-2 family, the thought positions of \textbf{graph-hopping} models show \textit{small but real} causal effect in both attention and MLP output decompositions. \textbf{Arithmetic-reasoning} models yet again show similar patterns as in the full residual stream (Figure~\ref{fig:causal_trace_llama}).

\paragraph{Gradient-Subspace Dimensionality.}
Table~\ref{tab:subspace_dim_full_gpt2} (GPT-2) and Table~\ref{tab:subspace_dim_full_llama} (Llama-3.2) report per-timestep gradient-subspace diagnostics: rank $k_t$ at $\rho{=}0.95$, mean rank $\bar k$ (excluding the degenerate final recurrent step), adjacent and off-diagonal $\overline{\cos^2}(B_t, B_{t'})$, and the norm fraction $\|h^c\|/\|h\|$ retained in the gradient-subspace (\S~\ref{sec:thought_use}). Causal influence concentrates along few directions (relative to full dimensions) across tasks and model families.

\textbf{GPT-2:}  On \textbf{graph-hopping}, the subspaces are extremely low-dimensional for all models; on \textbf{arithmetic-reasoning}, ranks grow (yet still small relative to full dimensionality) except for CODI whose mean rank stays comparable across tasks, plausibly a consequence of its LoRA training.

\textbf{Llama-3.2:} We see a similar story as in the GPT-2 counterpart. The exceptions being PaT on arithmetic-reasoning with its surprisingly low rank that compares to its graph-hopping counterpart, and CODI whose ranks across tasks no longer match.

\begin{table*}[h!]
\centering
\small
\setlength{\tabcolsep}{2.5pt}
\begin{tabular}{llrrrrrrrrccr}
\toprule
Task & Model & $k_{0}$ & $k_{1}$ & $k_{2}$ & $k_{3}$ & $k_{4}$ & $k_{5}$ & $k_{6}$ & $\bar k$ & Adj.\ $\overline{\cos^2}$ & Off-diag.\ $\overline{\cos^2}$ & $\|h^c\|/\|h\|$ \\
\midrule
\multirow{4}{*}{\makecell[l]{Graph-\\Hopping}}
 & PaT   & 15  & 16  & 16  & 16  & 17  & 17  & 17  & 16.3  & $0.871$ & $0.697 {\scriptstyle \pm 0.026}$ & $0.353$ \\
 & C     & 13  & 12  & 11  & 11  & 12  & 12  & \textcolor{gray}{--} & 11.8  & $0.972$ & $0.671 {\scriptstyle \pm 0.024}$ & $0.026$ \\
 & C$_u$ & 15  & 13  & 17  & 17  & 17  & 14  & \textcolor{gray}{--} & 15.5  & $0.486$ & $0.197 {\scriptstyle \pm 0.012}$ & $0.045$ \\
 & CODI  & 20  & 33  & 35  & 36  & 36  & 37  & \textcolor{gray}{--} & 32.8  & $0.898$ & $0.587 {\scriptstyle \pm 0.008}$ & $0.100$ \\
\midrule
\multirow{4}{*}{\makecell[l]{Arithmetic-\\Reasoning}}
 & PaT   & 160 & 145 & 146 & 109 & 122 & 156 & 153 & 141.6 & $0.711$ & $0.623 {\scriptstyle \pm 0.019}$ & $0.378$ \\
 & C     & 155 & 100 & 129 & 135 & 128 & 115 & \textcolor{gray}{--} & 127.0 & $0.512$ & $0.394 {\scriptstyle \pm 0.021}$ & $0.369$ \\
 & C$_u$ & 126 & 114 & 115 & 122 & 119 & 134 & \textcolor{gray}{--} & 121.7 & $0.574$ & $0.404 {\scriptstyle \pm 0.014}$ & $0.297$ \\
 & CODI  & 51  & 14  & 64  & 18  & 57  & 13  & \textcolor{gray}{--} & 36.2  & $0.271$ & $0.351 {\scriptstyle \pm 0.016}$ & $0.306$ \\
\bottomrule
\end{tabular}
\caption{Gradient-subspace diagnostics across tasks for the GPT-2 model family.}
\label{tab:subspace_dim_full_gpt2}
\end{table*}

\begin{table*}[h!]
\centering
\small
\setlength{\tabcolsep}{2.5pt}
\begin{tabular}{llrrrrrrrrccr}
\toprule
Task & Model & $k_{0}$ & $k_{1}$ & $k_{2}$ & $k_{3}$ & $k_{4}$ & $k_{5}$ & $k_{6}$ & $\bar k$ & Adj.\ $\overline{\cos^2}$ & Off-diag.\ $\overline{\cos^2}$ & $\|h^c\|/\|h\|$ \\
\midrule
\multirow{4}{*}{\makecell[l]{Graph-\\Hopping}} & PaT & 1 & 1 & 1 & 1 & 1 & 1 & 2 & 1.1 & $0.582$ & $0.304 {\scriptstyle \pm 0.033}$ & $0.019$ \\
 & C & 2 & 2 & 2 & 2 & 2 & 2 & \textcolor{gray}{--} & 2.0 & $0.863$ & $0.469 {\scriptstyle \pm 0.020}$ & $0.022$ \\
 & C$_u$ & 1 & 1 & 1 & 1 & 1 & 1 & \textcolor{gray}{--} & 1.0 & $0.901$ & $0.516 {\scriptstyle \pm 0.035}$ & $0.005$ \\
 & CODI & 2 & 2 & 2 & 2 & 2 & 2 & \textcolor{gray}{--} & 2.0 & $0.623$ & $0.306 {\scriptstyle \pm 0.026}$ & $0.018$ \\
\midrule
\multirow{4}{*}{\makecell[l]{Arithmetic-\\Reasoning}} & PaT & 1 & 24 & 11 & 5 & 3 & 3 & 3 & 7.1 & $0.870$ & $0.790 {\scriptstyle \pm 0.034}$ & $0.117$ \\
 & C & 397 & 345 & 291 & 261 & 251 & 270 & \textcolor{gray}{--} & 302.5 & $0.934$ & $0.590 {\scriptstyle \pm 0.004}$ & $0.574$ \\
 & C$_u$ & 259 & 357 & 386 & 305 & 268 & 250 & \textcolor{gray}{--} & 304.2 & $0.581$ & $0.354 {\scriptstyle \pm 0.006}$ & $0.385$ \\
 & CODI & 185 & 159 & 150 & 214 & 167 & 131 & \textcolor{gray}{--} & 167.7 & $0.423$ & $0.310 {\scriptstyle \pm 0.010}$ & $0.554$ \\
\bottomrule
\end{tabular}
\caption{Gradient-subspace diagnostics across tasks for the Llama-3.2 model family.}
\label{tab:subspace_dim_full_llama}
\end{table*}

\begin{figure*}[!t]
  \centering
  \begin{subfigure}{\textwidth}
    \centering
    \includegraphics[width=\textwidth]{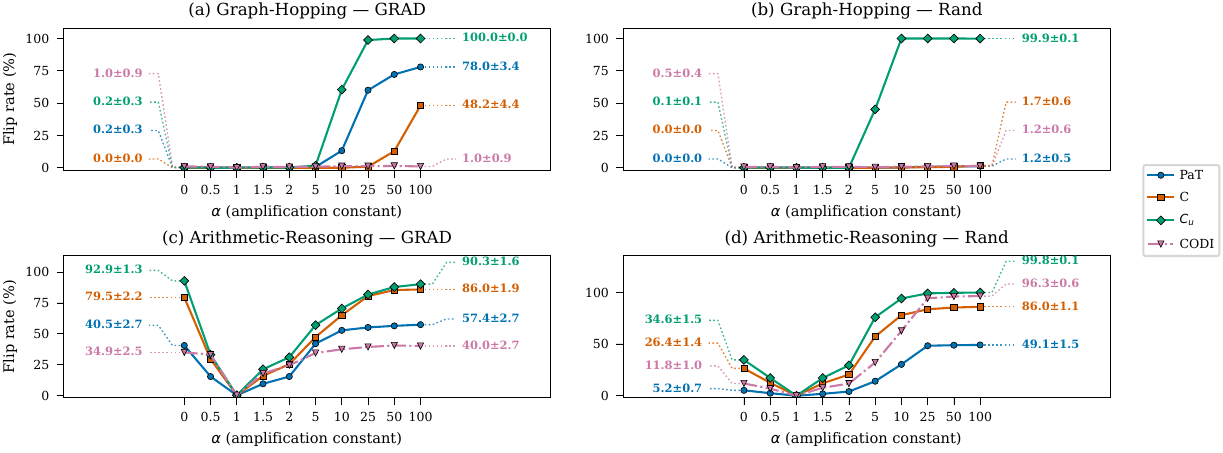}
    \caption{GPT-2 model-family, predicted-token subspace.}
    \label{fig:amplification_predtoken_gpt2}
  \end{subfigure}
  \\[0.5em]
  \begin{subfigure}{\textwidth}
    \centering
    \includegraphics[width=\textwidth]{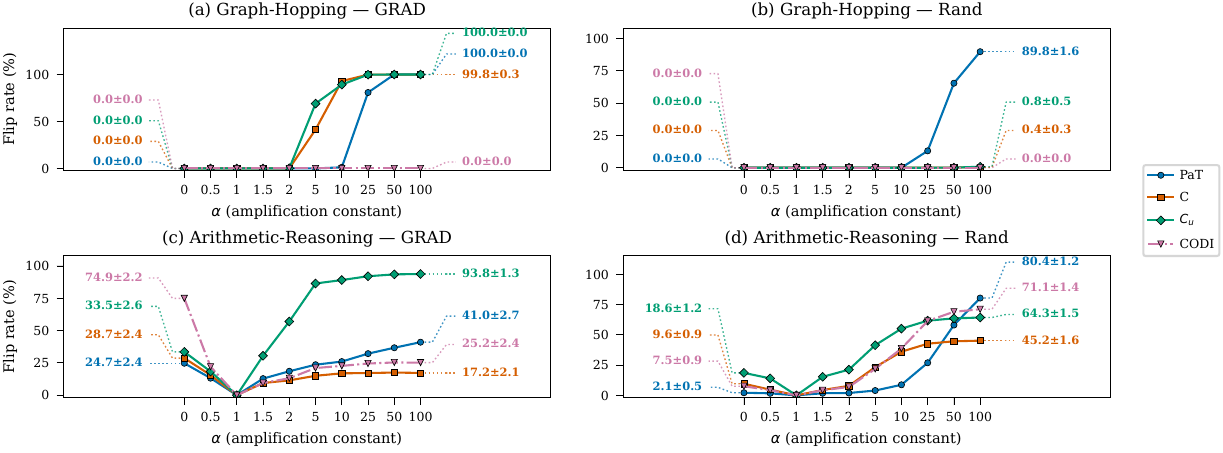}
    \caption{Llama-3.2 model-family, predicted-token subspace.}
    \label{fig:amplification_predtoken_llama}
  \end{subfigure}
  \caption{Gradient-subspace intervention flip rates (\%) across amplification strength $\alpha$ on the gradient-derived (predicted-token) vs random control directions (\S~\ref{sec:thought_use}). \textbf{Graph-Hopping:} Robust to ablation ($\alpha=0$) but flips at high-$\alpha$ on GRAD. \textbf{Arithmetic-Reasoning:} Degradation under GRAD ablation; GRAD and Rand amplifications cause comparable flip-rates. \textbf{\textit{Takeaway:}} Thought utilization is \textit{graded} and depends on their influence power over model behavior. Predicted-token subspaces replicate the interventional patterns of the gold subspace.}
  \label{fig:amplification_predtoken}
\end{figure*}

\begin{figure*}[!t]
  \centering
  \begin{subfigure}{0.49\linewidth}
    \centering
    \includegraphics[width=\linewidth]{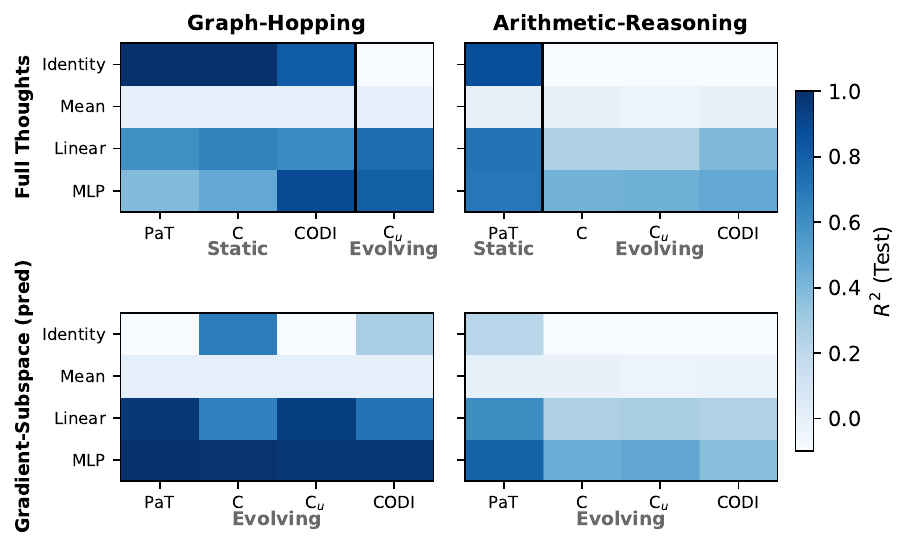}
    \caption{GPT-2 model-family, predicted-token subspace.}
    \label{fig:markov_heatmaps_gpt2_pred}
  \end{subfigure}
  \hfill
  \begin{subfigure}{0.49\linewidth}
    \centering
    \includegraphics[width=\linewidth]{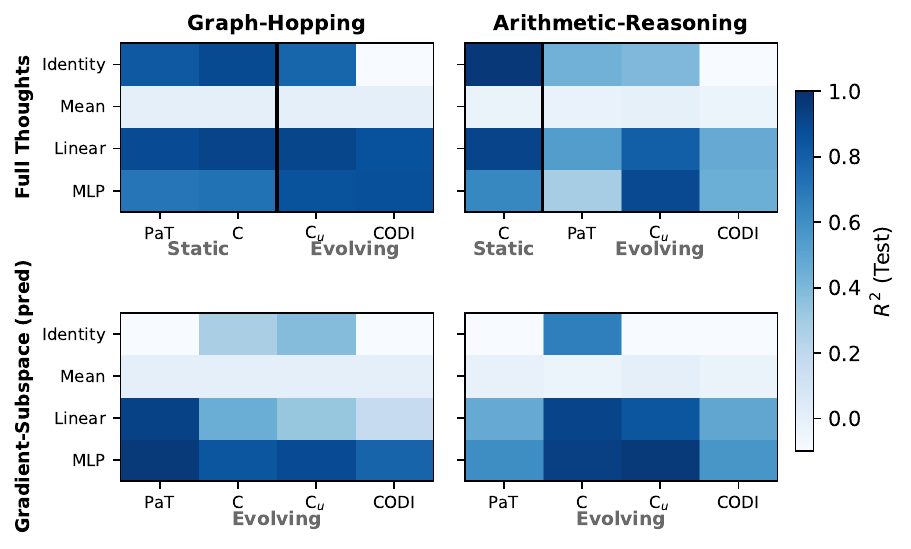}
    \caption{Llama-3.2 model-family, predicted-token subspace.}
    \label{fig:markov_heatmaps_llama_pred}
  \end{subfigure}
  \caption{Markovianity of latent thoughts (\S~\ref{sec:geometry}). (a) \textbf{\textit{GPT-2.}} \textbf{Graph-Hopping:} Static dynamics (except C$_u$); \textbf{Arithmetic-Reasoning:} Evolving dynamics. (b) \textbf{\textit{Llama-3.2.}} \textbf{Graph-Hopping:} Mixed dynamics; \textbf{Arithmetic-Reasoning:} Evolving dynamics (except C).  \textbf{\textit{Takeaway:}} In both families, static-appearing full-thoughts show evolving dynamics within the predicted-token gradient-subspace where influence concentrates. Predicted-token subspaces replicate the geometric patterns of the gold subspace.}
  \label{fig:markov_heatmaps_pred}
\end{figure*}

\begin{figure*}[!t]
  \centering
  \begin{subfigure}{0.49\linewidth}
    \centering
    \includegraphics[width=\linewidth]{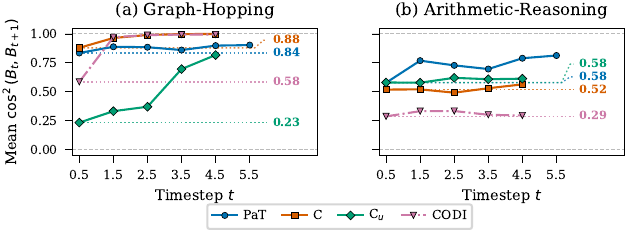}
    \caption{GPT-2 model-family, predicted-token subspace.}
    \label{fig:cos_panels_gpt2_pred}
  \end{subfigure}
  \hfill
  \begin{subfigure}{0.49\linewidth}
    \centering
    \includegraphics[width=\linewidth]{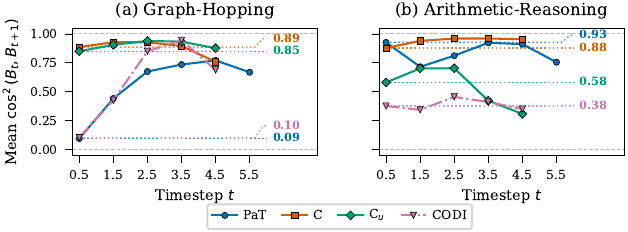}
    \caption{Llama-3.2 model-family, predicted-token subspace.}
    \label{fig:cos_panels_llama_pred}
  \end{subfigure}
  \caption{Geometric stability of the gradient-subspaces (predicted-token) (\S~\ref{sec:geometry}). (a) \textbf{\textit{GPT-2.}} \textbf{Graph-Hopping:} Stable subspaces (except C$_u$); \textbf{Arithmetic-Reasoning:} Rotating subspaces. (b) \textbf{\textit{Llama-3.2.}} \textbf{Graph-Hopping:} C and C$_u$ stable, PaT and CODI show initial rotations; \textbf{Arithmetic-Reasoning:} PaT and C stable, C$_u$ and CODI rotating. \textbf{\textit{Takeaway:}} The weak-influence$\rightarrow$stable-geometry dichotomy does not transfer from GPT-2 to Llama-3.2 where stability does not cleanly separate by task. Predicted-token subspaces replicate the geometric patterns of the gold subspace.}
  \label{fig:cos_panels_pred}
\end{figure*}

\begin{table*}[h!]
\centering
\small
\setlength{\tabcolsep}{2.5pt}
\begin{tabular}{llrrrrrrrrccr}
\toprule
Task & Model & $k_{0}$ & $k_{1}$ & $k_{2}$ & $k_{3}$ & $k_{4}$ & $k_{5}$ & $k_{6}$ & $\bar k$ & Adj.\ $\overline{\cos^2}$ & Off-diag.\ $\overline{\cos^2}$ & $\|h^c\|/\|h\|$ \\
\midrule
\multirow{4}{*}{\makecell[l]{Graph-\\Hopping}}
 & PaT   & 15  & 16  & 16  & 16  & 17  & 17  & 17  & 16.3  & $0.871$ & $0.697 {\scriptstyle \pm 0.026}$ & $0.353$ \\
 & C     & 13  & 12  & 11  & 11  & 12  & 12  & \textcolor{gray}{--} & 11.8  & $0.972$ & $0.671 {\scriptstyle \pm 0.024}$ & $0.026$ \\
 & C$_u$ & 15  & 13  & 17  & 17  & 17  & 14  & \textcolor{gray}{--} & 15.5  & $0.486$ & $0.197 {\scriptstyle \pm 0.012}$ & $0.045$ \\
 & CODI  & 20  & 33  & 35  & 36  & 36  & 37  & \textcolor{gray}{--} & 32.8  & $0.898$ & $0.587 {\scriptstyle \pm 0.008}$ & $0.100$ \\
\midrule
\multirow{4}{*}{\makecell[l]{Arithmetic-\\Reasoning}}
 & PaT   & 160 & 145 & 146 & 109 & 122 & 156 & 153 & 141.6 & $0.711$ & $0.623 {\scriptstyle \pm 0.019}$ & $0.378$ \\
 & C     & 155 & 100 & 129 & 135 & 128 & 115 & \textcolor{gray}{--} & 127.0 & $0.512$ & $0.394 {\scriptstyle \pm 0.021}$ & $0.369$ \\
 & C$_u$ & 126 & 114 & 115 & 122 & 119 & 134 & \textcolor{gray}{--} & 121.7 & $0.574$ & $0.404 {\scriptstyle \pm 0.014}$ & $0.297$ \\
 & CODI  & 51  & 14  & 64  & 18  & 57  & 13  & \textcolor{gray}{--} & 36.2  & $0.271$ & $0.351 {\scriptstyle \pm 0.016}$ & $0.306$ \\
\bottomrule
\end{tabular}
\caption{Gradient-subspace (predicted-token) diagnostics across tasks for the GPT-2 model family.}
\label{tab:subspace_geometry_gpt2_pred}
\end{table*}

\begin{table*}[h!]
\centering
\small
\setlength{\tabcolsep}{2.5pt}
\begin{tabular}{llrrrrrrrrccr}
\toprule
Task & Model & $k_{0}$ & $k_{1}$ & $k_{2}$ & $k_{3}$ & $k_{4}$ & $k_{5}$ & $k_{6}$ & $\bar k$ & Adj.\ $\overline{\cos^2}$ & Off-diag.\ $\overline{\cos^2}$ & $\|h^c\|/\|h\|$ \\
\midrule
\multirow{4}{*}{\makecell[l]{Graph-\\Hopping}} & PaT & 4 & 4 & 4 & 4 & 4 & 4 & 4 & 4.0 & $0.564$ & $0.269 {\scriptstyle \pm 0.032}$ & $0.039$ \\
 & C & 1 & 1 & 1 & 1 & 1 & 1 & \textcolor{gray}{--} & 1.0 & $0.879$ & $0.488 {\scriptstyle \pm 0.020}$ & $0.012$ \\
 & C$_u$ & 1 & 1 & 1 & 1 & 1 & 1 & \textcolor{gray}{--} & 1.0 & $0.901$ & $0.516 {\scriptstyle \pm 0.035}$ & $0.005$ \\
 & CODI & 2 & 2 & 2 & 2 & 2 & 2 & \textcolor{gray}{--} & 2.0 & $0.603$ & $0.288 {\scriptstyle \pm 0.026}$ & $0.015$ \\
\midrule
\multirow{4}{*}{\makecell[l]{Arithmetic-\\Reasoning}} & PaT & 2 & 16 & 8 & 6 & 3 & 5 & 5 & 6.4 & $0.841$ & $0.711 {\scriptstyle \pm 0.054}$ & $0.117$ \\
 & C & 412 & 347 & 288 & 256 & 247 & 267 & \textcolor{gray}{--} & 302.8 & $0.938$ & $0.596 {\scriptstyle \pm 0.004}$ & $0.558$ \\
 & C$_u$ & 250 & 354 & 354 & 270 & 231 & 248 & \textcolor{gray}{--} & 284.5 & $0.543$ & $0.335 {\scriptstyle \pm 0.008}$ & $0.368$ \\
 & CODI & 153 & 144 & 128 & 176 & 140 & 101 & \textcolor{gray}{--} & 140.3 & $0.387$ & $0.285 {\scriptstyle \pm 0.014}$ & $0.527$ \\
\bottomrule
\end{tabular}
\caption{Gradient-subspace (predicted-token) diagnostics across tasks for the Llama-3.2 model family.}
\label{tab:subspace_geometry_llama_pred}
\end{table*}

\paragraph{Predicted-Token Gradient Subspace.}
The gradient-subspaces in the main text are isolated from the loss on the gold answer token $y^\star$, i.e.\ $g_{i,t} = \nabla_{h_{i,t}} \mathcal{L}_i(y^\star)$. To verify that our conclusions do not hinge on this choice, we repeat the full pipeline using gradients of the loss on the model's own predicted token $\hat{y} = \arg\max_{y} p_\theta(y \mid x, e^{\mathrm{lat}}_{1:K})$, i.e.\ $\tilde{g}_{i,t} = \nabla_{h_{i,t}} \mathcal{L}_i(\hat{y})$, yielding a predicted-token gradient-subspace $\tilde{B}_t$. We rerun (i) the gradient-subspace interventions (Figure~\ref{fig:amplification_predtoken_gpt2} for GPT-2 and Figure~\ref{fig:amplification_predtoken_llama} for Llama-3.2), (ii) the Markovianity analysis on the predicted-token subspace-projected thoughts (Figure~\ref{fig:markov_heatmaps_gpt2_pred} and Figure~\ref{fig:markov_heatmaps_llama_pred}), and (iii) the geometric-stability analysis of the predicted-token subspaces (Figure~\ref{fig:cos_panels_gpt2_pred} and Figure~\ref{fig:cos_panels_llama_pred}). We also report the corresponding per-timestep dimensionality diagnostics in Table~\ref{tab:subspace_geometry_gpt2_pred} (GPT-2) and Table~\ref{tab:subspace_geometry_llama_pred} (Llama-3.2). Results across all experiments and the dimensionality diagnostic remain nearly identical to the gold-subspace.

\paragraph{Gold--Predicted Token Gradient Subspace Alignment.}

\begin{table*}[h!]
\centering
\small
\setlength{\tabcolsep}{4pt}
\begin{tabular}{llrrrrrrrrr}
\toprule
Task & Model & $t_{0}$ & $t_{1}$ & $t_{2}$ & $t_{3}$ & $t_{4}$ & $t_{5}$ & $t_{6}$ & $\overline{\cos^2\theta}$ & Control $\overline{\cos^2\theta}$ \\
\midrule
\multirow{4}{*}{\makecell[l]{Graph-Hopping}} & PaT & 0.620 & 0.700 & 0.708 & 0.656 & 0.768 & 0.709 & 0.663 & 0.689 & 0.021 \\
 & C & 0.611 & 0.563 & 0.529 & 0.566 & 0.530 & 0.566 & -- & 0.561 & 0.013 \\
 & C$_u$ & 0.536 & 0.534 & 0.593 & 0.554 & 0.541 & 0.639 & -- & 0.566 & 0.019 \\
 & CODI & 0.802 & 0.718 & 0.725 & 0.721 & 0.724 & 0.722 & -- & 0.735 & 0.039 \\
\midrule
\multirow{4}{*}{\makecell[l]{Arithmetic-Reasoning}} & PaT & 0.702 & 0.694 & 0.678 & 0.668 & 0.677 & 0.662 & 0.670 & 0.679 & 0.194 \\
 & C & 0.676 & 0.679 & 0.669 & 0.669 & 0.722 & 0.651 & -- & 0.678 & 0.153 \\
 & C$_u$ & 0.720 & 0.703 & 0.691 & 0.700 & 0.713 & 0.679 & -- & 0.701 & 0.153 \\
 & CODI & 0.796 & 0.903 & 0.838 & 0.931 & 0.857 & 0.952 & -- & 0.880 & 0.045 \\
\bottomrule
\end{tabular}
\caption{Gold-label vs.\ predicted-token subspace alignment across tasks for the GPT-2 model family. Each $t_i$ column reports $\overline{\cos^2\theta}$ between $B_t^{\text{gold}}$ and $B_t^{\text{pred}}$ at that timestep (1 = same subspace, 0 = orthogonal); the $\overline{\cos^2\theta}$ column is the mean over $t$, and the control column is the same statistic for rank-matched random subspaces.}
\label{tab:gold_pred_angle_gpt2}
\end{table*}
\begin{table*}[h!]
\centering
\small
\setlength{\tabcolsep}{4pt}
\begin{tabular}{llrrrrrrrrr}
\toprule
Task & Model & $t_{0}$ & $t_{1}$ & $t_{2}$ & $t_{3}$ & $t_{4}$ & $t_{5}$ & $t_{6}$ & $\overline{\cos^2\theta}$ & Control $\overline{\cos^2\theta}$ \\
\midrule
\multirow{4}{*}{\makecell[l]{Graph-Hopping}} & PaT & 0.999 & 0.999 & 0.982 & 0.974 & 0.990 & 0.995 & 0.999 & 0.991 & 0.002 \\
 & C & 1.000 & 1.000 & 1.000 & 1.000 & 1.000 & 0.999 & -- & 1.000 & 0.001 \\
 & C$_u$ & 1.000 & 1.000 & 0.999 & 1.000 & 1.000 & 1.000 & -- & 1.000 & 0.000 \\
 & CODI & 0.510 & 0.503 & 0.509 & 0.504 & 0.502 & 0.500 & -- & 0.505 & 0.001 \\
\midrule
\multirow{4}{*}{\makecell[l]{Arithmetic-Reasoning}} & PaT & 1.000 & 0.784 & 0.802 & 0.958 & 0.989 & 0.954 & 0.988 & 0.925 & 0.004 \\
 & C & 0.532 & 0.552 & 0.590 & 0.629 & 0.642 & 0.625 & -- & 0.595 & 0.128 \\
 & C$_u$ & 0.405 & 0.414 & 0.411 & 0.369 & 0.310 & 0.311 & -- & 0.370 & 0.127 \\
 & CODI & 0.593 & 0.546 & 0.621 & 0.621 & 0.556 & 0.648 & -- & 0.598 & 0.070 \\
\bottomrule
\end{tabular}
\caption{Gold-label vs.\ predicted-token subspace alignment across tasks for the Llama-3.2 model family. Each $t_i$ column reports $\overline{\cos^2\theta}$ between $B_t^{\text{gold}}$ and $B_t^{\text{pred}}$ at that timestep (1 = same subspace, 0 = orthogonal); the $\overline{\cos^2\theta}$ column is the mean over $t$, and the control column is the same statistic for rank-matched random subspaces.}
\label{tab:gold_pred_angle_llama}
\end{table*}

Tables~\ref{tab:gold_pred_angle_gpt2} (GPT-2 model family) \&~\ref{tab:gold_pred_angle_llama} (Llama-3.2 model family) report the alignment between the gold and predicted-token gradient subspaces. The subspaces remain in alignment far above the random control directions.

\paragraph{Variance Decomposition of Latent-Thoughts.}
We report the static organization of latent-thoughts by decomposing their total variance $h_{i,t} \in \mathbb{R}^D$ (instance $i$, timestep $t$) into timestep-specific, instance-specific, and residual components using $\mu = \mathbb{E}_{i,t}[h_{i,t}]$ (global mean), $\mu_t = \mathbb{E}_i[h_{i,t}]$ (temporal-mean) and $\mu_i = \mathbb{E}_t[h_{i,t}]$ (instance-mean) as:
$$
\begin{aligned}
Var(h_{i,t}) &= \underbrace{\mathbb{E}_t\|\mu_t - \mu\|^2}_{Var_{\mathrm{time}}} + \underbrace{\mathbb{E}_i\|\mu_i - \mu\|^2}_{Var_{\mathrm{inst}}} \\
&\quad + \underbrace{\mathbb{E}_{i,t}\|h_{i,t} - \mu_i - \mu_t + \mu\|^2}_{Var_{\mathrm{residual}}}.
\end{aligned}
$$
For PaT, $t$ indexes spatial tokens; for C and C$_u$, it indexes recurrence-steps. Results in Table~\ref{tab:variance_decomposition} (GPT-2) and Table~\ref{tab:variance_decomposition_llama} (Llama-3.2).

\begin{table}[!h]
\centering
\footnotesize
\setlength{\tabcolsep}{3pt}
\begin{tabular}{l cc | cc}
\toprule
& \multicolumn{2}{c}{\textbf{Graph-Hopping}} & \multicolumn{2}{c}{\textbf{Arithmetic-Reasoning}} \\
\cmidrule(lr){2-3} \cmidrule(lr){4-5}
\textbf{Model} & \textbf{Var}$_\textbf{time}$ & \textbf{Var}$_\textbf{inst}$ & \textbf{Var}$_\textbf{time}$ & \textbf{Var}$_\textbf{inst}$ \\
\midrule
PaT & $0.6 {\scriptstyle \pm 0.1}$ & $98.7 {\scriptstyle \pm 0.1}$ & $1.9 {\scriptstyle \pm 0.2}$ & $72.6 {\scriptstyle \pm 0.8}$ \\
\midrule
C & $22.6 {\scriptstyle \pm 1.9}$ & $68.9 {\scriptstyle \pm 2.1}$ & $25.0 {\scriptstyle \pm 0.4}$ & $26.0 {\scriptstyle \pm 0.6}$ \\
\midrule
C$_u$ & $99.1 {\scriptstyle \pm 0.0}$ & $0.4 {\scriptstyle \pm 0.0}$ & $10.8 {\scriptstyle \pm 0.6}$ & $26.1 {\scriptstyle \pm 0.5}$ \\
\midrule
CODI & $89.8 {\scriptstyle \pm 0.7}$ & $7.4 {\scriptstyle \pm 0.6}$ & $5.6 {\scriptstyle \pm 0.3}$ & $46.3 {\scriptstyle \pm 1.1}$ \\
\bottomrule
\end{tabular}
\caption{Variance decomposition of latent thoughts (GPT-2 model family) into temporal, instance, and residual components. \textbf{\textit{Takeaway:}} Lack of generalizable structure suggests that observable static decomposition is uninformative by itself. PaT carries 1-2 orders smaller $\text{Var}_\text{time}$ than other models across tasks.}
\label{tab:variance_decomposition}
\end{table}

\begin{table}[!h]
\centering
\footnotesize
\setlength{\tabcolsep}{3pt}
\begin{tabular}{l cc | cc}
\toprule
& \multicolumn{2}{c}{\textbf{Graph-Hopping}} & \multicolumn{2}{c}{\textbf{Arithmetic-Reasoning}} \\
\cmidrule(lr){2-3} \cmidrule(lr){4-5}
\textbf{Model} & \textbf{Var}$_\textbf{time}$ & \textbf{Var}$_\textbf{inst}$ & \textbf{Var}$_\textbf{time}$ & \textbf{Var}$_\textbf{inst}$ \\
\midrule
PaT & $25.1 {\scriptstyle \pm 0.4}$ & $64.9 {\scriptstyle \pm 0.4}$ & $11.2 {\scriptstyle \pm 0.2}$ & $39.2 {\scriptstyle \pm 0.3}$ \\
\midrule
C & $24.7 {\scriptstyle \pm 0.4}$ & $67.1 {\scriptstyle \pm 0.4}$ & $6.2 {\scriptstyle \pm 0.1}$ & $85.3 {\scriptstyle \pm 0.1}$ \\
\midrule
C$_u$ & $63.4 {\scriptstyle \pm 0.5}$ & $32.9 {\scriptstyle \pm 0.5}$ & $6.1 {\scriptstyle \pm 0.1}$ & $85.6 {\scriptstyle \pm 0.1}$ \\
\midrule
CODI & $89.0 {\scriptstyle \pm 0.5}$ & $6.6 {\scriptstyle \pm 0.3}$ & $10.7 {\scriptstyle \pm 0.2}$ & $30.9 {\scriptstyle \pm 0.3}$ \\
\bottomrule
\end{tabular}
\caption{Variance decomposition of latent thoughts (Llama-3.2 model family) into temporal, instance, and residual components. \textbf{\textit{Takeaway:}} Lack of generalizable structure suggests that observable static decomposition is uninformative by itself.}
\label{tab:variance_decomposition_llama}
\end{table}

\begin{figure*}[!t]
    \centering

    \begin{subfigure}[b]{0.24\textwidth}
        \centering
        \includegraphics[width=\linewidth]{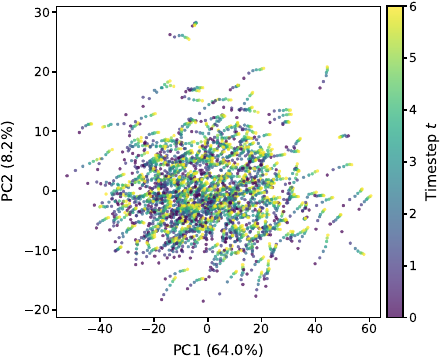}
        \caption{PaT graph-hopping}
        \label{fig:pca_pat_prosqa}
    \end{subfigure}
    \hfill
    \begin{subfigure}[b]{0.24\textwidth}
        \centering
        \includegraphics[width=\linewidth]{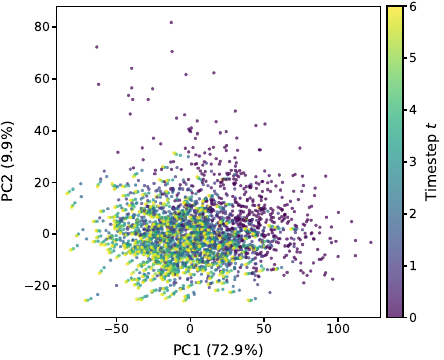}
        \caption{C graph-hopping}
        \label{fig:pca_c_prosqa}
    \end{subfigure}
    \hfill
    \begin{subfigure}[b]{0.24\textwidth}
        \centering
        \includegraphics[width=\linewidth]{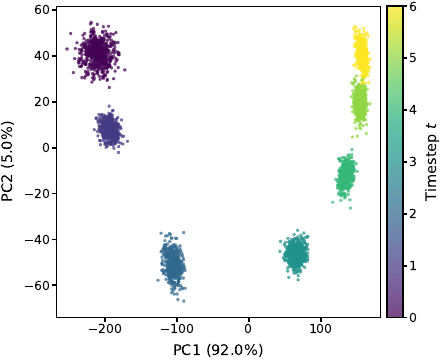}
        \caption{C$_u$ graph-hopping}
        \label{fig:pca_cu_prosqa}
    \end{subfigure}
    \hfill
    \begin{subfigure}[b]{0.24\textwidth}
        \centering
        \includegraphics[width=\linewidth]{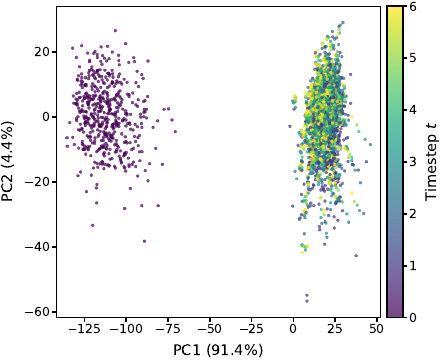}
        \caption{CODI graph-hopping}
        \label{fig:pca_codi_prosqa}
    \end{subfigure}

    \vspace{0.5em}

    \begin{subfigure}[b]{0.24\textwidth}
        \centering
        \includegraphics[width=\linewidth]{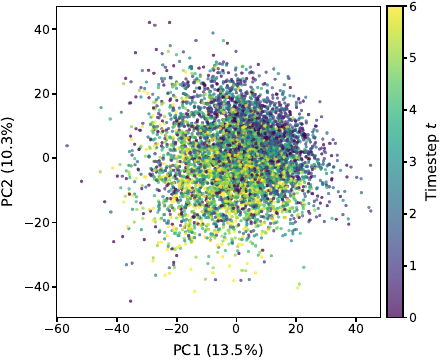}
        \caption{PaT arithmetic-reasoning}
        \label{fig:pca_pat_gsm}
    \end{subfigure}
    \hfill
    \begin{subfigure}[b]{0.24\textwidth}
        \centering
        \includegraphics[width=\linewidth]{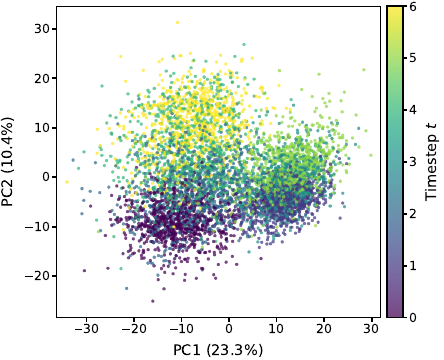}
        \caption{C arithmetic-reasoning}
        \label{fig:pca_c_gsm}
    \end{subfigure}
    \hfill
    \begin{subfigure}[b]{0.24\textwidth}
        \centering
        \includegraphics[width=\linewidth]{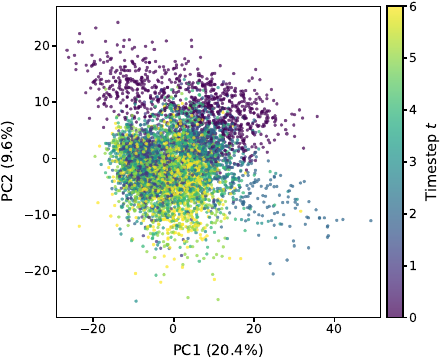}
        \caption{C$_u$ arithmetic-reasoning}
        \label{fig:pca_cu_gsm}
    \end{subfigure}
    \hfill
    \begin{subfigure}[b]{0.24\textwidth}
        \centering
        \includegraphics[width=\linewidth]{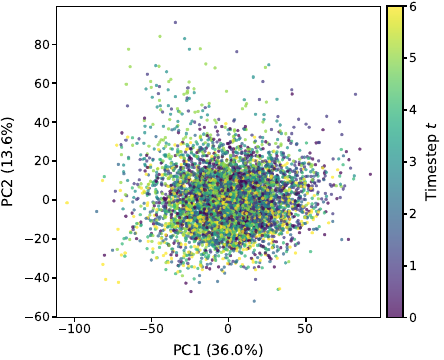}
        \caption{CODI arithmetic-reasoning}
        \label{fig:pca_codi_gsm}
    \end{subfigure}

    \caption{2D PCA projections of the latent space structure for the GPT-2 model family across the graph-hopping and arithmetic-reasoning tasks.}
    \label{fig:pca_gpt2}
\end{figure*}

\begin{figure*}[!t]
    \centering

    \begin{subfigure}[b]{0.24\textwidth}
        \centering
        \includegraphics[width=\linewidth]{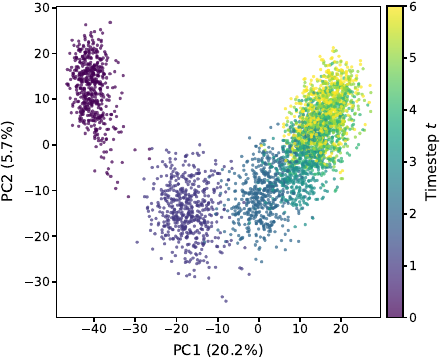}
        \caption{PaT graph-hopping}
        \label{fig:pca_pat_prosqa_llama}
    \end{subfigure}
    \hfill
    \begin{subfigure}[b]{0.24\textwidth}
        \centering
        \includegraphics[width=\linewidth]{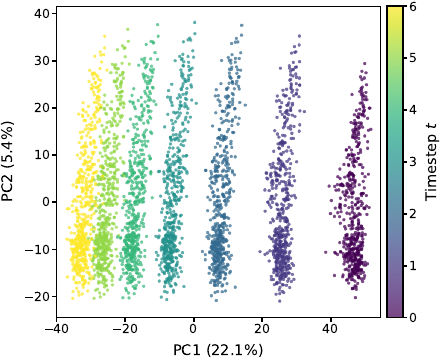}
        \caption{C graph-hopping}
        \label{fig:pca_c_prosqa_llama}
    \end{subfigure}
    \hfill
    \begin{subfigure}[b]{0.24\textwidth}
        \centering
        \includegraphics[width=\linewidth]{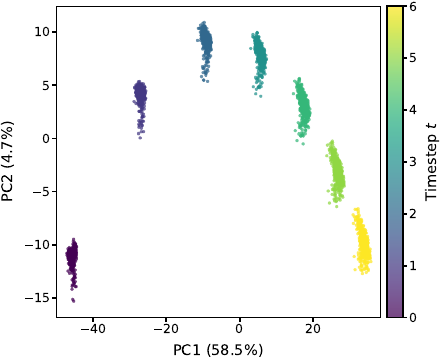}
        \caption{C$_u$ graph-hopping}
        \label{fig:pca_cu_prosqa_llama}
    \end{subfigure}
    \hfill
    \begin{subfigure}[b]{0.24\textwidth}
        \centering
        \includegraphics[width=\linewidth]{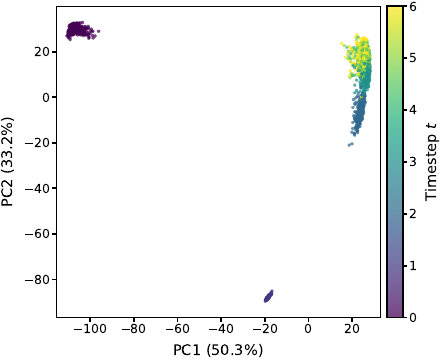}
        \caption{CODI graph-hopping}
        \label{fig:pca_codi_prosqa_llama}
    \end{subfigure}

    \vspace{0.5em}

    \begin{subfigure}[b]{0.24\textwidth}
        \centering
        \includegraphics[width=\linewidth]{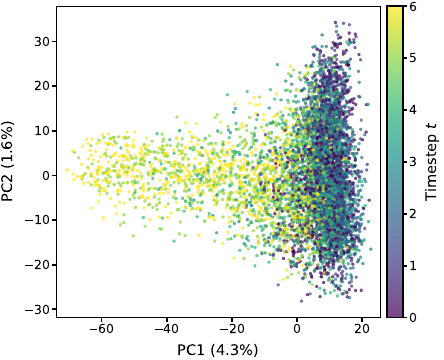}
        \caption{PaT arithmetic-reasoning}
        \label{fig:pca_pat_gsm_llama}
    \end{subfigure}
    \hfill
    \begin{subfigure}[b]{0.24\textwidth}
        \centering
        \includegraphics[width=\linewidth]{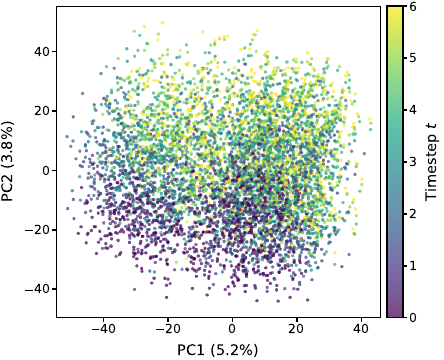}
        \caption{C arithmetic-reasoning}
        \label{fig:pca_c_gsm_llama}
    \end{subfigure}
    \hfill
    \begin{subfigure}[b]{0.24\textwidth}
        \centering
        \includegraphics[width=\linewidth]{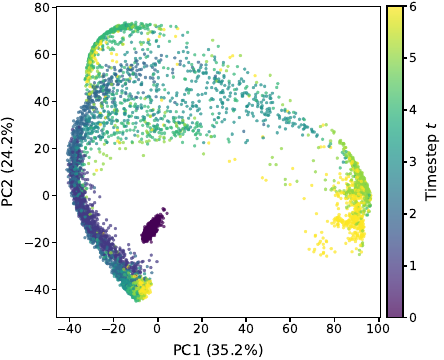}
        \caption{C$_u$ arithmetic-reasoning}
        \label{fig:pca_cu_gsm_llama}
    \end{subfigure}
    \hfill
    \begin{subfigure}[b]{0.24\textwidth}
        \centering
        \includegraphics[width=\linewidth]{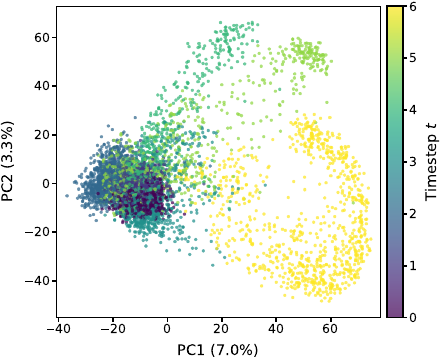}
        \caption{CODI arithmetic-reasoning}
        \label{fig:pca_codi_gsm_llama}
    \end{subfigure}

    \caption{2D PCA projections of the latent space structure for the Llama-3.2 model family across the graph-hopping and arithmetic-reasoning tasks.}
    \label{fig:pca_llama}
\end{figure*}

\textbf{GPT-2:} On \textbf{graph-hopping}, PaT and C are Var$_\text{inst}$ dominated, whereas C$_u$ is Var$_\text{time}$ dominated. Figure~\ref{fig:pca_gpt2} shows the two-dimensional PCA projections of the GPT-2 model family latents thoughts across tasks. In Figure~\ref{fig:pca_cu_prosqa}, we see C$_u$ forming nearly perfectly separable clusters corresponding to recurrence depth. For \textbf{arithmetic-reasoning}, all recurrence-based models are Var$_\text{residual}$ dominated (implicit in Table~\ref{tab:variance_decomposition} as the remaining variance). This is with the exception of PaT, which remains Var$_\text{inst}$ dominated with a temporal component that is 1-2 orders of magnitude smaller than the recurrence-based models.

\textbf{Llama-3.2:} On \textbf{graph-hopping}, PaT and C are Var$_\text{inst}$ dominated, whereas C$_u$ and CODI are Var$_\text{time}$ dominated. For \textbf{arithmetic-reasoning}, C and C$_u$ are heavily Var$_\text{inst}$ dominated. Figure~\ref{fig:pca_llama} shows the two-dimensional PCA projections of the Llama-3.2 model family latents thoughts across tasks. Figure~\ref{fig:pca_cu_prosqa_llama} shows the model forming perfectly separated clusters corresponding to recurrence depth for Llama-3.2 on the graph-hopping task again. In contrast, PaT and CODI are Var$_\text{residual}$ dominated (implicit in Table~\ref{tab:variance_decomposition_llama} as the remaining variance). The temporal component (Var$_\text{time}$) across all models on arithmetic-reasoning remains uniformly low.

\begin{figure*}
    \centering
    \begin{subfigure}{\linewidth}
        \centering
        \includegraphics[width=\linewidth]{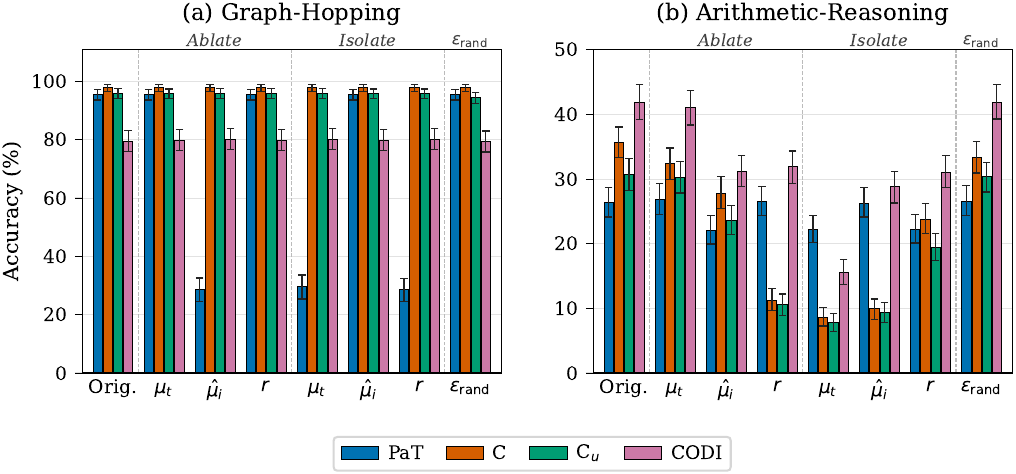}
        \caption{GPT-2.}
        \label{fig:mean_ablation_gpt2}
    \end{subfigure}

    \vspace{0.5em}

    \begin{subfigure}{\linewidth}
        \centering
        \includegraphics[width=\linewidth]{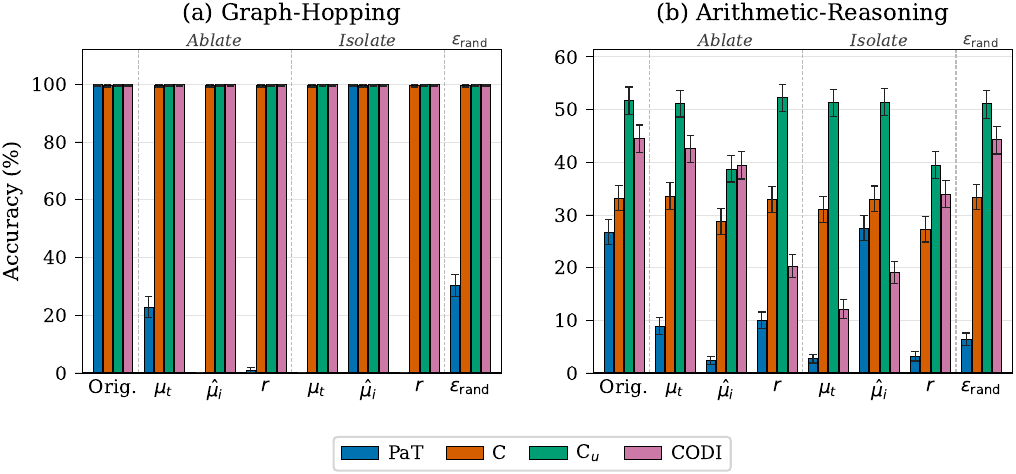}
        \caption{Llama-3.2.}
        \label{fig:mean_ablation_llama}
    \end{subfigure}

    \caption{Targeted interventions isolating and ablating specific variance components of the latent thoughts. \textbf{\textit{Takeaway:}} Thoughts carrying low influence are largely unaffected under perturbations. High influence carrying thoughts show varied dependence on components, affected least by the temporal and most by the residual. Temporal structure appears functionally inert across models and tasks.}
    \label{fig:mean_ablation}
\end{figure*}

\paragraph{Mean-Ablations and Isolations.} We apply six counterfactual interventions to the latent thoughts $h_{i,t}$ at inference to identify component-wise importance. \textit{Ablations} remove one component: temporal ($h'_{i,t} = h_{i,t} - \mu_t + \mu$), instance ($h'_{i,t} = h_{i,t} - \hat{\mu}_i + \mu$, with $\hat{\mu}_i = \frac{1}{K+1}\sum_{k=0}^K h_{i,k}$), or residual ($h'_{i,t} = \mu_t + \hat{\mu}_i - \mu$). \textit{Isolations} retain only one component: $h'_{i,t} = \mu_t$, $h'_{i,t} = \hat{\mu}_i$, or $h'_{i,t} = h_{i,t} - \mu_t - \hat{\mu}_i + 2\mu$, respectively. A \emph{matched-norm random control} adds isotropic noise to $h_{i,t}$ rescaled to the temporal-deviation norm: $h'_{i,t} = h_{i,t} + \varepsilon \cdot \|\mu_t - \mu\|/\|\varepsilon\|$ ($\varepsilon \sim \mathcal{N}(0, I_D)$). To prevent test-set leakage, all reference means ($\mu, \mu_t$) are computed from the training split. Results in Figure~\ref{fig:mean_ablation}.


\textbf{GPT-2:} On \textbf{graph-hopping}, models are robust to perturbations except PaT, which requires $\mu_i$ for stability. Notably, neither variance nor gradient-subspace interventions recover C$_u$’s thought-ablation drop, suggesting that its signal is distributed rather than localized. On \textbf{arithmetic reasoning}, PaT shows the same pattern. Ablating $\mu_t$ yields the smallest drop while isolating it causes the largest, indicating temporal structure is largely functionally inert---contrasting~\citet{wei2025sim}'s account of vocabulary-space drift as a key vulnerability. Ablating the residual component produces the largest drop, but its isolation has minimal impact. The matched-norm control preserves performance across all tasks.

\textbf{Llama-3.2:} \textbf{Graph-hopping} models are robust to perturbations except PaT, which degrades under all ablations, and all isolations except $\hat\mu_i$ which preserves full performance. On \textbf{arithmetic reasoning}, all models are affected with varying degree under all ablations and isolations, demonstrating the combined effect played by the components.




\section{Statistical Evaluation}
\label{app:statistical_tests}

This section presents the full statistical-tests for every experiment conducted in \S~\ref{sec:epiphenomena}, \S~\ref{sec:thought_use} and \S~\ref{sec:geometry}. Unless stated otherwise, all confidence intervals are 95\% percentile bootstrap intervals computed with 1{,}000 resamples over per-instance outcome vectors, and paired comparisons report exact two-sided McNemar $p$-values on discordant pairs $(b, c)$, where $b$ counts instances correct only under condition~A and $c$ counts instances correct only under condition~B. Significance markers follow the convention $^{*}p{<}0.05$, $^{**}p{<}0.01$, $^{***}p{<}0.001$. Where multiple McNemar tests are reported within a single table, we additionally report Benjamini–Hochberg FDR-adjusted $q$-values (bracketed), with significance stars on $q$ following the same convention.

\paragraph{Epiphenomenal Patterns in LRM Interpretability (\S~\ref{sec:epiphenomena}).}

\begin{table*}[!t]
\centering
\tiny
\setlength{\tabcolsep}{3pt}
\begin{tabular}{l cccccc cccccc}
\toprule
 & \multicolumn{6}{c}{$H/\log_2 N$} & \multicolumn{6}{c}{Top-1 correct (\%)} \\
\cmidrule(lr){2-7} \cmidrule(lr){8-13}
$k$ & B & CoT & PaT & C & C$_u$ & CODI & B & CoT & PaT & C & C$_u$ & CODI \\
\midrule
0 & $0.58 {\scriptstyle \pm 0.03}$ & $0.16 {\scriptstyle \pm 0.02}$ & $0.51 {\scriptstyle \pm 0.03}$ & $0.43 {\scriptstyle \pm 0.03}$ & $0.44 {\scriptstyle \pm 0.03}$ & $0.34 {\scriptstyle \pm 0.03}$ & $81.3 {\scriptstyle \pm 3.7}$ & $84.2 {\scriptstyle \pm 3.5}$ & $65.1 {\scriptstyle \pm 4.2}$ & $63.8 {\scriptstyle \pm 4.1}$ & $90.7 {\scriptstyle \pm 2.5}$ & $64.4 {\scriptstyle \pm 4.6}$ \\
1 & $0.33 {\scriptstyle \pm 0.03}$ & $0.15 {\scriptstyle \pm 0.02}$ & $0.51 {\scriptstyle \pm 0.03}$ & $0.47 {\scriptstyle \pm 0.03}$ & $0.46 {\scriptstyle \pm 0.03}$ & $0.33 {\scriptstyle \pm 0.03}$ & $68.8 {\scriptstyle \pm 4.1}$ & $83.9 {\scriptstyle \pm 3.6}$ & $64.9 {\scriptstyle \pm 4.4}$ & $65.1 {\scriptstyle \pm 4.1}$ & $90.7 {\scriptstyle \pm 2.5}$ & $64.4 {\scriptstyle \pm 4.4}$ \\
2 & $0.37 {\scriptstyle \pm 0.02}$ & $0.24 {\scriptstyle \pm 0.02}$ & $0.48 {\scriptstyle \pm 0.02}$ & $0.48 {\scriptstyle \pm 0.02}$ & $0.32 {\scriptstyle \pm 0.02}$ & $0.32 {\scriptstyle \pm 0.02}$ & $55.6 {\scriptstyle \pm 4.1}$ & $70.8 {\scriptstyle \pm 4.1}$ & $71.6 {\scriptstyle \pm 4.0}$ & $63.2 {\scriptstyle \pm 4.1}$ & $89.1 {\scriptstyle \pm 2.8}$ & $51.0 {\scriptstyle \pm 4.3}$ \\
3 & $0.39 {\scriptstyle \pm 0.03}$ & $0.26 {\scriptstyle \pm 0.02}$ & $0.23 {\scriptstyle \pm 0.03}$ & $0.26 {\scriptstyle \pm 0.03}$ & $0.37 {\scriptstyle \pm 0.03}$ & $0.22 {\scriptstyle \pm 0.03}$ & $51.3 {\scriptstyle \pm 4.7}$ & $65.1 {\scriptstyle \pm 4.6}$ & $88.6 {\scriptstyle \pm 3.2}$ & $85.2 {\scriptstyle \pm 3.4}$ & $86.6 {\scriptstyle \pm 3.3}$ & $66.7 {\scriptstyle \pm 4.3}$ \\
4 & $0.40 {\scriptstyle \pm 0.05}$ & $0.27 {\scriptstyle \pm 0.04}$ & $0.13 {\scriptstyle \pm 0.03}$ & $0.14 {\scriptstyle \pm 0.04}$ & $0.38 {\scriptstyle \pm 0.05}$ & $0.19 {\scriptstyle \pm 0.04}$ & $57.7 {\scriptstyle \pm 7.1}$ & $70.3 {\scriptstyle \pm 6.3}$ & $97.3 {\scriptstyle \pm 2.2}$ & $92.9 {\scriptstyle \pm 3.8}$ & $70.9 {\scriptstyle \pm 6.9}$ & $78.6 {\scriptstyle \pm 6.0}$ \\
\midrule
 & \multicolumn{6}{c}{$P(\text{correct})$} & \multicolumn{6}{c}{Cand.\ mass} \\
\cmidrule(lr){2-7} \cmidrule(lr){8-13}
$k$ & B & CoT & PaT & C & C$_u$ & CODI & B & CoT & PaT & C & C$_u$ & CODI \\
\midrule
0 & $0.05 {\scriptstyle \pm 0.01}$ & $0.36 {\scriptstyle \pm 0.04}$ & $0.00 {\scriptstyle \pm 0.00}$ & $0.00 {\scriptstyle \pm 0.00}$ & $0.84 {\scriptstyle \pm 0.02}$ & $0.02 {\scriptstyle \pm 0.01}$ & $0.07 {\scriptstyle \pm 0.01}$ & $0.40 {\scriptstyle \pm 0.04}$ & $0.00 {\scriptstyle \pm 0.00}$ & $0.00 {\scriptstyle \pm 0.00}$ & $0.95 {\scriptstyle \pm 0.01}$ & $0.03 {\scriptstyle \pm 0.01}$ \\
1 & $0.00 {\scriptstyle \pm 0.00}$ & $0.39 {\scriptstyle \pm 0.04}$ & $0.00 {\scriptstyle \pm 0.00}$ & $0.00 {\scriptstyle \pm 0.00}$ & $0.85 {\scriptstyle \pm 0.02}$ & $0.01 {\scriptstyle \pm 0.01}$ & $0.01 {\scriptstyle \pm 0.00}$ & $0.44 {\scriptstyle \pm 0.04}$ & $0.00 {\scriptstyle \pm 0.00}$ & $0.00 {\scriptstyle \pm 0.00}$ & $0.97 {\scriptstyle \pm 0.01}$ & $0.03 {\scriptstyle \pm 0.01}$ \\
2 & $0.00 {\scriptstyle \pm 0.00}$ & $0.09 {\scriptstyle \pm 0.02}$ & $0.01 {\scriptstyle \pm 0.01}$ & $0.00 {\scriptstyle \pm 0.00}$ & $0.71 {\scriptstyle \pm 0.03}$ & $0.01 {\scriptstyle \pm 0.01}$ & $0.01 {\scriptstyle \pm 0.00}$ & $0.10 {\scriptstyle \pm 0.02}$ & $0.01 {\scriptstyle \pm 0.01}$ & $0.00 {\scriptstyle \pm 0.00}$ & $0.81 {\scriptstyle \pm 0.02}$ & $0.03 {\scriptstyle \pm 0.01}$ \\
3 & $0.00 {\scriptstyle \pm 0.00}$ & $0.04 {\scriptstyle \pm 0.01}$ & $0.34 {\scriptstyle \pm 0.04}$ & $0.37 {\scriptstyle \pm 0.04}$ & $0.38 {\scriptstyle \pm 0.03}$ & $0.15 {\scriptstyle \pm 0.03}$ & $0.00 {\scriptstyle \pm 0.00}$ & $0.04 {\scriptstyle \pm 0.02}$ & $0.34 {\scriptstyle \pm 0.04}$ & $0.37 {\scriptstyle \pm 0.04}$ & $0.40 {\scriptstyle \pm 0.03}$ & $0.16 {\scriptstyle \pm 0.03}$ \\
4 & $0.00 {\scriptstyle \pm 0.00}$ & $0.02 {\scriptstyle \pm 0.02}$ & $0.58 {\scriptstyle \pm 0.06}$ & $0.63 {\scriptstyle \pm 0.07}$ & $0.23 {\scriptstyle \pm 0.05}$ & $0.16 {\scriptstyle \pm 0.05}$ & $0.00 {\scriptstyle \pm 0.00}$ & $0.03 {\scriptstyle \pm 0.02}$ & $0.59 {\scriptstyle \pm 0.07}$ & $0.63 {\scriptstyle \pm 0.07}$ & $0.24 {\scriptstyle \pm 0.05}$ & $0.16 {\scriptstyle \pm 0.05}$ \\
\bottomrule
\end{tabular}
\caption{Full results for superposition and BFS-probing across timesteps and models with statistical testing for the GPT-2 model family.}
\label{tab:appendix_bfs_probing_gpt2}
\end{table*}

\begin{table*}[!t]
\centering
\tiny
\setlength{\tabcolsep}{3pt}
\begin{tabular}{l cccccc cccccc}
\toprule
 & \multicolumn{6}{c}{$H/\log_2 N$} & \multicolumn{6}{c}{Top-1 correct (\%)} \\
\cmidrule(lr){2-7} \cmidrule(lr){8-13}
$k$ & B & CoT & PaT & C & C$_u$ & CODI & B & CoT & PaT & C & C$_u$ & CODI \\
\midrule
0 & $0.60 {\scriptstyle \pm 0.03}$ & $0.28 {\scriptstyle \pm 0.03}$ & $0.52 {\scriptstyle \pm 0.03}$ & $0.50 {\scriptstyle \pm 0.03}$ & $0.45 {\scriptstyle \pm 0.03}$ & $0.14 {\scriptstyle \pm 0.02}$ & $79.8 {\scriptstyle \pm 3.8}$ & $94.1 {\scriptstyle \pm 2.3}$ & $66.8 {\scriptstyle \pm 4.0}$ & $68.8 {\scriptstyle \pm 4.3}$ & $91.1 {\scriptstyle \pm 2.5}$ & $95.2 {\scriptstyle \pm 2.0}$ \\
1 & $0.41 {\scriptstyle \pm 0.03}$ & $0.33 {\scriptstyle \pm 0.03}$ & $0.52 {\scriptstyle \pm 0.03}$ & $0.49 {\scriptstyle \pm 0.03}$ & $0.44 {\scriptstyle \pm 0.03}$ & $0.16 {\scriptstyle \pm 0.02}$ & $69.2 {\scriptstyle \pm 4.2}$ & $88.7 {\scriptstyle \pm 2.8}$ & $67.0 {\scriptstyle \pm 4.2}$ & $66.8 {\scriptstyle \pm 4.2}$ & $91.1 {\scriptstyle \pm 2.5}$ & $92.8 {\scriptstyle \pm 2.4}$ \\
2 & $0.42 {\scriptstyle \pm 0.02}$ & $0.27 {\scriptstyle \pm 0.02}$ & $0.46 {\scriptstyle \pm 0.02}$ & $0.43 {\scriptstyle \pm 0.02}$ & $0.35 {\scriptstyle \pm 0.02}$ & $0.19 {\scriptstyle \pm 0.02}$ & $66.7 {\scriptstyle \pm 4.2}$ & $89.5 {\scriptstyle \pm 2.9}$ & $69.5 {\scriptstyle \pm 4.3}$ & $71.6 {\scriptstyle \pm 3.9}$ & $85.2 {\scriptstyle \pm 3.1}$ & $86.8 {\scriptstyle \pm 2.9}$ \\
3 & $0.49 {\scriptstyle \pm 0.03}$ & $0.22 {\scriptstyle \pm 0.02}$ & $0.25 {\scriptstyle \pm 0.03}$ & $0.34 {\scriptstyle \pm 0.02}$ & $0.28 {\scriptstyle \pm 0.03}$ & $0.17 {\scriptstyle \pm 0.02}$ & $67.9 {\scriptstyle \pm 4.4}$ & $90.7 {\scriptstyle \pm 2.6}$ & $84.3 {\scriptstyle \pm 3.3}$ & $84.1 {\scriptstyle \pm 3.4}$ & $86.3 {\scriptstyle \pm 3.1}$ & $86.8 {\scriptstyle \pm 3.2}$ \\
4 & $0.50 {\scriptstyle \pm 0.04}$ & $0.18 {\scriptstyle \pm 0.04}$ & $0.17 {\scriptstyle \pm 0.04}$ & $0.27 {\scriptstyle \pm 0.04}$ & $0.22 {\scriptstyle \pm 0.04}$ & $0.09 {\scriptstyle \pm 0.03}$ & $74.2 {\scriptstyle \pm 6.3}$ & $96.2 {\scriptstyle \pm 2.5}$ & $91.8 {\scriptstyle \pm 4.1}$ & $90.7 {\scriptstyle \pm 4.4}$ & $91.2 {\scriptstyle \pm 4.1}$ & $89.6 {\scriptstyle \pm 4.4}$ \\
\midrule
 & \multicolumn{6}{c}{$P(\text{correct})$} & \multicolumn{6}{c}{Cand.\ mass} \\
\cmidrule(lr){2-7} \cmidrule(lr){8-13}
$k$ & B & CoT & PaT & C & C$_u$ & CODI & B & CoT & PaT & C & C$_u$ & CODI \\
\midrule
0 & $0.12 {\scriptstyle \pm 0.01}$ & $0.24 {\scriptstyle \pm 0.03}$ & $0.00 {\scriptstyle \pm 0.00}$ & $0.03 {\scriptstyle \pm 0.01}$ & $0.54 {\scriptstyle \pm 0.03}$ & $0.24 {\scriptstyle \pm 0.03}$ & $0.15 {\scriptstyle \pm 0.01}$ & $0.25 {\scriptstyle \pm 0.03}$ & $0.01 {\scriptstyle \pm 0.00}$ & $0.05 {\scriptstyle \pm 0.01}$ & $0.60 {\scriptstyle \pm 0.03}$ & $0.24 {\scriptstyle \pm 0.03}$ \\
1 & $0.00 {\scriptstyle \pm 0.00}$ & $0.09 {\scriptstyle \pm 0.02}$ & $0.00 {\scriptstyle \pm 0.00}$ & $0.03 {\scriptstyle \pm 0.01}$ & $0.54 {\scriptstyle \pm 0.03}$ & $0.24 {\scriptstyle \pm 0.03}$ & $0.00 {\scriptstyle \pm 0.00}$ & $0.10 {\scriptstyle \pm 0.02}$ & $0.01 {\scriptstyle \pm 0.00}$ & $0.05 {\scriptstyle \pm 0.01}$ & $0.60 {\scriptstyle \pm 0.03}$ & $0.25 {\scriptstyle \pm 0.04}$ \\
2 & $0.02 {\scriptstyle \pm 0.00}$ & $0.14 {\scriptstyle \pm 0.02}$ & $0.02 {\scriptstyle \pm 0.01}$ & $0.14 {\scriptstyle \pm 0.02}$ & $0.10 {\scriptstyle \pm 0.01}$ & $0.12 {\scriptstyle \pm 0.03}$ & $0.02 {\scriptstyle \pm 0.00}$ & $0.14 {\scriptstyle \pm 0.02}$ & $0.03 {\scriptstyle \pm 0.01}$ & $0.20 {\scriptstyle \pm 0.02}$ & $0.11 {\scriptstyle \pm 0.01}$ & $0.12 {\scriptstyle \pm 0.03}$ \\
3 & $0.04 {\scriptstyle \pm 0.01}$ & $0.14 {\scriptstyle \pm 0.02}$ & $0.38 {\scriptstyle \pm 0.04}$ & $0.30 {\scriptstyle \pm 0.03}$ & $0.13 {\scriptstyle \pm 0.02}$ & $0.19 {\scriptstyle \pm 0.03}$ & $0.06 {\scriptstyle \pm 0.01}$ & $0.15 {\scriptstyle \pm 0.02}$ & $0.39 {\scriptstyle \pm 0.04}$ & $0.35 {\scriptstyle \pm 0.03}$ & $0.13 {\scriptstyle \pm 0.02}$ & $0.19 {\scriptstyle \pm 0.03}$ \\
4 & $0.03 {\scriptstyle \pm 0.01}$ & $0.17 {\scriptstyle \pm 0.04}$ & $0.60 {\scriptstyle \pm 0.06}$ & $0.39 {\scriptstyle \pm 0.05}$ & $0.20 {\scriptstyle \pm 0.04}$ & $0.32 {\scriptstyle \pm 0.06}$ & $0.04 {\scriptstyle \pm 0.01}$ & $0.17 {\scriptstyle \pm 0.04}$ & $0.60 {\scriptstyle \pm 0.06}$ & $0.41 {\scriptstyle \pm 0.05}$ & $0.20 {\scriptstyle \pm 0.04}$ & $0.32 {\scriptstyle \pm 0.06}$ \\
\bottomrule
\end{tabular}
\caption{Full results for superposition and BFS-probing across timesteps and models with statistical testing for the Llama-3.2 model family.}
\label{tab:appendix_bfs_probing_llama}
\end{table*}

\begin{table*}[h!]
\centering
\small
\begin{tabular}{ll ccc}
\toprule
\textbf{Model} & \textbf{Step} & \textbf{Hit Rate} & \textbf{Superpos.} & \textbf{Step Align.} \\
\midrule
B & 0 & 0.1 \textsubscript{(0.0, 0.2)} & 0.0 \textsubscript{(0.0, 0.0)} & 0.0 \textsubscript{(0.0, 0.0)} \\
 & 1 & 1.4 \textsubscript{(0.8, 2.1)} & 0.0 \textsubscript{(0.0, 0.0)} & 0.3 \textsubscript{(0.0, 0.6)} \\
 & 2 & 0.3 \textsubscript{(0.1, 0.6)} & 0.0 \textsubscript{(0.0, 0.0)} & 0.0 \textsubscript{(0.0, 0.0)} \\
 & 3 & 0.3 \textsubscript{(0.1, 0.6)} & 0.0 \textsubscript{(0.0, 0.0)} & 0.1 \textsubscript{(0.0, 0.2)} \\
 & 4 & 0.2 \textsubscript{(0.0, 0.5)} & 0.0 \textsubscript{(0.0, 0.0)} & 0.0 \textsubscript{(0.0, 0.0)} \\
 & 5 & 0.2 \textsubscript{(0.0, 0.4)} & 0.0 \textsubscript{(0.0, 0.0)} & 0.0 \textsubscript{(0.0, 0.0)} \\
 & 6 & 0.2 \textsubscript{(0.0, 0.4)} & 0.0 \textsubscript{(0.0, 0.0)} & 0.0 \textsubscript{(0.0, 0.0)} \\
\midrule
CoT & 0 & 51.6 \textsubscript{(49.0, 54.3)} & 15.4 \textsubscript{(13.4, 17.3)} & 35.2 \textsubscript{(32.4, 37.8)} \\
 & 1 & 7.1 \textsubscript{(5.8, 8.5)} & 0.6 \textsubscript{(0.2, 1.1)} & 1.7 \textsubscript{(1.0, 2.5)} \\
 & 2 & 40.6 \textsubscript{(38.0, 43.4)} & 11.7 \textsubscript{(10.0, 13.5)} & 9.7 \textsubscript{(8.1, 11.4)} \\
 & 3 & 10.3 \textsubscript{(8.8, 11.8)} & 1.3 \textsubscript{(0.8, 2.0)} & 0.6 \textsubscript{(0.2, 1.1)} \\
 & 4 & 39.4 \textsubscript{(36.8, 42.0)} & 11.5 \textsubscript{(9.8, 13.2)} & 2.0 \textsubscript{(1.3, 2.8)} \\
 & 5 & 12.1 \textsubscript{(10.5, 13.9)} & 1.7 \textsubscript{(1.0, 2.5)} & 0.2 \textsubscript{(0.0, 0.4)} \\
 & 6 & 36.2 \textsubscript{(33.4, 38.8)} & 10.5 \textsubscript{(8.9, 12.1)} & 0.0 \textsubscript{(0.0, 0.0)} \\
\midrule
PaT & 0 & 65.4 \textsubscript{(62.9, 67.9)} & 22.6 \textsubscript{(20.4, 24.9)} & 48.3 \textsubscript{(45.7, 51.0)} \\
 & 1 & 65.6 \textsubscript{(63.1, 68.1)} & 24.1 \textsubscript{(22.0, 26.5)} & 40.7 \textsubscript{(37.9, 43.4)} \\
 & 2 & 64.3 \textsubscript{(61.9, 66.9)} & 23.8 \textsubscript{(21.4, 26.1)} & 18.7 \textsubscript{(16.7, 20.9)} \\
 & 3 & 62.2 \textsubscript{(59.6, 64.7)} & 23.1 \textsubscript{(20.9, 25.4)} & 7.9 \textsubscript{(6.5, 9.5)} \\
 & 4 & 60.0 \textsubscript{(57.6, 62.6)} & 23.4 \textsubscript{(21.1, 25.8)} & 4.2 \textsubscript{(3.1, 5.2)} \\
 & 5 & 58.9 \textsubscript{(56.3, 61.5)} & 22.8 \textsubscript{(20.7, 25.1)} & 1.3 \textsubscript{(0.7, 1.9)} \\
 & 6 & 58.6 \textsubscript{(55.9, 61.1)} & 21.9 \textsubscript{(19.7, 24.1)} & 0.2 \textsubscript{(0.0, 0.4)} \\
\midrule
C & 0 & 29.4 \textsubscript{(27.0, 32.1)} & 4.2 \textsubscript{(3.1, 5.4)} & 17.7 \textsubscript{(15.7, 19.7)} \\
 & 1 & 73.0 \textsubscript{(70.6, 75.5)} & 20.8 \textsubscript{(18.7, 23.2)} & 20.8 \textsubscript{(18.7, 23.0)} \\
 & 2 & 19.7 \textsubscript{(17.6, 21.9)} & 2.9 \textsubscript{(2.1, 3.8)} & 4.5 \textsubscript{(3.5, 5.6)} \\
 & 3 & 71.6 \textsubscript{(69.3, 74.1)} & 18.8 \textsubscript{(16.5, 20.8)} & 6.3 \textsubscript{(5.0, 7.7)} \\
 & 4 & 30.4 \textsubscript{(28.0, 32.9)} & 5.5 \textsubscript{(4.3, 6.8)} & 1.5 \textsubscript{(0.8, 2.2)} \\
 & 5 & 64.8 \textsubscript{(62.3, 67.4)} & 16.6 \textsubscript{(14.5, 18.7)} & 0.7 \textsubscript{(0.2, 1.2)} \\
 & 6 & 29.2 \textsubscript{(26.7, 31.7)} & 6.1 \textsubscript{(4.8, 7.5)} & 0.2 \textsubscript{(0.0, 0.4)} \\
\midrule
C$_u$ & 0 & 44.4 \textsubscript{(41.8, 47.3)} & 11.1 \textsubscript{(9.5, 12.9)} & 28.3 \textsubscript{(26.0, 30.7)} \\
 & 1 & 44.4 \textsubscript{(41.8, 47.3)} & 11.1 \textsubscript{(9.5, 12.9)} & 16.3 \textsubscript{(14.2, 18.4)} \\
 & 2 & 52.6 \textsubscript{(50.0, 55.2)} & 11.1 \textsubscript{(9.5, 12.9)} & 6.6 \textsubscript{(5.3, 8.0)} \\
 & 3 & 26.1 \textsubscript{(23.7, 28.7)} & 4.8 \textsubscript{(3.6, 5.9)} & 1.5 \textsubscript{(0.9, 2.2)} \\
 & 4 & 50.7 \textsubscript{(48.0, 53.5)} & 10.9 \textsubscript{(9.3, 12.7)} & 1.2 \textsubscript{(0.6, 1.7)} \\
 & 5 & 30.1 \textsubscript{(27.7, 32.7)} & 3.6 \textsubscript{(2.7, 4.8)} & 0.2 \textsubscript{(0.0, 0.4)} \\
 & 6 & 46.6 \textsubscript{(44.0, 49.1)} & 13.1 \textsubscript{(11.3, 14.9)} & 0.2 \textsubscript{(0.0, 0.5)} \\
\midrule
CODI & 0 & 75.1 \textsubscript{(72.8, 77.2)} & 21.9 \textsubscript{(19.7, 24.1)} & 63.9 \textsubscript{(61.3, 66.3)} \\
 & 1 & 0.5 \textsubscript{(0.2, 0.8)} & 0.0 \textsubscript{(0.0, 0.0)} & 0.1 \textsubscript{(0.0, 0.2)} \\
 & 2 & 73.0 \textsubscript{(70.6, 75.5)} & 22.1 \textsubscript{(19.8, 24.4)} & 15.3 \textsubscript{(13.4, 17.2)} \\
 & 3 & 0.4 \textsubscript{(0.1, 0.8)} & 0.1 \textsubscript{(0.0, 0.2)} & 0.0 \textsubscript{(0.0, 0.0)} \\
 & 4 & 70.8 \textsubscript{(68.5, 73.2)} & 21.8 \textsubscript{(19.6, 23.9)} & 3.2 \textsubscript{(2.2, 4.2)} \\
 & 5 & 0.5 \textsubscript{(0.2, 0.8)} & 0.0 \textsubscript{(0.0, 0.0)} & 0.0 \textsubscript{(0.0, 0.0)} \\
 & 6 & 70.1 \textsubscript{(67.8, 72.6)} & 22.1 \textsubscript{(20.1, 24.4)} & 0.2 \textsubscript{(0.0, 0.4)} \\
\midrule
\multicolumn{5}{l}{\textbf{Pooled (all steps)}} \\
B & -- & 0.4 \textsubscript{(0.2, 0.5)} & 0.0 \textsubscript{(0.0, 0.0)} & -- \\
CoT & -- & 28.2 \textsubscript{(27.3, 29.1)} & 7.5 \textsubscript{(7.0, 8.0)} & -- \\
PaT & -- & 62.1 \textsubscript{(61.2, 63.2)} & 23.1 \textsubscript{(22.3, 23.9)} & -- \\
C & -- & 45.4 \textsubscript{(44.5, 46.6)} & 10.7 \textsubscript{(10.0, 11.4)} & -- \\
C$_u$ & -- & 42.1 \textsubscript{(41.2, 43.1)} & 9.4 \textsubscript{(8.8, 10.0)} & -- \\
CODI & -- & 41.5 \textsubscript{(40.4, 42.5)} & 12.6 \textsubscript{(11.8, 13.2)} & -- \\
\bottomrule
\end{tabular}
\caption{Full results for scratchpad-thinking probing across timesteps and models with statistical testing for the GPT-2 model family.}
\label{tab:scratchpad_gpt2}
\end{table*}

\begin{table*}[h!]
\centering
\small
\begin{tabular}{ll ccc}
\toprule
\textbf{Model} & \textbf{Step} & \textbf{Hit Rate} & \textbf{Superpos.} & \textbf{Step Align.} \\
\midrule
B & 0 & 34.0 \textsubscript{(31.4, 36.5)} & 7.8 \textsubscript{(6.3, 9.2)} & 18.4 \textsubscript{(16.2, 20.5)} \\
 & 1 & 34.0 \textsubscript{(31.4, 36.5)} & 7.8 \textsubscript{(6.3, 9.2)} & 12.0 \textsubscript{(10.3, 13.8)} \\
 & 2 & 35.0 \textsubscript{(32.5, 37.7)} & 7.0 \textsubscript{(5.6, 8.5)} & 7.6 \textsubscript{(6.1, 9.1)} \\
 & 3 & 34.9 \textsubscript{(32.3, 37.4)} & 6.8 \textsubscript{(5.5, 8.2)} & 4.1 \textsubscript{(3.0, 5.1)} \\
 & 4 & 31.0 \textsubscript{(28.5, 33.5)} & 6.1 \textsubscript{(4.8, 7.5)} & 2.0 \textsubscript{(1.3, 2.8)} \\
 & 5 & 25.7 \textsubscript{(23.4, 28.3)} & 4.2 \textsubscript{(3.2, 5.4)} & 0.2 \textsubscript{(0.0, 0.5)} \\
 & 6 & 20.1 \textsubscript{(18.0, 22.4)} & 2.8 \textsubscript{(1.9, 3.7)} & 0.1 \textsubscript{(0.0, 0.2)} \\
\midrule
CoT & 0 & 58.8 \textsubscript{(56.2, 61.3)} & 21.4 \textsubscript{(19.3, 23.6)} & 45.1 \textsubscript{(42.4, 47.5)} \\
 & 1 & 58.8 \textsubscript{(56.2, 61.3)} & 21.4 \textsubscript{(19.3, 23.6)} & 22.2 \textsubscript{(20.1, 24.5)} \\
 & 2 & 52.0 \textsubscript{(49.3, 54.7)} & 16.9 \textsubscript{(15.0, 19.0)} & 12.6 \textsubscript{(10.8, 14.4)} \\
 & 3 & 10.2 \textsubscript{(8.5, 11.8)} & 2.4 \textsubscript{(1.5, 3.2)} & 0.7 \textsubscript{(0.3, 1.2)} \\
 & 4 & 6.3 \textsubscript{(5.0, 7.5)} & 1.7 \textsubscript{(1.0, 2.4)} & 0.3 \textsubscript{(0.1, 0.6)} \\
 & 5 & 3.3 \textsubscript{(2.4, 4.2)} & 1.0 \textsubscript{(0.5, 1.5)} & 0.1 \textsubscript{(0.0, 0.2)} \\
 & 6 & 2.4 \textsubscript{(1.6, 3.2)} & 0.6 \textsubscript{(0.2, 1.1)} & 0.0 \textsubscript{(0.0, 0.0)} \\
\midrule
PaT & 0 & 59.0 \textsubscript{(56.3, 61.8)} & 17.0 \textsubscript{(14.8, 19.1)} & 43.7 \textsubscript{(41.0, 46.6)} \\
 & 1 & 65.4 \textsubscript{(62.9, 67.9)} & 21.6 \textsubscript{(19.5, 24.1)} & 25.0 \textsubscript{(22.7, 27.5)} \\
 & 2 & 68.6 \textsubscript{(66.3, 71.1)} & 23.2 \textsubscript{(21.1, 25.4)} & 13.8 \textsubscript{(12.0, 15.7)} \\
 & 3 & 65.9 \textsubscript{(63.3, 68.3)} & 22.5 \textsubscript{(20.2, 24.8)} & 7.1 \textsubscript{(5.8, 8.6)} \\
 & 4 & 60.3 \textsubscript{(57.6, 63.0)} & 20.4 \textsubscript{(18.1, 22.7)} & 2.4 \textsubscript{(1.5, 3.2)} \\
 & 5 & 46.5 \textsubscript{(44.0, 49.1)} & 12.5 \textsubscript{(10.8, 14.3)} & 0.8 \textsubscript{(0.3, 1.2)} \\
 & 6 & 41.4 \textsubscript{(38.8, 44.0)} & 9.5 \textsubscript{(7.9, 11.1)} & 0.2 \textsubscript{(0.0, 0.5)} \\
\midrule
C & 0 & 75.3 \textsubscript{(72.9, 77.6)} & 32.8 \textsubscript{(30.4, 35.2)} & 54.9 \textsubscript{(52.0, 57.6)} \\
 & 1 & 75.3 \textsubscript{(72.9, 77.6)} & 32.8 \textsubscript{(30.4, 35.2)} & 38.4 \textsubscript{(35.8, 41.1)} \\
 & 2 & 74.2 \textsubscript{(71.7, 76.5)} & 32.4 \textsubscript{(29.8, 35.0)} & 18.3 \textsubscript{(16.3, 20.4)} \\
 & 3 & 73.6 \textsubscript{(71.0, 75.9)} & 30.2 \textsubscript{(27.6, 32.8)} & 7.2 \textsubscript{(5.8, 8.5)} \\
 & 4 & 71.6 \textsubscript{(69.0, 73.9)} & 27.7 \textsubscript{(25.3, 30.3)} & 3.2 \textsubscript{(2.3, 4.2)} \\
 & 5 & 69.9 \textsubscript{(67.1, 72.3)} & 24.7 \textsubscript{(22.4, 27.1)} & 0.7 \textsubscript{(0.3, 1.2)} \\
 & 6 & 67.6 \textsubscript{(64.9, 70.0)} & 22.1 \textsubscript{(19.9, 24.4)} & 0.1 \textsubscript{(0.0, 0.2)} \\
\midrule
C$_u$ & 0 & 10.3 \textsubscript{(8.8, 11.9)} & 0.5 \textsubscript{(0.2, 1.0)} & 4.5 \textsubscript{(3.5, 5.8)} \\
 & 1 & 10.3 \textsubscript{(8.8, 11.9)} & 0.5 \textsubscript{(0.2, 1.0)} & 3.2 \textsubscript{(2.3, 4.2)} \\
 & 2 & 13.7 \textsubscript{(12.0, 15.6)} & 1.5 \textsubscript{(0.8, 2.2)} & 2.8 \textsubscript{(2.0, 3.7)} \\
 & 3 & 12.5 \textsubscript{(10.8, 14.4)} & 1.5 \textsubscript{(0.9, 2.2)} & 1.6 \textsubscript{(0.9, 2.3)} \\
 & 4 & 10.9 \textsubscript{(9.3, 12.6)} & 1.9 \textsubscript{(1.2, 2.7)} & 0.8 \textsubscript{(0.4, 1.4)} \\
 & 5 & 3.9 \textsubscript{(2.8, 4.9)} & 0.4 \textsubscript{(0.1, 0.8)} & 0.1 \textsubscript{(0.0, 0.2)} \\
 & 6 & 1.8 \textsubscript{(1.1, 2.5)} & 0.0 \textsubscript{(0.0, 0.0)} & 0.0 \textsubscript{(0.0, 0.0)} \\
\midrule
CODI & 0 & 71.8 \textsubscript{(69.3, 74.4)} & 18.4 \textsubscript{(16.3, 20.5)} & 51.8 \textsubscript{(49.0, 54.4)} \\
 & 1 & 19.8 \textsubscript{(17.8, 22.1)} & 2.7 \textsubscript{(1.8, 3.5)} & 6.9 \textsubscript{(5.6, 8.3)} \\
 & 2 & 0.8 \textsubscript{(0.4, 1.4)} & 0.0 \textsubscript{(0.0, 0.0)} & 0.2 \textsubscript{(0.0, 0.5)} \\
 & 3 & 44.3 \textsubscript{(41.8, 47.0)} & 12.0 \textsubscript{(10.1, 13.8)} & 7.4 \textsubscript{(6.0, 8.8)} \\
 & 4 & 19.5 \textsubscript{(17.2, 21.7)} & 3.2 \textsubscript{(2.2, 4.2)} & 0.7 \textsubscript{(0.3, 1.2)} \\
 & 5 & 1.8 \textsubscript{(1.2, 2.6)} & 0.2 \textsubscript{(0.0, 0.4)} & 0.0 \textsubscript{(0.0, 0.0)} \\
 & 6 & 53.3 \textsubscript{(50.4, 56.0)} & 17.9 \textsubscript{(15.9, 20.1)} & 0.1 \textsubscript{(0.0, 0.2)} \\
\midrule
\multicolumn{5}{l}{\textbf{Pooled (all steps)}} \\
B & -- & 30.7 \textsubscript{(29.7, 31.6)} & 6.1 \textsubscript{(5.6, 6.6)} & -- \\
CoT & -- & 27.4 \textsubscript{(26.5, 28.3)} & 9.4 \textsubscript{(8.7, 9.9)} & -- \\
PaT & -- & 58.2 \textsubscript{(57.2, 59.2)} & 18.1 \textsubscript{(17.3, 18.9)} & -- \\
C & -- & 72.5 \textsubscript{(71.6, 73.5)} & 29.0 \textsubscript{(28.0, 29.9)} & -- \\
C$_u$ & -- & 9.0 \textsubscript{(8.5, 9.6)} & 0.9 \textsubscript{(0.7, 1.1)} & -- \\
CODI & -- & 30.2 \textsubscript{(29.3, 31.1)} & 7.8 \textsubscript{(7.3, 8.3)} & -- \\
\bottomrule
\end{tabular}
\caption{Full results for scratchpad-thinking probing across timesteps and models with statistical testing for the Llama-3.2 model family.}
\label{tab:scratchpad_llama}
\end{table*}

Tables~\ref{tab:appendix_bfs_probing_gpt2} (GPT-2) and~\ref{tab:appendix_bfs_probing_llama} (Llama-3.2) report the results for case study 1 (superposition and BFS-like search on graph-hopping); Table~\ref{tab:scratchpad_gpt2} (GPT-2) and~\ref{tab:scratchpad_llama} (Llama-3.2) report results for case study 2 (scratchpad reasoning on arithmetic-reasoning via logit-lens).

\paragraph{When and How do LRMs use Latent Thoughts (\S~\ref{sec:thought_use})?}

\begin{table*}[h!]
\centering
\small
\setlength{\tabcolsep}{4pt}
\begin{tabular}{l ccc cc}
\toprule
Model & Acc$_{K_{\max}}$ [\% CI] & Acc$_{K{=}0}$ [\% CI] & $\Delta$ [\% CI] & McNemar $p$ & $n$ \\
\midrule
\multicolumn{6}{c}{\textbf{Graph-Hopping}} \\
\midrule
B & 2.4 [1.2, 3.8] & -- & -- & -- & 500 \\
CoT & 79.0 [75.4, 82.6] & -- & -- & -- & 500 \\
PaT & 95.4 [93.6, 97.2] & 95.6 [93.6, 97.2] & -0.2 [-0.6, 0.0] & 1.0000 (b=0, c=1) [q=1.0000] & 500 \\
C & 98.0 [96.6, 99.0] & 97.8 [96.4, 99.0] & +0.2 [0.0, 0.6] & 1.0000 (b=1, c=0) [q=1.0000] & 500 \\
C$_u$ & 96.0 [94.2, 97.6] & 91.8 [89.4, 94.0] & +4.2 [1.6, 6.8] & 0.0019$^{**}$ (b=32, c=11) [q=0.0030$^{**}$] & 500 \\
CODI & 79.6 [76.0, 83.2] & 79.8 [76.2, 83.4] & -0.2 [-1.2, 0.8] & 1.0000 (b=3, c=4) [q=1.0000] & 500 \\
\midrule
\multicolumn{6}{c}{\textbf{Arithmetic-Reasoning}} \\
\midrule
B & 1.4 [0.8, 2.0] & -- & -- & -- & 1319 \\
CoT & 41.9 [39.3, 44.6] & -- & -- & -- & 1319 \\
PaT & 26.4 [24.1, 28.7] & 21.4 [19.4, 23.7] & +5.0 [3.5, 6.7] & 0.0000$^{***}$ (b=95, c=29) [q=0.0000$^{***}$] & 1319 \\
C & 35.7 [33.4, 38.1] & 7.7 [6.4, 9.2] & +28.1 [25.6, 30.5] & 0.0000$^{***}$ (b=391, c=21) [q=0.0000$^{***}$] & 1319 \\
C$_u$ & 30.8 [28.3, 33.2] & 39.1 [36.4, 41.6] & -8.3 [-10.7, -6.0] & 0.0000$^{***}$ (b=74, c=184) [q=0.0000$^{***}$] & 1319 \\
CODI & 41.8 [39.2, 44.7] & 24.8 [22.4, 27.1] & +17.1 [14.4, 19.9] & 0.0000$^{***}$ (b=297, c=72) [q=0.0000$^{***}$] & 1319 \\
\bottomrule
\end{tabular}
\caption{Full results for latent thought ablation at test-time with statistical testing for the GPT-2 model family.}
\label{tab:ablation_stats_gpt2}
\end{table*}
\begin{table*}[h!]
\centering
\small
\setlength{\tabcolsep}{4pt}
\begin{tabular}{l ccc cc}
\toprule
Model & Acc$_{K_{\max}}$ [\% CI] & Acc$_{K{=}0}$ [\% CI] & $\Delta$ [\% CI] & McNemar $p$ & $n$ \\
\midrule
\multicolumn{6}{c}{\textbf{Graph-Hopping}} \\
\midrule
B & 2.2 [1.0, 3.6] & -- & -- & -- & 500 \\
CoT & 96.8 [95.2, 98.2] & -- & -- & -- & 500 \\
PaT & 99.8 [99.4, 100.0] & 99.8 [99.4, 100.0] & +0.0 [0.0, 0.0] & 1.0000 (b=0, c=0) [q=1.0000] & 500 \\
C & 99.6 [99.0, 100.0] & 99.6 [99.0, 100.0] & +0.0 [0.0, 0.0] & 1.0000 (b=0, c=0) [q=1.0000] & 500 \\
C$_u$ & 99.8 [99.4, 100.0] & 99.8 [99.4, 100.0] & +0.0 [0.0, 0.0] & 1.0000 (b=0, c=0) [q=1.0000] & 500 \\
CODI & 99.8 [99.4, 100.0] & 99.8 [99.4, 100.0] & +0.0 [0.0, 0.0] & 1.0000 (b=0, c=0) [q=1.0000] & 500 \\
\midrule
\multicolumn{6}{c}{\textbf{Arithmetic-Reasoning}} \\
\midrule
B & 7.7 [6.1, 9.1] & -- & -- & -- & 1319 \\
CoT & 53.3 [50.6, 55.9] & -- & -- & -- & 1319 \\
PaT & 26.8 [24.4, 29.2] & 3.9 [3.0, 4.9] & +22.8 [20.5, 25.4] & 0.0000$^{***}$ (b=315, c=14) [q=0.0000$^{***}$] & 1319 \\
C & 33.2 [30.8, 35.6] & 12.1 [10.5, 13.9] & +21.2 [18.6, 23.7] & 0.0000$^{***}$ (b=307, c=28) [q=0.0000$^{***}$] & 1319 \\
C$_u$ & 51.7 [49.1, 54.3] & 49.4 [46.8, 52.1] & +2.4 [0.5, 3.9] & 0.0171$^{*}$ (b=95, c=64) [q=0.0342$^{*}$] & 1319 \\
CODI & 44.6 [41.8, 47.1] & 23.5 [21.2, 25.8] & +21.1 [18.5, 23.6] & 0.0000$^{***}$ (b=317, c=39) [q=0.0000$^{***}$] & 1319 \\
\bottomrule
\end{tabular}
\caption{Full results for latent thought ablation at test-time with statistical testing for the Llama-3.2 model family.}
\label{tab:ablation_stats_llama}
\end{table*}

\begin{table*}[h!]
\centering
\tiny
\setlength{\tabcolsep}{2.5pt}
\begin{tabular}{l ccc c c c c r}
\toprule
Model & P$_{\text{full}}$ & P$_{\max}$ & P$_b$ & A$_b$ & $\overline{\text{KL}_{c\to c'}}$ & Wilc.\ $p$ & \makecell{McN.\ $p$\\ ($\tau{=}0.5$)} & $n$ \\
\midrule
\multicolumn{9}{c}{\textbf{Graph-Hopping}} \\
\midrule
PaT & 1.355 [0.977, 1.817] & 5.705 [5.008, 6.388] & 14.577 [13.697, 15.488] & 14.657 [13.791, 15.565] & 15.158 [14.280, 16.085] & 0.0000$^{***}$ & \makecell{0.0000$^{***}$ (0,421)\\ q=0.0000$^{***}$} & 500 \\
C & 1.554 [1.145, 1.992] & 5.315 [4.740, 5.934] & 14.204 [13.248, 15.077] & 14.143 [13.196, 15.032] & 15.320 [14.382, 16.198] & 0.0000$^{***}$ & \makecell{0.0000$^{***}$ (0,435)\\ q=0.0000$^{***}$} & 500 \\
C$_u$ & 1.694 [1.201, 2.208] & 4.499 [3.953, 5.085] & 13.776 [12.953, 14.662] & 13.528 [12.741, 14.351] & 13.960 [13.122, 14.835] & 0.0000$^{***}$ & \makecell{0.0000$^{***}$ (0,464)\\ q=0.0000$^{***}$} & 500 \\
CODI & 0.861 [0.607, 1.167] & 5.324 [4.755, 5.911] & 11.135 [10.410, 11.857] & 11.218 [10.509, 11.945] & 11.540 [10.831, 12.255] & 0.0000$^{***}$ & \makecell{0.0000$^{***}$ (0,455)\\ q=0.0000$^{***}$} & 500 \\
\midrule
\multicolumn{9}{c}{\textbf{Arithmetic-Reasoning}} \\
\midrule
PaT & 5.361 [5.162, 5.572] & 6.496 [6.255, 6.743] & 10.408 [10.089, 10.730] & 10.517 [10.195, 10.847] & 11.736 [11.390, 12.085] & 0.0000$^{***}$ & \makecell{0.0000$^{***}$ (136,423)\\ q=0.0000$^{***}$} & 1319 \\
\bottomrule
\end{tabular}
\caption{Full causal tracing results with statistical testing (prompt-positions) for the GPT-2 model family.}
\label{tab:causal_trace_prompt_gpt2}
\end{table*}

\begin{table*}[h!]
\centering
\tiny
\setlength{\tabcolsep}{2.5pt}
\begin{tabular}{l ccccccc}
\toprule
Model & T$_1$ & T$_2$ & T$_3$ & T$_4$ & T$_5$ & T$_6$ & T$_{\text{full}}$ \\
\midrule
\multicolumn{8}{c}{\textbf{Graph-Hopping}} \\
\midrule
PaT & 14.610 [13.735, 15.521] & 14.621 [13.747, 15.532] & 14.636 [13.767, 15.537] & 14.637 [13.768, 15.537] & 14.650 [13.783, 15.555] & 14.651 [13.786, 15.560] & 13.612 [12.721, 14.475] \\
C & 15.305 [14.369, 16.184] & 15.307 [14.371, 16.187] & 15.287 [14.332, 16.166] & 15.288 [14.332, 16.166] & 15.310 [14.373, 16.190] & 14.203 [13.252, 15.091] & 14.205 [13.255, 15.087] \\
C$_u$ & 13.842 [13.021, 14.730] & 13.846 [13.023, 14.735] & 13.779 [12.973, 14.640] & 13.808 [12.995, 14.681] & 13.808 [12.995, 14.681] & 13.761 [12.936, 14.622] & 13.661 [12.854, 14.504] \\
CODI & 11.192 [10.474, 11.918] & 11.234 [10.514, 11.963] & 11.221 [10.510, 11.940] & 11.241 [10.523, 11.965] & 11.228 [10.505, 11.964] & 11.249 [10.544, 11.987] & 11.033 [10.301, 11.750] \\
\midrule
\multicolumn{8}{c}{\textbf{Arithmetic-Reasoning}} \\
\midrule
PaT & 10.278 [9.967, 10.596] & 10.153 [9.838, 10.472] & 10.011 [9.691, 10.325] & 9.955 [9.642, 10.270] & 10.019 [9.710, 10.333] & 10.105 [9.794, 10.422] & 6.745 [6.483, 6.998] \\
C & 6.276 [6.061, 6.471] & 6.071 [5.871, 6.256] & 6.092 [5.886, 6.293] & 6.638 [6.413, 6.852] & 8.012 [7.754, 8.290] & 9.406 [9.112, 9.707] & 3.165 [3.006, 3.303] \\
C$_u$ & 5.254 [5.046, 5.469] & 5.019 [4.826, 5.222] & 5.005 [4.806, 5.212] & 5.049 [4.875, 5.252] & 5.975 [5.765, 6.210] & 7.655 [7.397, 7.942] & 2.812 [2.658, 2.986] \\
CODI & 8.105 [7.841, 8.376] & 8.324 [8.054, 8.601] & 8.193 [7.923, 8.478] & 8.241 [7.974, 8.521] & 8.317 [8.051, 8.615] & 9.895 [9.584, 10.220] & 5.449 [5.232, 5.664] \\
\bottomrule
\end{tabular}
\caption{Full causal tracing results with statistical testing (thought-positions) for the GPT-2 model family.}
\label{tab:causal_trace_thought_gpt2}
\end{table*}

\begin{table*}[h!]
\centering
\tiny
\setlength{\tabcolsep}{2.5pt}
\begin{tabular}{l ccc c c c c r}
\toprule
Model & P$_{\text{full}}$ & P$_{\max}$ & P$_b$ & A$_b$ & $\overline{\text{KL}_{c\to c'}}$ & Wilc.\ $p$ & \makecell{McN.\ $p$\\ ($\tau{=}0.5$)} & $n$ \\
\midrule
\multicolumn{9}{c}{\textbf{Graph-Hopping}} \\
\midrule
PaT & 4.305 [3.798, 4.829] & 10.939 [10.322, 11.513] & 14.356 [13.820, 14.844] & 15.731 [15.224, 16.209] & 17.452 [16.954, 17.875] & 0.0000$^{***}$ & \makecell{0.0000$^{***}$ (3,284)\\ q=0.0000$^{***}$} & 500 \\
C & 2.266 [1.942, 2.609] & 6.785 [6.434, 7.142] & 9.373 [8.980, 9.754] & 11.544 [11.183, 11.896] & 13.230 [12.883, 13.590] & 0.0000$^{***}$ & \makecell{0.0000$^{***}$ (1,348)\\ q=0.0000$^{***}$} & 500 \\
C$_u$ & 3.135 [2.617, 3.626] & 9.265 [8.599, 9.893] & 12.714 [12.111, 13.271] & 15.831 [15.192, 16.420] & 18.117 [17.475, 18.760] & 0.0000$^{***}$ & \makecell{0.0000$^{***}$ (0,330)\\ q=0.0000$^{***}$} & 500 \\
CODI & 0.323 [0.179, 0.506] & 6.008 [5.512, 6.477] & 14.996 [14.328, 15.601] & 15.156 [14.496, 15.767] & 15.263 [14.617, 15.884] & 0.0000$^{***}$ & \makecell{0.0000$^{***}$ (0,492)\\ q=0.0000$^{***}$} & 500 \\
\midrule
\multicolumn{9}{c}{\textbf{Arithmetic-Reasoning}} \\
\midrule
PaT & 1.660 [1.490, 1.842] & 2.571 [2.352, 2.785] & 3.913 [3.652, 4.181] & 0.552 [0.467, 0.637] & 4.561 [4.284, 4.841] & 0.0000$^{***}$ & \makecell{0.0000$^{***}$ (48,340)\\ q=0.0000$^{***}$} & 1318 \\
C & 4.386 [4.225, 4.550] & 5.840 [5.649, 6.057] & 6.995 [6.803, 7.208] & 2.268 [2.156, 2.383] & 8.535 [8.320, 8.780] & 0.0000$^{***}$ & \makecell{0.0000$^{***}$ (9,477)\\ q=0.0000$^{***}$} & 1319 \\
C$_u$ & 3.757 [3.627, 3.902] & 4.216 [4.079, 4.365] & 4.977 [4.835, 5.131] & 1.425 [1.346, 1.498] & 5.435 [5.292, 5.594] & 0.0000$^{***}$ & \makecell{0.0000$^{***}$ (15,238)\\ q=0.0000$^{***}$} & 1319 \\
CODI & 3.555 [3.413, 3.690] & 4.913 [4.754, 5.078] & 5.944 [5.789, 6.105] & 7.343 [7.175, 7.522] & 7.978 [7.778, 8.178] & 0.0000$^{***}$ & \makecell{0.0000$^{***}$ (58,646)\\ q=0.0000$^{***}$} & 1319 \\
\bottomrule
\end{tabular}
\caption{Full causal tracing results with statistical testing (prompt-positions) for the Llama-3.2 model family.}
\label{tab:causal_trace_prompt_llama}
\end{table*}

\begin{table*}[h!]
\centering
\tiny
\setlength{\tabcolsep}{2.5pt}
\begin{tabular}{l ccccccc}
\toprule
Model & T$_1$ & T$_2$ & T$_3$ & T$_4$ & T$_5$ & T$_6$ & T$_{\text{full}}$ \\
\midrule
\multicolumn{8}{c}{\textbf{Graph-Hopping}} \\
\midrule
PaT & 14.801 [14.303, 15.268] & 15.890 [15.364, 16.362] & 16.362 [15.850, 16.836] & 16.517 [15.993, 16.971] & 16.485 [15.947, 16.931] & 16.308 [15.807, 16.770] & 12.986 [12.472, 13.470] \\
C & 12.152 [11.845, 12.479] & 12.333 [12.026, 12.639] & 12.428 [12.108, 12.750] & 12.582 [12.245, 12.912] & 12.656 [12.303, 12.988] & 10.726 [10.349, 11.120] & 10.565 [10.200, 10.943] \\
C$_u$ & 15.266 [14.618, 15.933] & 17.039 [16.409, 17.691] & 17.374 [16.751, 18.009] & 17.561 [16.936, 18.184] & 17.850 [17.224, 18.478] & 14.989 [14.389, 15.583] & 13.129 [12.517, 13.711] \\
CODI & 15.107 [14.450, 15.717] & 15.156 [14.493, 15.766] & 15.154 [14.493, 15.766] & 15.173 [14.512, 15.783] & 15.156 [14.494, 15.767] & 15.197 [14.554, 15.808] & 15.107 [14.449, 15.717] \\
\midrule
\multicolumn{8}{c}{\textbf{Arithmetic-Reasoning}} \\
\midrule
PaT & 3.751 [3.502, 4.022] & 3.703 [3.461, 3.970] & 3.524 [3.277, 3.788] & 3.437 [3.197, 3.684] & 3.910 [3.653, 4.176] & 4.086 [3.834, 4.356] & 2.413 [2.206, 2.631] \\
C & 7.594 [7.398, 7.811] & 7.754 [7.553, 7.986] & 7.815 [7.610, 8.052] & 7.902 [7.700, 8.148] & 8.009 [7.803, 8.232] & 7.389 [7.191, 7.622] & 6.800 [6.609, 7.017] \\
C$_u$ & 5.184 [5.048, 5.345] & 5.187 [5.051, 5.349] & 5.206 [5.069, 5.367] & 5.210 [5.076, 5.371] & 5.252 [5.113, 5.409] & 4.804 [4.658, 4.957] & 4.700 [4.557, 4.857] \\
CODI & 6.450 [6.283, 6.621] & 6.292 [6.121, 6.466] & 7.005 [6.836, 7.173] & 7.166 [6.980, 7.346] & 7.095 [6.920, 7.272] & 7.603 [7.422, 7.791] & 5.692 [5.519, 5.860] \\
\bottomrule
\end{tabular}
\caption{Full causal tracing results with statistical testing (thought-positions) for the Llama-3.2 model family.}
\label{tab:causal_trace_thought_llama}
\end{table*}

\begin{table*}[h!]
\centering
\tiny
\setlength{\tabcolsep}{3pt}
\begin{tabular}{l ccc cc cc}
\toprule
Model & Acc$_{\text{orig}}$ [\% CI] & Acc$_{\text{grad}}$ [\% CI] & Acc$_{\text{rand}}$ [\% CI] & $\Delta_{\text{grad}}$ [\% CI] & McNemar$_{\text{grad}}$ $p$ & $\Delta_{\text{rand}}$ [\% CI] & McNemar$_{\text{rand}}$ $p$ \\
\midrule
\multicolumn{8}{c}{\textbf{Gold Subspace}} \\
\midrule
\multicolumn{8}{c}{\textbf{Graph-Hopping}} \\
\midrule
PaT & $95.4 {\scriptstyle \pm 1.8}$ & $95.8 {\scriptstyle \pm 1.7}$ & $95.4 {\scriptstyle \pm 1.0}$ & $-0.4 {\scriptstyle \pm 0.5}$ & 0.5000 (b=0, c=2) [q=0.8000] & $+0.0 {\scriptstyle \pm 0.0}$ & 1.0000 (b=0, c=0) [q=1.0000] \\
C & $98.0 {\scriptstyle \pm 1.2}$ & $98.0 {\scriptstyle \pm 1.2}$ & $98.0 {\scriptstyle \pm 0.7}$ & $+0.0 {\scriptstyle \pm 0.0}$ & 1.0000 (b=0, c=0) [q=1.0000] & $+0.0 {\scriptstyle \pm 0.0}$ & 1.0000 (b=0, c=0) [q=1.0000] \\
C$_u$ & $96.0 {\scriptstyle \pm 1.7}$ & $96.0 {\scriptstyle \pm 1.7}$ & $96.0 {\scriptstyle \pm 1.0}$ & $+0.0 {\scriptstyle \pm 0.0}$ & 1.0000 (b=0, c=0) [q=1.0000] & $+0.0 {\scriptstyle \pm 0.0}$ & 1.0000 (b=0, c=0) [q=1.0000] \\
CODI & $80.4 {\scriptstyle \pm 3.5}$ & $79.8 {\scriptstyle \pm 3.6}$ & $80.0 {\scriptstyle \pm 2.0}$ & $+0.6 {\scriptstyle \pm 0.9}$ & 0.3750 (b=4, c=1) [q=0.6667] & $+0.4 {\scriptstyle \pm 0.4}$ & 0.1094 (b=8, c=2) [q=0.2501] \\
\midrule
\multicolumn{8}{c}{\textbf{Arithmetic-Reasoning}} \\
\bottomrule
\end{tabular}
\caption{Full gradient-subspace ablation results with statistical testing for the GPT-2 model family.}
\label{tab:gradsubspace_ablation_stats_gpt2}
\end{table*}

\begin{table*}[h!]
\centering
\tiny
\setlength{\tabcolsep}{4pt}
\begin{tabular}{l l cccc c}
\toprule
Model & $\alpha$ & Flip$_{\text{grad}}$ [\% CI] & Flip$_{\text{rand}}$ [\% CI] & $\Delta$ [\% CI] & McNemar $p$ & $n$ \\
\midrule
\multicolumn{7}{c}{\textbf{Graph-Hopping}} \\
\midrule
PaT & 1.5 & $0.0 {\scriptstyle \pm 0.0}$ & $0.0 {\scriptstyle \pm 0.0}$ & $+0.0 {\scriptstyle \pm 0.0}$ & 1.0000 (b=0, c=0) [q=1.0000] & 500 \\
 & 2 & $0.2 {\scriptstyle \pm 0.3}$ & $0.0 {\scriptstyle \pm 0.0}$ & $+0.2 {\scriptstyle \pm 0.2}$ & 0.2500 (b=3, c=0) [q=0.3182] &  \\
 & 5 & $1.0 {\scriptstyle \pm 0.9}$ & $0.1 {\scriptstyle \pm 0.1}$ & $+0.9 {\scriptstyle \pm 0.5}$ & 0.0005$^{***}$ (b=15, c=1) [q=0.0008$^{***}$] &  \\
 & 10 & $2.0 {\scriptstyle \pm 1.3}$ & $0.2 {\scriptstyle \pm 0.2}$ & $+1.8 {\scriptstyle \pm 0.8}$ & 0.0000$^{***}$ (b=30, c=3) [q=0.0000$^{***}$] &  \\
 & 25 & $13.0 {\scriptstyle \pm 2.8}$ & $1.0 {\scriptstyle \pm 0.5}$ & $+12.0 {\scriptstyle \pm 1.7}$ & 0.0000$^{***}$ (b=183, c=3) [q=0.0000$^{***}$] &  \\
 & 50 & $29.6 {\scriptstyle \pm 4.0}$ & $1.0 {\scriptstyle \pm 0.5}$ & $+28.6 {\scriptstyle \pm 2.4}$ & 0.0000$^{***}$ (b=432, c=3) [q=0.0000$^{***}$] &  \\
\bottomrule
\end{tabular}
\caption{Full gradient-subspace amplification results with statistical testing for the GPT-2 model family (gold subspace).}
\label{tab:gradsubspace_amp_stats_gold_gpt2}
\end{table*}

\begin{table*}[h!]
\centering
\tiny
\setlength{\tabcolsep}{4pt}
\begin{tabular}{l l cccc c}
\toprule
Model & $\alpha$ & Flip$_{\text{grad}}$ [\% CI] & Flip$_{\text{rand}}$ [\% CI] & $\Delta$ [\% CI] & McNemar $p$ & $n$ \\
\midrule
\multicolumn{7}{c}{\textbf{Graph-Hopping}} \\
\midrule
PaT & 1.5 & $0.0 {\scriptstyle \pm 0.0}$ & $0.0 {\scriptstyle \pm 0.0}$ & $+0.0 {\scriptstyle \pm 0.0}$ & 1.0000 (b=0, c=0) [q=1.0000] & 500 \\
 & 2 & $0.2 {\scriptstyle \pm 0.3}$ & $0.0 {\scriptstyle \pm 0.0}$ & $+0.2 {\scriptstyle \pm 0.2}$ & 0.2500 (b=3, c=0) [q=0.3500] &  \\
 & 5 & $0.6 {\scriptstyle \pm 0.7}$ & $0.0 {\scriptstyle \pm 0.0}$ & $+0.6 {\scriptstyle \pm 0.4}$ & 0.0039$^{**}$ (b=9, c=0) [q=0.0062$^{**}$] &  \\
 & 10 & $13.2 {\scriptstyle \pm 2.8}$ & $0.2 {\scriptstyle \pm 0.2}$ & $+13.0 {\scriptstyle \pm 1.8}$ & 0.0000$^{***}$ (b=198, c=3) [q=0.0000$^{***}$] &  \\
 & 25 & $60.0 {\scriptstyle \pm 4.1}$ & $0.9 {\scriptstyle \pm 0.5}$ & $+59.1 {\scriptstyle \pm 2.6}$ & 0.0000$^{***}$ (b=890, c=3) [q=0.0000$^{***}$] &  \\
 & 50 & $72.2 {\scriptstyle \pm 4.0}$ & $0.9 {\scriptstyle \pm 0.5}$ & $+71.3 {\scriptstyle \pm 2.4}$ & 0.0000$^{***}$ (b=1072, c=3) [q=0.0000$^{***}$] &  \\
\bottomrule
\end{tabular}
\caption{Full gradient-subspace amplification results with statistical testing for the GPT-2 model family  (predicted-token subspace).}
\label{tab:gradsubspace_amp_stats_predtoken_gpt2}
\end{table*}

\begin{table*}[h!]
\centering
\tiny
\setlength{\tabcolsep}{3pt}
\begin{tabular}{l ccc cc cc}
\toprule
Model & Acc$_{\text{orig}}$ [\% CI] & Acc$_{\text{grad}}$ [\% CI] & Acc$_{\text{rand}}$ [\% CI] & $\Delta_{\text{grad}}$ [\% CI] & McNemar$_{\text{grad}}$ $p$ & $\Delta_{\text{rand}}$ [\% CI] & McNemar$_{\text{rand}}$ $p$ \\
\midrule
\multicolumn{8}{c}{\textbf{Gold Subspace}} \\
\midrule
\multicolumn{8}{c}{\textbf{Graph-Hopping}} \\
\midrule
PaT & $99.8 {\scriptstyle \pm 0.3}$ & $99.8 {\scriptstyle \pm 0.3}$ & $99.8 {\scriptstyle \pm 0.2}$ & $+0.0 {\scriptstyle \pm 0.0}$ & 1.0000 (b=0, c=0) [q=1.0000] & $+0.0 {\scriptstyle \pm 0.0}$ & 1.0000 (b=0, c=0) [q=1.0000] \\
C & $99.6 {\scriptstyle \pm 0.5}$ & $99.6 {\scriptstyle \pm 0.5}$ & $99.6 {\scriptstyle \pm 0.3}$ & $+0.0 {\scriptstyle \pm 0.0}$ & 1.0000 (b=0, c=0) [q=1.0000] & $+0.0 {\scriptstyle \pm 0.0}$ & 1.0000 (b=0, c=0) [q=1.0000] \\
C$_u$ & $99.8 {\scriptstyle \pm 0.3}$ & $99.8 {\scriptstyle \pm 0.3}$ & $99.8 {\scriptstyle \pm 0.2}$ & $+0.0 {\scriptstyle \pm 0.0}$ & 1.0000 (b=0, c=0) [q=1.0000] & $+0.0 {\scriptstyle \pm 0.0}$ & 1.0000 (b=0, c=0) [q=1.0000] \\
CODI & $99.8 {\scriptstyle \pm 0.3}$ & $99.8 {\scriptstyle \pm 0.3}$ & $99.8 {\scriptstyle \pm 0.2}$ & $+0.0 {\scriptstyle \pm 0.0}$ & 1.0000 (b=0, c=0) [q=1.0000] & $+0.0 {\scriptstyle \pm 0.0}$ & 1.0000 (b=0, c=0) [q=1.0000] \\
\midrule
\multicolumn{8}{c}{\textbf{Arithmetic-Reasoning}} \\
\bottomrule
\end{tabular}
\caption{Full gradient-subspace ablation results with statistical testing for the Llama-3.2 model family.}
\label{tab:gradsubspace_ablation_stats_llama}
\end{table*}

\begin{table*}[h!]
\centering
\tiny
\setlength{\tabcolsep}{4pt}
\begin{tabular}{l l cccc c}
\toprule
Model & $\alpha$ & Flip$_{\text{grad}}$ [\% CI] & Flip$_{\text{rand}}$ [\% CI] & $\Delta$ [\% CI] & McNemar $p$ & $n$ \\
\midrule
\multicolumn{7}{c}{\textbf{Graph-Hopping}} \\
\midrule
PaT & 1.5 & $0.0 {\scriptstyle \pm 0.0}$ & $0.0 {\scriptstyle \pm 0.0}$ & $+0.0 {\scriptstyle \pm 0.0}$ & 1.0000 (b=0, c=0) [q=1.0000] & 500 \\
 & 2 & $0.0 {\scriptstyle \pm 0.0}$ & $0.0 {\scriptstyle \pm 0.0}$ & $+0.0 {\scriptstyle \pm 0.0}$ & 1.0000 (b=0, c=0) [q=1.0000] &  \\
 & 5 & $0.0 {\scriptstyle \pm 0.0}$ & $0.0 {\scriptstyle \pm 0.0}$ & $+0.0 {\scriptstyle \pm 0.0}$ & 1.0000 (b=0, c=0) [q=1.0000] &  \\
 & 10 & $0.0 {\scriptstyle \pm 0.0}$ & $0.0 {\scriptstyle \pm 0.0}$ & $+0.0 {\scriptstyle \pm 0.0}$ & 1.0000 (b=0, c=0) [q=1.0000] &  \\
 & 25 & $0.8 {\scriptstyle \pm 0.7}$ & $0.0 {\scriptstyle \pm 0.0}$ & $+0.8 {\scriptstyle \pm 0.4}$ & 0.0005$^{***}$ (b=12, c=0) [q=0.0007$^{***}$] &  \\
 & 50 & $0.0 {\scriptstyle \pm 0.0}$ & $0.1 {\scriptstyle \pm 0.1}$ & $-0.1 {\scriptstyle \pm 0.1}$ & 1.0000 (b=0, c=1) [q=1.0000] &  \\
\bottomrule
\end{tabular}
\caption{Full gradient-subspace amplification results with statistical testing for the Llama-3.2 model family (gold subspace).}
\label{tab:gradsubspace_amp_stats_gold_llama}
\end{table*}

\begin{table*}[h!]
\centering
\tiny
\setlength{\tabcolsep}{4pt}
\begin{tabular}{l l cccc c}
\toprule
Model & $\alpha$ & Flip$_{\text{grad}}$ [\% CI] & Flip$_{\text{rand}}$ [\% CI] & $\Delta$ [\% CI] & McNemar $p$ & $n$ \\
\midrule
\multicolumn{7}{c}{\textbf{Graph-Hopping}} \\
\midrule
PaT & 1.5 & $0.0 {\scriptstyle \pm 0.0}$ & $0.0 {\scriptstyle \pm 0.0}$ & $+0.0 {\scriptstyle \pm 0.0}$ & 1.0000 (b=0, c=0) [q=1.0000] & 500 \\
 & 2 & $0.0 {\scriptstyle \pm 0.0}$ & $0.0 {\scriptstyle \pm 0.0}$ & $+0.0 {\scriptstyle \pm 0.0}$ & 1.0000 (b=0, c=0) [q=1.0000] &  \\
 & 5 & $0.0 {\scriptstyle \pm 0.0}$ & $0.0 {\scriptstyle \pm 0.0}$ & $+0.0 {\scriptstyle \pm 0.0}$ & 1.0000 (b=0, c=0) [q=1.0000] &  \\
 & 10 & $1.2 {\scriptstyle \pm 0.9}$ & $0.0 {\scriptstyle \pm 0.0}$ & $+1.2 {\scriptstyle \pm 0.5}$ & 0.0000$^{***}$ (b=18, c=0) [q=0.0000$^{***}$] &  \\
 & 25 & $80.8 {\scriptstyle \pm 3.4}$ & $13.1 {\scriptstyle \pm 1.6}$ & $+67.7 {\scriptstyle \pm 2.6}$ & 0.0000$^{***}$ (b=1046, c=30) [q=0.0000$^{***}$] &  \\
 & 50 & $100.0 {\scriptstyle \pm 0.0}$ & $65.4 {\scriptstyle \pm 2.4}$ & $+34.6 {\scriptstyle \pm 2.4}$ & 0.0000$^{***}$ (b=519, c=0) [q=0.0000$^{***}$] &  \\
\bottomrule
\end{tabular}
\caption{Full gradient-subspace amplification results with statistical testing for the Llama-3.2 model family (predicted-token subspace).}
\label{tab:gradsubspace_amp_stats_predtoken_llama}
\end{table*}

Tables~\ref{tab:ablation_stats_gpt2} (GPT-2) and~\ref{tab:ablation_stats_llama} (Llama-3.2) report results on the latent thought ablation at test-time experiment; Tables~\ref{tab:causal_trace_prompt_gpt2} \&~\ref{tab:causal_trace_thought_gpt2} (prompt and thought-positions respectively for GPT-2), and Tables~\ref{tab:causal_trace_prompt_llama} \&~\ref{tab:causal_trace_thought_llama} (prompt and thought-positions respectively for Llama-3.2) on the causal-tracing experiment; and Tables~\ref{tab:gradsubspace_ablation_stats_gpt2}, \ref{tab:gradsubspace_amp_stats_gold_gpt2} \&~\ref{tab:gradsubspace_amp_stats_predtoken_gpt2} (ablation and amplification, gold and predicted-token subspace respectively, for GPT-2), and Tables~\ref{tab:gradsubspace_ablation_stats_llama}, \ref{tab:gradsubspace_amp_stats_gold_llama} \&~\ref{tab:gradsubspace_amp_stats_predtoken_llama} (ablation and amplification, gold and predicted-token subspace respectively, for Llama-3.2) on the gradient-subspace interventions experiment.

\paragraph{The Dynamics and Geometry of Latent Thoughts (\S~\ref{sec:geometry}).}

\begin{table*}[h!]
\centering
\tiny
\setlength{\tabcolsep}{4pt}
\begin{tabular}{llrrrrrrrr}
\toprule
& & \multicolumn{4}{c}{\textbf{Graph-Hopping}} & \multicolumn{4}{c}{\textbf{Arithmetic-Reasoning}} \\
\cmidrule(lr){3-6}\cmidrule(lr){7-10}
Proj. & Model & Identity & Mean & Linear & MLP & Identity & Mean & Linear & MLP \\
\midrule
\multirow{4}{*}{Full} & PaT & 0.999 [0.999, 0.999] & -0.003 & 0.597 [0.595, 0.599] & 0.384 [0.365, 0.399] & 0.867 [0.861, 0.873] & -0.013 & 0.716 [0.713, 0.720] & 0.697 [0.695, 0.699] \\
 & C & 0.992 [0.991, 0.992] & -0.002 & 0.653 [0.651, 0.655] & 0.477 [0.463, 0.488] & -0.658 [-0.670, -0.648] & -0.022 & 0.256 [0.253, 0.258] & 0.428 [0.425, 0.430] \\
 & C$_u$ & -0.646 [-0.677, -0.616] & -0.000 & 0.745 [0.740, 0.748] & 0.795 [0.792, 0.797] & -0.713 [-0.726, -0.701] & -0.032 & 0.255 [0.251, 0.258] & 0.439 [0.436, 0.441] \\
 & CODI & 0.813 [0.801, 0.826] & -0.002 & 0.621 [0.618, 0.624] & 0.884 [0.883, 0.885] & -0.683 [-0.693, -0.675] & -0.023 & 0.395 [0.390, 0.399] & 0.480 [0.477, 0.483] \\
\midrule
\multirow{4}{*}{\makecell{Subspace\\(Gold)}} & PaT & -0.466 [-0.485, -0.452] & -0.000 & 0.971 [0.970, 0.971] & 0.996 [0.996, 0.996] & 0.264 [0.257, 0.270] & -0.009 & 0.605 [0.602, 0.609] & 0.780 [0.778, 0.783] \\
 & C & 0.775 [0.764, 0.786] & -0.001 & 0.686 [0.681, 0.690] & 0.976 [0.975, 0.977] & -0.893 [-0.906, -0.880] & -0.018 & 0.250 [0.247, 0.253] & 0.441 [0.438, 0.444] \\
 & C$_u$ & -0.758 [-0.796, -0.728] & -0.000 & 0.921 [0.919, 0.923] & 0.968 [0.968, 0.969] & -0.932 [-0.945, -0.920] & -0.030 & 0.281 [0.278, 0.284] & 0.475 [0.472, 0.478] \\
 & CODI & 0.296 [0.270, 0.322] & -0.003 & 0.771 [0.766, 0.776] & 0.974 [0.973, 0.975] & -1.076 [-1.105, -1.050] & -0.032 & 0.236 [0.231, 0.240] & 0.341 [0.337, 0.346] \\
\midrule
\multirow{4}{*}{\makecell{Subspace\\(Pred)}} & PaT & -0.538 [-0.559, -0.520] & -0.000 & 0.967 [0.967, 0.968] & 0.995 [0.995, 0.995] & 0.220 [0.213, 0.226] & -0.009 & 0.606 [0.603, 0.610] & 0.785 [0.783, 0.787] \\
 & C & 0.673 [0.649, 0.694] & -0.001 & 0.657 [0.651, 0.661] & 0.984 [0.983, 0.984] & -0.922 [-0.936, -0.910] & -0.018 & 0.258 [0.255, 0.261] & 0.458 [0.455, 0.461] \\
 & C$_u$ & -0.936 [-0.970, -0.911] & -0.000 & 0.931 [0.930, 0.933] & 0.968 [0.967, 0.969] & -0.952 [-0.967, -0.939] & -0.033 & 0.277 [0.273, 0.280] & 0.484 [0.481, 0.488] \\
 & CODI & 0.279 [0.245, 0.313] & -0.004 & 0.716 [0.711, 0.721] & 0.973 [0.972, 0.974] & -1.065 [-1.093, -1.039] & -0.030 & 0.250 [0.245, 0.254] & 0.365 [0.361, 0.369] \\
\bottomrule
\end{tabular}
\caption{Full results for Markovianity of thought trajectories experiment with statistical testing for the GPT-2 model family.}
\label{tab:markovianity_gpt2}
\end{table*}

\begin{table*}[h!]
\centering
\tiny
\setlength{\tabcolsep}{4pt}
\begin{tabular}{llrrrrrrrr}
\toprule
& & \multicolumn{4}{c}{\textbf{Graph-Hopping}} & \multicolumn{4}{c}{\textbf{Arithmetic-Reasoning}} \\
\cmidrule(lr){3-6}\cmidrule(lr){7-10}
Proj. & Model & Identity & Mean & Linear & MLP & Identity & Mean & Linear & MLP \\
\midrule
\multirow{4}{*}{Full} & PaT & 0.826 [0.820, 0.833] & -0.002 & 0.880 [0.879, 0.881] & 0.707 [0.705, 0.708] & 0.432 [0.419, 0.445] & -0.023 & 0.528 [0.521, 0.535] & 0.284 [0.282, 0.287] \\
 & C & 0.887 [0.885, 0.890] & -0.002 & 0.914 [0.913, 0.915] & 0.724 [0.723, 0.726] & 0.965 [0.965, 0.965] & -0.028 & 0.913 [0.912, 0.913] & 0.631 [0.630, 0.632] \\
 & C$_u$ & 0.776 [0.771, 0.780] & -0.002 & 0.902 [0.900, 0.902] & 0.852 [0.851, 0.853] & 0.398 [0.384, 0.411] & -0.008 & 0.800 [0.797, 0.803] & 0.891 [0.889, 0.892] \\
 & CODI & -2.086 [-2.257, -1.907] & -0.001 & 0.862 [0.858, 0.866] & 0.863 [0.861, 0.864] & -0.488 [-0.496, -0.481] & -0.034 & 0.465 [0.461, 0.470] & 0.449 [0.446, 0.452] \\
\midrule
\multirow{4}{*}{\makecell{Subspace\\(Gold)}} & PaT & -0.760 [-0.788, -0.736] & -0.000 & 0.865 [0.862, 0.868] & 0.958 [0.957, 0.960] & -1.505 [-1.553, -1.458] & -0.017 & 0.517 [0.510, 0.525] & 0.596 [0.591, 0.601] \\
 & C & 0.330 [0.308, 0.348] & -0.001 & 0.540 [0.532, 0.547] & 0.855 [0.850, 0.860] & 0.661 [0.657, 0.664] & -0.034 & 0.911 [0.910, 0.911] & 0.895 [0.894, 0.895] \\
 & C$_u$ & 0.396 [0.371, 0.418] & -0.003 & 0.336 [0.329, 0.341] & 0.880 [0.874, 0.885] & -0.668 [-0.678, -0.660] & -0.003 & 0.848 [0.846, 0.851] & 0.959 [0.959, 0.960] \\
 & CODI & -0.794 [-0.870, -0.723] & -0.001 & 0.226 [0.223, 0.229] & 0.795 [0.788, 0.801] & -0.725 [-0.734, -0.717] & -0.029 & 0.484 [0.479, 0.489] & 0.562 [0.558, 0.565] \\
\midrule
\multirow{4}{*}{\makecell{Subspace\\(Pred)}} & PaT & -0.458 [-0.472, -0.448] & -0.000 & 0.922 [0.920, 0.923] & 0.961 [0.960, 0.962] & -0.952 [-0.988, -0.917] & -0.017 & 0.465 [0.459, 0.472] & 0.604 [0.599, 0.608] \\
 & C & 0.273 [0.245, 0.297] & -0.001 & 0.452 [0.442, 0.461] & 0.840 [0.833, 0.847] & 0.664 [0.660, 0.668] & -0.034 & 0.912 [0.911, 0.912] & 0.930 [0.929, 0.930] \\
 & C$_u$ & 0.377 [0.351, 0.401] & -0.003 & 0.332 [0.325, 0.338] & 0.880 [0.874, 0.885] & -0.799 [-0.813, -0.786] & -0.003 & 0.840 [0.838, 0.843] & 0.957 [0.957, 0.958] \\
 & CODI & -0.696 [-0.767, -0.628] & -0.001 & 0.172 [0.168, 0.176] & 0.777 [0.769, 0.784] & -0.765 [-0.775, -0.757] & -0.031 & 0.486 [0.481, 0.491] & 0.564 [0.561, 0.567] \\
\bottomrule
\end{tabular}
\caption{Full results for Markovianity of thought trajectories experiment with statistical testing for the Llama-3.2 model family.}
\label{tab:markovianity_llama}
\end{table*}

Tables~\ref{tab:markovianity_gpt2} (GPT-2) and~\ref{tab:markovianity_llama} (Llama-3.2) report results on the Markovianity of thought trajectories experiment; Tables~\ref{tab:subspace_dim_full_gpt2} (gold subspace) \&~\ref{tab:subspace_geometry_gpt2_pred} (predicted-token subspace) (GPT-2), and Tables~\ref{tab:subspace_dim_full_llama} (gold subspace) \&~\ref{tab:subspace_geometry_llama_pred} (predicted-token subspace) (Llama-3.2) on the geometric stability of gradient-subspaces experiment.
\section{Models, Controls, and Training Details}
\label{app:models_and_hyperparameters}

In this appendix, we provide details on the dataset, models, and training details.

\subsection{Task and Dataset}
\label{app:dataset_motivation}

\paragraph{Tasks.}
In this work, we use two case studies corresponding to prominent mechanistic claims about latent reasoning models.

Prior work has interpreted latent trajectories as breadth-first-search-like reasoning over graph frontiers in the graph-hopping task~\citep{coconut,reasoningbysuperposition25}. More recent works have investigated this claim more carefully, questioning its validity~\citep{cui2026how,rizvi-martel2026the}. All these works used ProsQA~\citep{coconut} as the benchmark for this investigation, which makes it a natural choice for our study. Each ProsQA instance is a directed acyclic graph of logical relations paired with a query and a gold target node; the dataset comprises $17{,}886$ training, $300$ validation, and $500$ test instances.

Prior work has interpreted decodable intermediate values as evidence that latent thoughts act as a compressed thought scratchpad in arithmetic reasoning~\citep{shen2025codi}. More recent works also question the validity of this claim, raising the possibility of shortcut behavior or task-specific heuristics~\citep{cui2026how,liang2026latentcotmodelsthinkstepbystep,LatentReasoningInterpretable2026,wei2025sim}. All these works used GSM8k~\citep{cobbe2021training} as one of the benchmarks for this investigation, which makes it a natural choice for our study. Each GSM8k instance is a grade-school arithmetic word problem paired with a step-by-step solution; we use the GSM8K-Aug variant of~\citet{deng2023implicitchainthoughtreasoning}, which augments each problem with a compressed chain-of-thought decomposition, giving $385{,}620$ training, $500$ validation, and $1{,}319$ test instances.

Together, the two tasks (graph traversal and arithmetic reasoning) allow us to evaluate whether observational patterns of latent reasoning are robust across qualitatively different reasoning settings.

\paragraph{Training data.} All models are trained separately for each task. Following~\citet{coconut} and~\citet{shen2025codi}, we train on the GSM8K-Aug training data of~\citet{deng2023implicitchainthoughtreasoning} for GSM8k and on the original ProsQA training split~\citep{coconut} for ProsQA. The held out test splits ($500$ for ProsQA, $1{,}319$ for GSM8k) are used for all reported results.

\subsection{Role of Each Model and Control}
\label{app:model_roles}

Table~\ref{tab:model_roles} summarizes the role of each model in our design. The controls are not used only as task-performance baselines: each removes or perturbs one factor that could otherwise make an observable latent-state pattern look like evidence for a latent reasoning mechanism.

\begin{table}[!htbp]
\centering
\scriptsize
\setlength{\tabcolsep}{3pt}
\begin{tabular}{p{0.16\linewidth}p{0.22\linewidth}p{0.50\linewidth}}
\toprule
\textbf{Model} & \textbf{Role} & \textbf{Question addressed} \\
\midrule
\textbf{Coconut (C)}
& Target LRM
& Does staged recurrent latent computation exhibit the BFS-like graph-frontier patterns attributed to Coconut? \\

\textbf{CODI}
& Target LRM
& Do distilled latent rationale states contain decodable arithmetic intermediates that are also behaviorally relevant? \\

\textbf{PaT}
& Recurrence control
& Do the same patterns persist when Coconut's format and curriculum are kept but recurrent latent-state feedback is removed? \\

\textbf{Coconut$_u$ (C$_u$)}
& Curriculum control
& Do the patterns survive a controlled perturbation of Coconut's staged curriculum while the recurrent architecture is preserved? \\

\textbf{Base (B)}
& Observational control
& Can the probes recover apparent structure without task-specific latent-thought training? \\

\textbf{Explicit-CoT (CoT)}
& Observational control
& Can similar patterns arise from ordinary task solving or explicit-rationale supervision rather than latent-thought mechanisms? \\
\bottomrule
\end{tabular}
\caption{Role of each model and control in the experimental design.}
\label{tab:model_roles}
\end{table}

\paragraph{Target latent reasoning models.}
Coconut~\citep{coconut} and CODI~\citep{shen2025codi} are the target LRMs because they instantiate the two mechanistic claims studied in this work: BFS-like latent search on graph-hopping and scratchpad-like latent arithmetic. Both insert dedicated latent thought positions, but train them differently. Coconut gradually replaces explicit reasoning segments with recurrent latent states through a staged curriculum. CODI compresses textual rationales into latent representations by distillation. Studying both lets us test whether our conclusions are tied to one training recipe or hold across two representative ways of constructing latent thought states.

\paragraph{Pause-as-thought as a recurrence control.}
PaT is designed to isolate recurrent latent-state feedback. It follows Coconut's input format and staged curriculum, inspired by pause-token training~\citep{goyal2024think}, but replaces feedback from previous hidden states with learned thought-position embeddings. Thus, PaT preserves the thought slots, task formatting, and curriculum pressures that may shape observable representations, while removing the mechanism by which Coconut feeds latent states back into subsequent latent computation. If PaT reproduces Coconut-like patterns, those patterns are not sufficient evidence that recurrence is the cause.

\paragraph{Coconut$_u$ as a curriculum control.}
Coconut$_u$ preserves Coconut's recurrent latent architecture and number of thought positions, but perturbs the order in which curriculum stages are sampled. At each training step, the model follows the current stage with probability $1-u$ and samples a non-current stage with probability $u$. We set $u=0.3$ to create a moderate perturbation: the model still trains mostly on the intended stage ($70\%$ of updates), keeping the comparison close to Coconut, while the remaining updates are frequent enough to test whether the reported patterns depend on a nearly deterministic stage schedule. This control separates effects of recurrence from effects of the curriculum trajectory.

\paragraph{Base GPT-2 \& Llama-3.2 and Explicit-CoT GPT-2 \& Llama-3.2 as observational controls.}
Base and Explicit-CoT controls are included only in the observational analyses of \S~\ref{sec:epiphenomena}. Base tests whether the readout methods themselves can recover apparently interpretable structure from a model without task-specific latent-thought training. Explicit-CoT tests whether similar structure appears after ordinary explicit-rationale supervision. These controls assess the specificity of observational readouts. They are excluded from causal and geometric analyses because those analyses intervene on dedicated latent thought positions, which these models are not trained to use.

\subsection{Training Details}
\label{app:training_details}

Table~\ref{tab:baseline_training_hyperparameters} reports the training configuration for the explicit-CoT baseline and the Coconut-family models: PaT, Coconut, and Coconut$_u$. Table~\ref{tab:codi_training_hyperparameters} then reports the CODI configuration\inlinecomment{ for both GPT-2 and Llama-3.2-1B-Instruct}. CODI is separately listed since it uses a different training recipe.

For the baseline table, $c$ denotes the number of thought tokens per reasoning step. Effective batch size is computed as batch size per GPU $\times$ gradient accumulation steps $\times$ number of GPUs. Coconut and Coconut$_u$ share the same hyperparameters for a given dataset; Coconut$_u$ differs only by using stage-mixing probability $u=0.3$, while standard Coconut uses $u=0.0$. GPT-2 runs use bf16 disabled and weight decay $0.01$; Llama runs use bf16 and backbone-specific learning rates as reported in the tables.

Table~\ref{tab:codi_training_hyperparameters} reports the CODI hyperparameters for GPT-2 and Llama-3.2-1B. These models use
LoRA fine-tuning with a projection head, following the configuration of~\citet{shen2025codi}. The ProsQA CODI settings follow the
configuration of~\citet{cui2026how}.\inlinecomment{ For Llama-3.2-1B-Instruct, we keep the same
CODI structure and latent count, but adjust the projection dimension to the
backbone hidden size and use the Llama-specific loss weight and learning rate
reported in the table.}


\begin{table*}[h!]
    \centering
    \small
    \setlength{\tabcolsep}{4pt}
    \begin{tabular}{lcccccccc}
    \toprule
    \textbf{Dataset} & \textbf{$c$} & \textbf{Eps./stage} & \textbf{Max stage} &
    \textbf{$u$} & \textbf{Resume} & \textbf{Eff.\ batch} & \textbf{Epochs} & \textbf{LR} \\
    \midrule
    \multicolumn{9}{l}{\textit{GPT-2}} \\
    \addlinespace[2pt]
    \multicolumn{9}{l}{\textbf{CoT}} \\
    GSM8K-Aug & 0 & 1 & 0 & 0.0     & 0 & 128 & 25 & $1 \times 10^{-4}$ \\
    ProsQA    & 0 & 1 & 0 & 0.0     & 0 & 128 & 50 & $1 \times 10^{-4}$ \\
    \midrule
    \multicolumn{9}{l}{\textbf{PaT}} \\
    GSM8K-Aug & 2 & 3 & 3 & 0.0     & 3 & 128 & 25 & $1 \times 10^{-4}$ \\
    ProsQA    & 1 & 5 & 6 & 0.0     & 0 & 128 & 50 & $1 \times 10^{-4}$ \\
    \midrule
    \multicolumn{9}{l}{\textbf{Coconut}} \\
    GSM8K-Aug & 2 & 3 & 3 & 0.0/0.3 & 3 & 128 & 25 & $1 \times 10^{-4}$ \\
    ProsQA    & 1 & 5 & 6 & 0.0/0.3 & 0 & 256 & 50 & $1 \times 10^{-4}$ \\
    \midrule
    \multicolumn{9}{l}{\textit{Llama-3.2-1B-Instruct}} \\
    \addlinespace[2pt]
    \multicolumn{9}{l}{\textbf{CoT}} \\
    GSM8K-Aug & 0 & 1 & 0 & 0.0     & 0 & 128    & 10 & $5 \times 10^{-5}$ \\
    ProsQA    & 0 & 1 & 0 & 0.0     & 0 & 64     & 50 & $5 \times 10^{-5}$ \\
    \midrule
    \multicolumn{9}{l}{\textbf{PaT}} \\
    GSM8K-Aug & 2 & 3 & 3 & 0.0     & 3 & 128    & 25 & $5 \times 10^{-5}$ \\
    ProsQA    & 1 & 5 & 6 & 0.0     & 0 & 64     & 50 & $5 \times 10^{-5}$ \\
    \midrule
    \multicolumn{9}{l}{\textbf{Coconut}} \\
    GSM8K-Aug & 2 & 3 & 3 & 0.0/0.3 & 3 & 128    & 25 & $5 \times 10^{-5}$ \\
    ProsQA    & 1 & 5 & 6 & 0.0/0.3 & 0 & 64/128 & 50 & $5 \times 10^{-5}$ \\
    \bottomrule
    \end{tabular}
    \caption{Hyperparameters used to train the GPT-2 and Llama-3.2-1B-Instruct CoT,
    PaT, Coconut, and Coconut$_u$ models. Effective batch size is computed as
    $\text{batch\_size\_training} \times \text{gradient\_accumulation\_steps} 
    \times N_{\text{GPUs}}$. Coconut rows report both the standard setting
    ($u=0.0$) and the stage-mixed Coconut$_u$ setting ($u=0.3$). For ProsQA Coconut on Llama, effective batch size is 64 at $u=0.0$ and 128 at $u=0.3$.}
    \label{tab:baseline_training_hyperparameters}
    \end{table*}

\begin{table*}[h!]
    \centering
    \small
    \setlength{\tabcolsep}{4pt}
    \begin{tabular}{lcccc}
    \toprule
    \textbf{Hyperparameter} &
    \multicolumn{2}{c}{\textbf{GPT-2}} &
    \multicolumn{2}{c}{\textbf{Llama-3.2-1B-Instruct}} \\
    \cmidrule(lr){2-3}\cmidrule(lr){4-5}
    & \textbf{GSM8K-Aug} & \textbf{ProsQA} & \textbf{GSM8K-Aug} & \textbf{ProsQA} \\
    \midrule
    Model max length & 1024 & 1024 & 1024 & 1024 \\
    Precision & bf16 & bf16 & bf16 & bf16 \\
    CODI loss weight & 1.0 & 1.0 & 20 & 20 \\
    Include last CoT & False & False & False & False \\
    Number of latents & 6 & 6 & 6 & 6 \\
    Use projection & True & True & True & True \\
    Projection dimension & 768 & 768 & 2048 & 2048 \\
    Projection dropout & 0.0 & 0.0 & 0.0 & 0.0 \\
    Use LoRA & True & True & True & True \\
    LoRA rank & 128 & 128 & 128 & 128 \\
    LoRA alpha & 32 & 32 & 32 & 32 \\
    Learning rate & $3 \times 10^{-3}$ & $1 \times 10^{-3}$ & $2 \times 10^{-4}$ & $2 \times 10^{-4}$ \\
    LR scheduler & Cosine & Cosine & Cosine & Cosine \\
    Warmup ratio & 0.03 & 0.03 & 0.03 & 0.03 \\
    Optimizer & AdamW & AdamW & AdamW & AdamW \\
    Total Batch size & 128 & 128 & 128 & 128 \\
    Weight decay & 0.1 & 0.01 & 0.1 & 0.01 \\
    Gradient clipping & 2.0 & 2.0 & 2.0 & 2.0 \\
    Epochs & 40 & 20 & 30 & 20 \\
    \bottomrule
    \end{tabular}
    \caption{Hyperparameters used to train the GPT-2 and Llama-3.2-1B-Instruct CODI models.}
    \label{tab:codi_training_hyperparameters}
    \end{table*}


\paragraph{Model size.}
All GPT-2 models are built on the pretrained GPT-2 small ($124$M parameters;~\citealp{radford2019language}). The Llama models are built on Llama-3.2-1B-Instruct ($1.24$B parameters;~\cite{llama3herdmodels2024}). CODI uses LoRA fine-tuning (rank $128$) on these backbones; all other models are fully fine-tuned. Full per-model, per-task hyperparameters appear in Tables~\ref{tab:baseline_training_hyperparameters} and~\ref{tab:codi_training_hyperparameters}.

\end{document}